\documentclass[shortAfour,sageh,times]{sagej}

\usepackage{moreverb,url}

\usepackage[colorlinks,bookmarksopen,bookmarksnumbered,citecolor=black,urlcolor=black]{hyperref}

\newcommand\BibTeX{{\rmfamily B\kern-.05em \textsc{i\kern-.025em b}\kern-.08em
T\kern-.1667em\lower.7ex\hbox{E}\kern-.125emX}}

\usepackage{times}
\usepackage{amsmath,amssymb}
\usepackage{tensor}
\usepackage{graphicx}
\usepackage{caption}
\usepackage{subcaption}
\usepackage{overpic}
\graphicspath{{./figs/}}
\usepackage{rotating}
\usepackage{verbatim}
\usepackage{color}
\usepackage{soul}
\usepackage{bm}
\usepackage{secdot}
\usepackage{float}
\usepackage{subcaption}
\usepackage{todonotes}
\usepackage[ruled,vlined]{algorithm2e}

\DeclareCaptionSubType*[Alph]{table}
\DeclareCaptionLabelFormat{mystyle}{Table~\bothIfFirst{#1}{ ̃}#2}
\DeclareMathOperator*{\argmin}{arg\!\min}
\DeclareMathOperator*{\argmax}{arg\!\max}
\newcommand*\diff{\mathop{}\!\mathrm{d}}

\begin{document}

\runninghead{Mukadam et al.}

\title{Continuous-time Gaussian process motion planning via probabilistic inference}

\author{Mustafa Mukadam\affilnum{*}, Jing Dong\affilnum{*}, Xinyan Yan, Frank Dellaert and Byron Boots}
\affiliation{Institute for Robotics \& Intelligent Machines, Georgia Institute of Technology, Atlanta, GA, USA.\\ \affilnum{*}Equal contribution to this article.}
\corrauth{Mustafa Mukadam, Georgia Institute of Technology, Atlanta, GA, USA.}
\email{mmukadam3@gatech.edu}

%%%%%%%%%%%%%%%%%%%%%%%%%%%%%%%%%%%%%%%%%%%%%%%%%%%%%%%%%%%%%%%%%%%%%%%%%%%%%%%%%%%%%%%%%%%%%%%%%%%%%%%%%%%%%%%%

\begin{abstract}
\textit{We introduce a novel formulation of motion planning, for continuous-time trajectories, as probabilistic inference. We first show how smooth continuous-time trajectories can be represented by a small number of states using sparse Gaussian process (GP) models. We next develop an efficient gradient-based optimization algorithm that exploits this sparsity and Gaussian process interpolation. We call this algorithm the Gaussian Process Motion Planner (GPMP).  We then detail how motion planning problems can be formulated as probabilistic inference on a factor graph. This forms the basis for GPMP2, a very efficient algorithm that combines GP representations of trajectories with fast, structure-exploiting inference via numerical optimization. Finally, we extend GPMP2 to an incremental algorithm, iGPMP2, that can efficiently replan when conditions change. We benchmark our algorithms against several sampling-based and trajectory optimization-based motion planning algorithms on planning problems in multiple environments. Our evaluation reveals that GPMP2 is several times faster than previous algorithms while retaining robustness. We also benchmark iGPMP2 on replanning problems, and show that it can find successful solutions in a fraction of the time required by GPMP2 to replan from scratch.}
\end{abstract}


\maketitle

%%%%%%%%%%%%%%%%%%%%%%%%%%%%%%%%%%%%%%%%%%%%%%%%%%%%%%%%%%%%%%%%%%%%%%%%%%%%%%%%%%%%%%%%%%%%%%%%%%%%%%%%%%%%%%%%

\section{Introduction}\label{sec:intro}

Motion planning is a key tool in robotics, used to find trajectories of robot states that achieve a desired task. While searching for a solution, motion planners evaluate trajectories based on two criteria: \emph{feasiblity} and \emph{optimality}. The exact notion of feasibility and optimality can vary depending on the system, tasks, and other problem-specific requirements. In general, feasibility evaluates a trajectory based on whether or not it respects the robot or task-specific constraints such as avoiding obstacles, while reaching the desired goal. In other words, feasibility is often binary: a trajectory is feasible or it is not. In contrast with feasibility, optimality often evaluates the quality of trajectories without reference to task-specific constraints. For example, optimality may refer to the smoothness of a trajectory and encourage the motion planner to minimize dynamical criteria like velocity or acceleration. A variety of motion planning algorithms have been proposed to find trajectories that are both feasible and optimal. These approaches can be roughly divided into two categories: sampling-based algorithms and trajectory optimization algorithms.

Sampling-based algorithms~\citep{kavraki1996probabilistic,kuffner2000rrt,lavalle2006planning} can effectively find feasible trajectories for high-dimensional systems but the trajectories often exhibit jerky and redundant motion and therefore require post processing to address optimality. Although optimal planners~\citep{karaman2010incremental} have been proposed, they are computationally inefficient on high-dimensional problems with challenging constraints. 

Trajectory optimization algorithms~\citep{ratliff2009chomp,zucker2013chomp,kalakrishnan2011stomp,he2013multigrid,byravan2014space,Marinho-RSS-16} minimize an objective function that encourages trajectories to be both feasible and optimal. A drawback of these approaches is that, in practice, a fine discretization of the trajectory is necessary to integrate cost information when reasoning about thin obstacles and tight constraints. Additionally, trajectory optimization is locally optimal, and may need to be rerun with different initial conditions to find a feasible solution, which can incur high computational cost. A solution to this latter problem is to initialize trajectory optimization with the solution discovered by a sampling-based algorithm.

TrajOpt~\citep{schulman2013finding,schulman2014motion} attempts to avoid finely discretized trajectories by formulating trajectory optimization as sequential quadratic programming.  It achieves reduced computational costs by parameterizing the trajectory with a small number of states and employing continuous-time collision checking. However, due to the discrete-time representation of the trajectory, a sparse solution may need post-processing for execution and may not remain collision-free. In other words, a fine discretization may still be necessary on problems in complex environments.

A continuous-time trajectory representation can avoid some of these challenges to yield a more efficient approach~\citep{Elbanhawi15itits,Marinho-RSS-16}. In this work, we adopt a continuous-time representation of trajectories; specifically, we view trajectories as functions that map time to robot state. We assume these functions are sampled from a Gaussian process (GP)~\citep{rasmussen2006gaussian}. We will show that GPs can inherently provide a notion of trajectory optimality through a \emph{prior}. Efficient structure-exploiting GP regression (GPR) can be used to query the trajectory at any time of interest in $O(1)$. Using this representation, we develop a gradient-based optimization algorithm called GPMP (Gaussian Process Motion Planner) that can efficiently overcome the large computational costs of fine discretization while still maintaining smoothness in the result.

Through the GP formulation, we can view motion planning as probabilistic inference~\citep{toussaint2009robot,toussaint2010bayesian}. Similar to how the notion of trajectory optimality is captured by a \emph{prior} on trajectories,  the notion of feasibility can also be viewed probabilistically as well and encoded in a likelihood function.  Bayesian inference can then be used to compute a solution to our motion planning problem efficiently through the use of factor graphs~\citep{kschischang2001factor}. The duality between inference and optimization allows us to perform efficient inference on factor graphs by solving sparse least squares problems, thereby exploiting the structure of the underlying system. Similar techniques have been used to solve large-scale Simultaneous Localization and Mapping (SLAM) problems~\citep{dellaert2006square}. With this key insight we can use preexisting efficient optimization tools developed by the SLAM community, and use them in the context of motion planning. This results in the GPMP2 algorithm,  which is more efficient than previous motion planning algorithms.

\begin{figure}
\centering
\includegraphics[width=0.44\columnwidth]{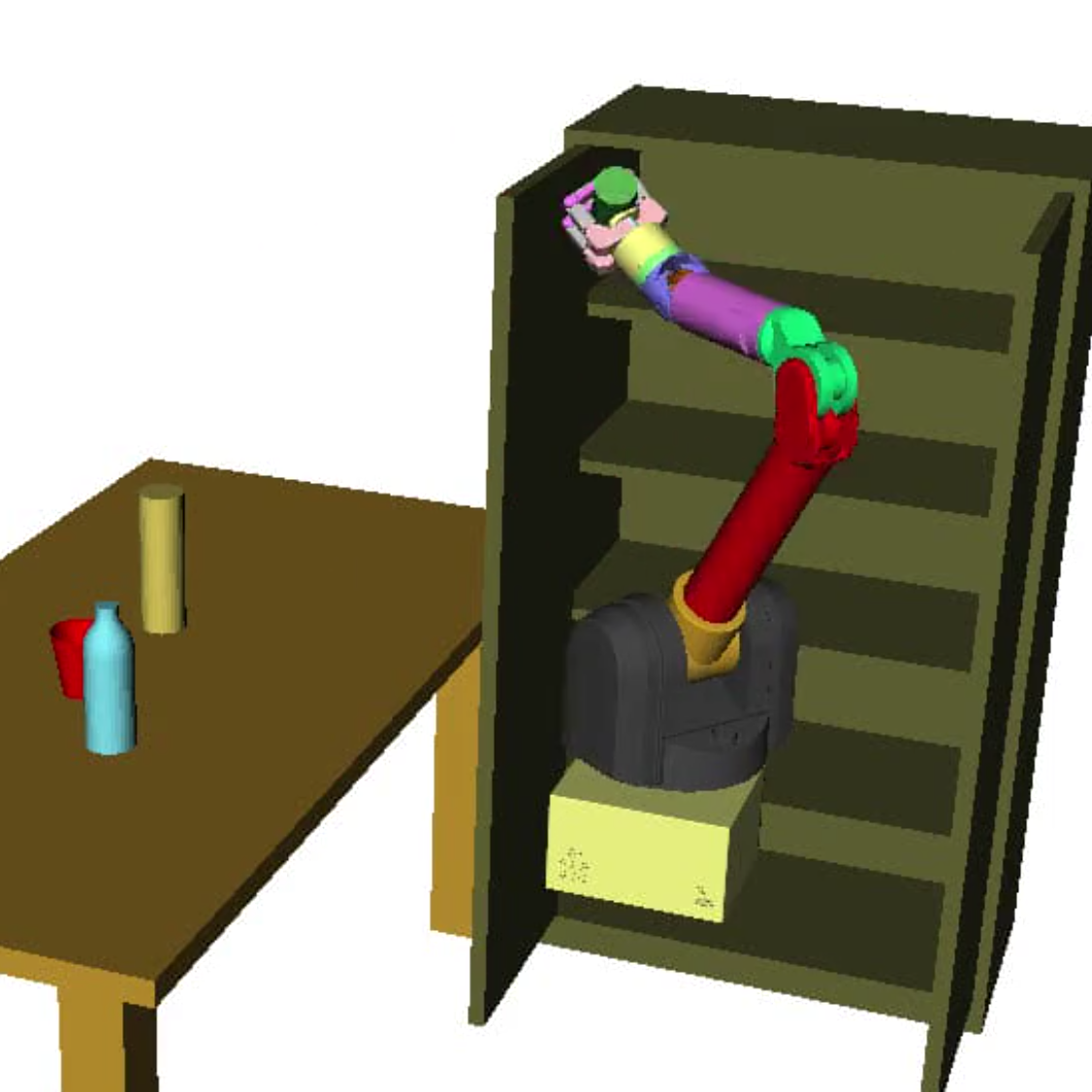}
\includegraphics[width=0.44\columnwidth]{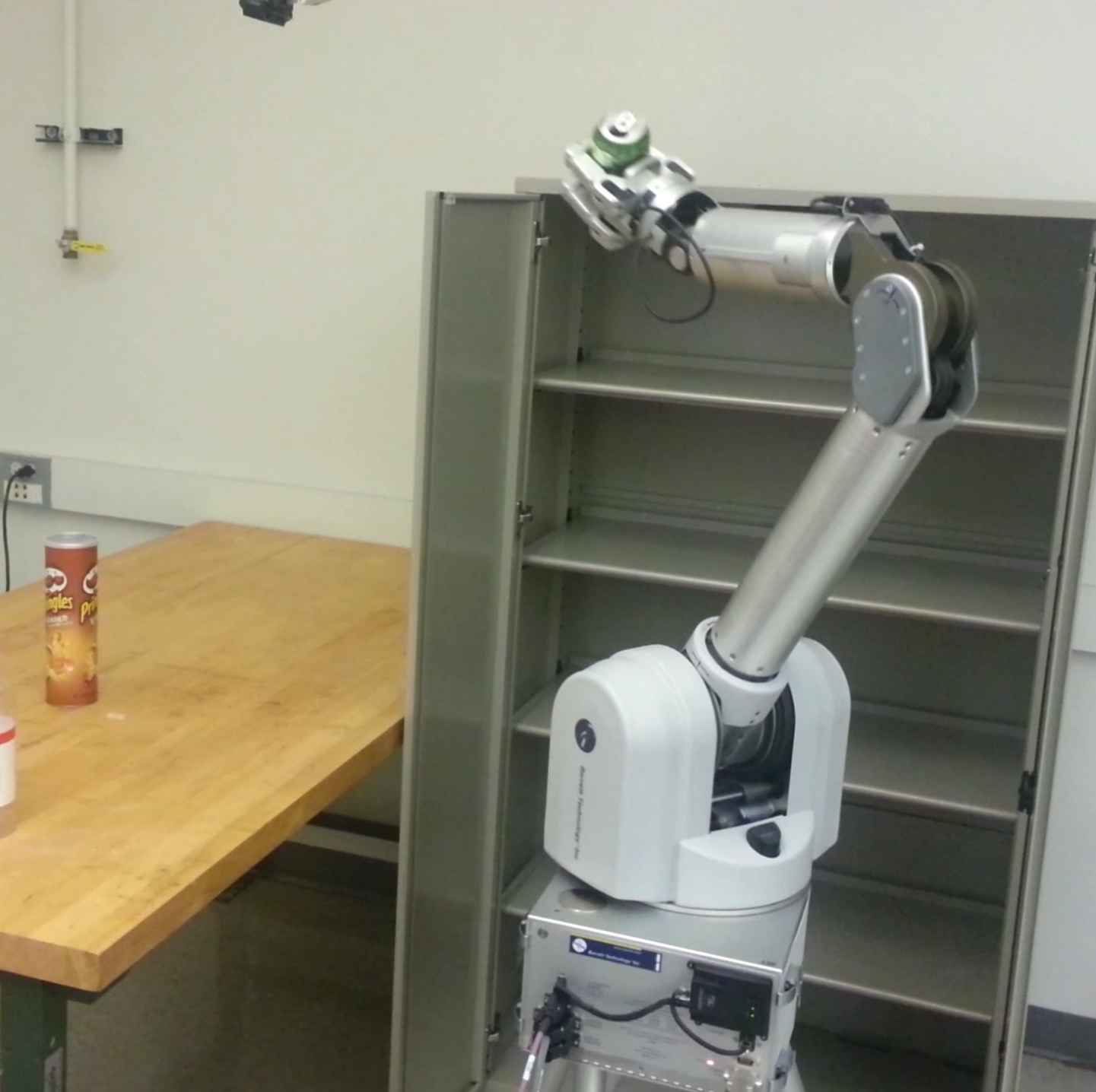}
\includegraphics[width=0.44\columnwidth]{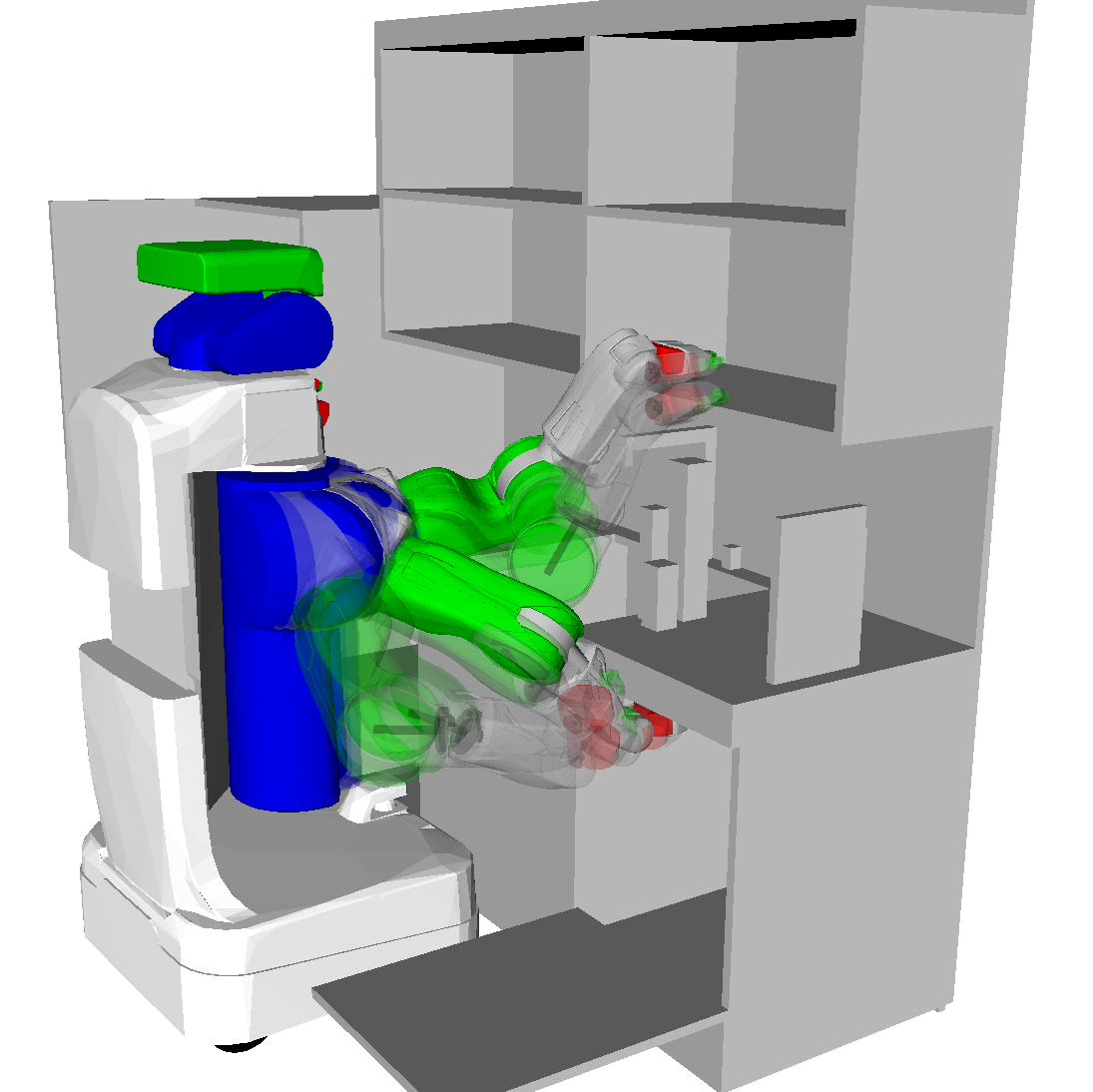}
\includegraphics[width=0.44\columnwidth]{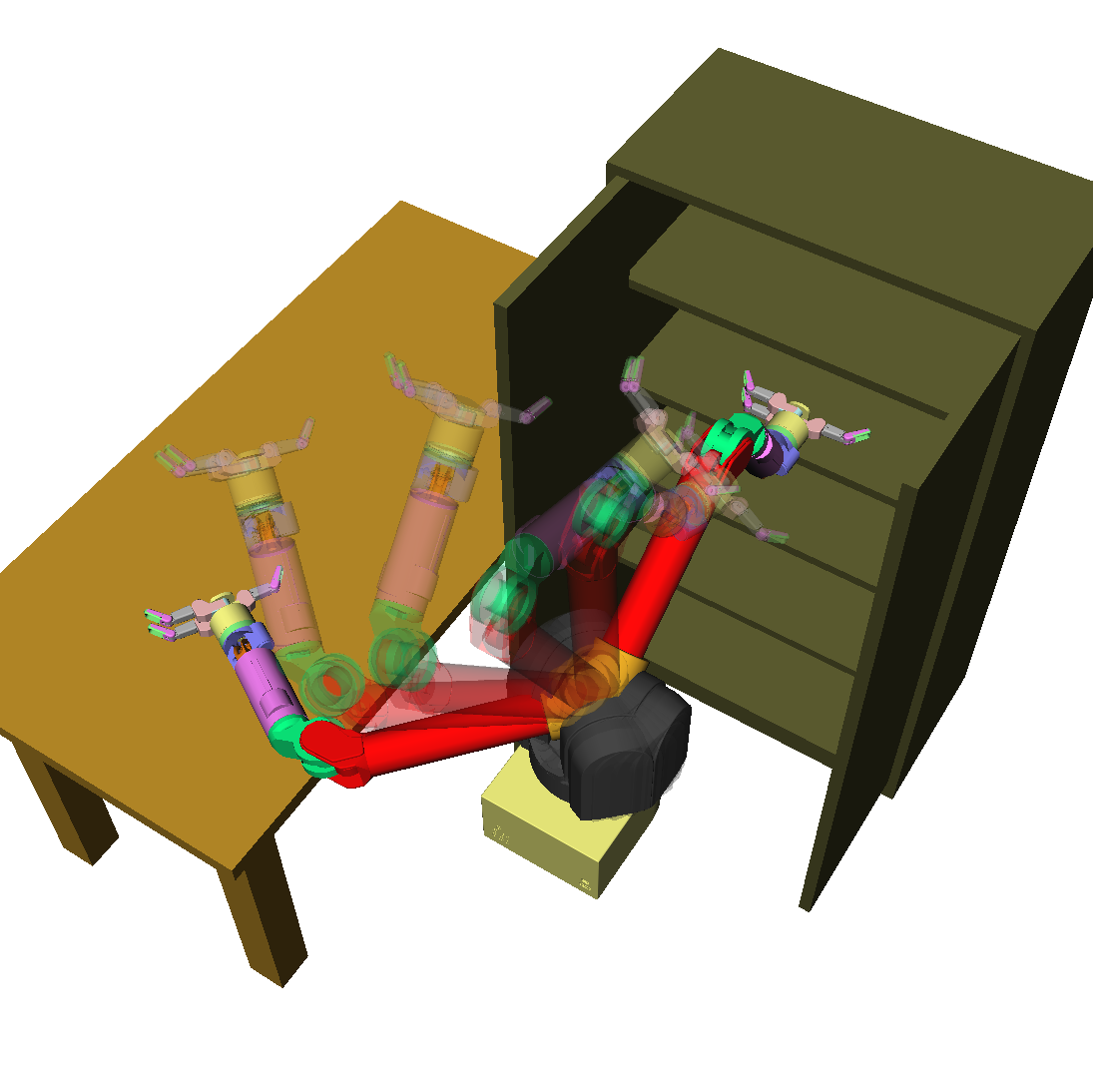}
\protect\caption{Optimized trajectory found by GPMP2 is used to place  a soda can on a shelf in simulation (top left) and with a real WAM arm (top right). Examples of successful trajectories generated by GPMP2 are shown in the countertop (bottom left) and lab (bottom right) environments with the PR2 and WAM robots respectively.
\label{fig:intro}}
\end{figure}

Another advantage of GPMP2 is that we can easily extend the algorithm using techniques designed for incremental inference on factor graphs developed in the context of SLAM. For example, Incremental Smoothing and Mapping (iSAM)~\citep{kaess2008isam,kaess2011isam2} can be adapted to efficiently solve replanning problems.

In this paper, we provide a revised and extended version of our previous work~\citep{Mukadam-ICRA-16,Dong-RSS-16}, give more theoretical insight, and provide a proof for the sparsity of the linear system in GPMP2. We also conduct additional benchmarks on larger datasets and compare GPMP and GPMP2 against leading trajectory optimization-based motion planning algorithms~\citep{zucker2013chomp,schulman2014motion} as well as sampling-based motion planning algorithms~\citep{kuffner2000rrt,csucan2009kinodynamic,sucan2012open} in multiple reaching tasks (Figure~\ref{fig:intro}). Our results show GPMP2 to be several times faster than the state-of-the-art with higher success rates. We also benchmark GPMP2 against our incremental planner, iGPMP2, on replanning tasks and show that iGPMP2 can incrementally solve replanning problems an order of magnitude faster than GPMP2 solving from scratch.

%%%%%%%%%%%%%%%%%%%%%%%%%%%%%%%%%%%%%%%%%%%%%%%%%%%%%%%%%%%%%%%%%%%%%%%%%%%%%%%%%%%%%%%%%%%%%%%%%%%%%%%%%%%%%%%%

\section{Related work}\label{sec:related_work}

Most motion planning algorithms are categorized as either sampling-based algorithms or trajectory optimization-based algorithms. Sampling-based planners such as probabilistic roadmaps (PRMs)~\citep{kavraki1996probabilistic} construct a dense graph from random samples in obstacle free areas of the robot's configuration space. PRMs can be used for multiple queries by finding the shortest path between a start and goal configuration in the graph. Rapidly exploring random trees (RRTs)~\citep{kuffner2000rrt} find trajectories by incrementally building space-filling trees through directed sampling. RRTs are very good at finding feasible solutions in highly constrained problems and high-dimensional search spaces. Both PRMs and RRTs offer probabilistic completeness, ensuring that, given enough time, a feasible trajectory can be found, if one exists. Despite guarantees, sampling-based algorithms may be difficult to use in real-time applications due to computational challenges. Often computation is wasted exploring regions that may not lead to a solution. Recent work in informed techniques~\citep{gammell2015batch} combat this problem by biasing the sampling approach to make search more tractable.

In contrast with sampling-based planners, trajectory optimization starts with an initial, possibly infeasible, trajectory and then optimizes the trajectory by minimizing a cost function. Covariant Hamiltonian Optimization for Motion Planning (CHOMP) and related methods~\citep{ratliff2009chomp,zucker2013chomp,he2013multigrid,byravan2014space,Marinho-RSS-16} optimize a cost functional using covariant gradient descent, while Stochastic Trajectory Optimization for Motion Planning (STOMP)~\citep{kalakrishnan2011stomp} optimizes non-differentiable costs by stochastic sampling of noisy trajectories. TrajOpt~\citep{schulman2013finding,schulman2014motion} solves a sequential quadratic program and performs convex continuous-time collision checking. In contrast to sampling-based planners, trajectory optimization methods are very fast, but only find locally optimal solution. The computational bottleneck results from evaluating costs on a fine discretization of the trajectory or, in difficult problems, repeatedly changing the initial conditions until a feasible trajectory is discovered. 

Continuous-time trajectory representations can overcome the computational cost incurred by finely discretizing the trajectory. Linear interpolation \citep{Bosse09icra,Li13icra,Dong2014fsr}, splines \citep{Bibby10icra,Anderson13icra,Furgale13iros,Leutenegger15ijrr,Patron15ijcv,Furgale15ijrr}, and hierarchical wavelets \citep{Anderson14icra} have been used  to represent trajectories in filtering and state estimation. B-Splines~\citep{Elbanhawi15itits} have similarly been used to represent trajectories in motion planning problems.
Compared to parametric representations like splines and wavelets, Gaussian processes provide a natural notion of uncertainty on top of allowing a sparse parameterization of the continuous-time trajectory. A critical distinction in motion planning problems is that even with a sparse parameterization, the collision cost has to be evaluated at a finer resolution. Therefore, if the interpolation procedure for a chosen continuous-time representation is computationally expensive, the resulting speedup obtained from a sparse representation is negligible and may result in an overall slower algorithm. Recent work by~\citet{Marinho-RSS-16} works to optimize trajectories in RKHS with RBF kernels, but ignores the cost between sparse waypoints. Even without interpolation, these dense kernels result in relatively computationally expensive updates. In this work, we use structured Gaussian processes (GPs) that allow us to exploit the underlying sparsity in the problem to perform efficient inference. We are able to use fast GP regression to interpolate the trajectory and evaluate obstacle cost on a finer resolution, while the trajectory can be parameterized by a small number of support states. We also show in this work that the probabilistic representation naturally allows us to represent the motion planning problem with a factor graph and the GP directly corresponds to the system dynamics or motion model thus giving it a physical meaning.

GPs have been used for function approximation in supervised learning~\citep{vijayakumar2005incremental,kersting2007most}, inverse dynamics modeling~\citep{nguyen2008learning,sturm2009body}, reinforcement learning~\citep{deisenroth2011pilco}, path prediction~\citep{tay2008modelling}, simultaneous localization and mapping~\citep{barfoot2014batch,Yan17ras}, state estimation~\citep{ko2009gp,tong2012gaussian}, and controls~\citep{theodorou2010stochastic}, but to our knowledge GPs have not been used in motion planning.

We also consider motion planning from the perspective of probabilistic inference. Early work by \cite{attias2003planning} uses inference  to solve Markov decision processes. More recently, solutions to planning and control problems have used probabilistic tools such as expectation propagation~\citep{toussaint2009robot}, expectation maximization~\citep{toussaint2006probabilistic,levine2013variational}, and KL-minimization~\citep{rawlik2012stochastic}. We exploit the duality between inference and optimization to perform inference on factor graphs by solving nonlinear least square problems. While this is an established and efficient approach~\citep{dellaert2006square} to solving large scale SLAM problems, we introduce this technique in the context of motion planning. Incremental inference can also be performed efficiently on factor graphs~\citep{kaess2008isam, kaess2011isam2}, a fact we take advantage of to solve replanning problems.

Replanning involves adapting a previously solved solution to changing conditions. Early replanning work like D\textsuperscript{*}~\citep{koenig2003performance} and Anytime A\textsuperscript{*}~\citep{likhachev2005anytime}  need a finely discretized state space and therefore do not scale well with high-dimensional problems. Recent trajectory optimization algorithms inspired from CHOMP~\citep{ratliff2009chomp} like incremental trajectory optimization for motion planning (ITOMP)~\citep{park2012itomp} can fluently replan using a scheduler that enforces timing restrictions but the solution cannot guarantee feasibility.  GPUs have been suggested as a way to increase the speed of replanning~\citep{park2013real}, with some success. Our algorithm is inspired from the incremental approach to SLAM problems~\citep{kaess2011isam2} that can efficiently update factor graphs to generate new solutions without performing redundant calculations. During planning, we use this method to update the trajectory only where necessary, thus reducing computational costs and making fast replanning possible.

%%%%%%%%%%%%%%%%%%%%%%%%%%%%%%%%%%%%%%%%%%%%%%%%%%%%%%%%%%%%%%%%%%%%%%%%%%%%%%%%%%%%%%%%%%%%%%%%%%%%%%%%%%%%%%%%

\section{Motion planning as trajectory optimization}\label{sec:problem}

The goal of motion planning via trajectory optimization is to find trajectories, $\bm \theta (t): t \to \mathbb{R}^D$, where $D$ is dimensionality of the state, that satisfy constraints and minimize cost~\citep{zucker2013chomp,kalakrishnan2011stomp,schulman2014motion}. Motion planning can therefore be formalized as
\begin{align} \label{eq:optimization}
	\text{minimize} & \quad \mathcal{F}[\bm \theta (t)] \\
	\text{subject to} & \quad \mathcal{G}_i[\bm{\theta} (t)] \leq 0, \ i = 1,\dots,m_{ineq} \nonumber \\
	& \quad \mathcal{H}_i[\bm{\theta} (t)] = 0, \ i = 1,\dots,m_{eq}\nonumber
\end{align}
where the trajectory $\bm{\theta} (t)$ is a continuous-time function, mapping time $t$ to robot states, which are generally configurations (and possibly higher-order derivatives). $\mathcal{F}[\bm{\theta}(t)]$ is an objective or cost functional that evaluates the quality of a trajectory and usually encodes \emph{smoothness} that minimizes higher-order derivatives of the robot states (for example, velocity or acceleration) and collision costs that enforces the trajectory to be \emph{collision-free}. $\mathcal{G}_i[\bm{\theta}(t)]$ are inequality constraint functionals such as joint angle limits, and $\mathcal{H}_i[\bm{\theta}(t)]$ are task-dependent equality constraints, such as the desired start and end configurations and velocities, or the desired end-effector orientation (for example, holding a cup filled with water upright). The number of inequality or equality constraints may be zero, depending on the specific problem. Based on the optimization technique used to solve Eq.~\eqref{eq:optimization}, collision cost may also appear as an obstacle avoidance inequality constraint~\citep{schulman2014motion}. In practice, most existing trajectory optimization algorithms work with a fine discretization of the trajectory, which can be used to reason about thin obstacles or tight navigation constraints, but can incur a large computational cost.

%%%%%%%%%%%%%%%%%%%%%%%%%%%%%%%%%%%%%%%%%%%%%%%%%%%%%%%%%%%%%%%%%%%%%%%%%%%%%%%%%%%%%%%%%%%%%%%%%%%%%%%%%%%%%%%%

\section{Gaussian processes for continuous-time trajectories}\label{sec:gp}

A vector-valued Gaussian process (GP)~\citep{rasmussen2006gaussian} provides a principled way to reason about continuous-time trajectories, where the trajectories are viewed as functions that map time to state. In this section, we describe how GPs can be used to encode a prior on trajectories  such that optimality properties like smoothness are naturally encouraged (Section~\ref{sec:gpprior}). We also consider a class of structured priors for trajectories that will be useful in efficient optimization (Section~\ref{sec:ltvsde}), and we provide details about how fast GP interpolation can be used to query the trajectory at any time of interest (Section~\ref{sec:gpintp}).

%------------------------------------------------------------------------------------------------------------------------------------------------%
\subsection{The GP prior}\label{sec:gpprior}

We consider continuous-time trajectories as samples from a vector-valued GP, $\bm{\theta}(t) \sim \mathcal{GP}(\bm{\mu}(t), \bm{\mathcal{K}}(t, t'))$, where $\bm{\mu}(t)$ is a vector-valued mean function and $\bm{\mathcal{K}}(t,t')$ is a matrix-valued covariance function. A vector-valued GP is a collection of random variables, any finite number of which have a joint Gaussian distribution. Using the GP framework, we can say that for any collection of times ${\bm t} = \{ t_0, \dots, t_N\}$, $\bm{\theta}$ has a joint Gaussian distribution:
\begin{equation}
	\bm{\theta} = \begin{bmatrix} \bm{\theta}_0 & \hdots & \bm{\theta}_N \end{bmatrix}^\top \sim \mathcal{N}(\bm{\mu}, \bm{\mathcal{K}})
\end{equation}
with the mean vector $\bm{\mu}$ and covariance kernel $\bm{\mathcal{K}}$ defined as
\begin{equation} 
	\bm{\mu} = \begin{bmatrix} \bm{\mu}_0 & \hdots & \bm{\mu}_N \end{bmatrix}^\top\hspace{-2mm}, \,
	\bm{\mathcal{K}} = [\bm{\mathcal{K}}(t_i, t_j)]\Bigr|_{ij, 0 \leq i,j \leq N}.
\end{equation}
We use bold $\bm{\theta}$ to denote the matrix formed by vectors $\bm{\theta}_i \in \mathbb{R}^D$, which are \emph{support states} that parameterize the continuous-time trajectory $\bm{\theta}(t)$. Simnilar notation is used for $\bm{\mu}$.

\begin{figure}[t]
	\centering
	\includegraphics[width=0.85\columnwidth]{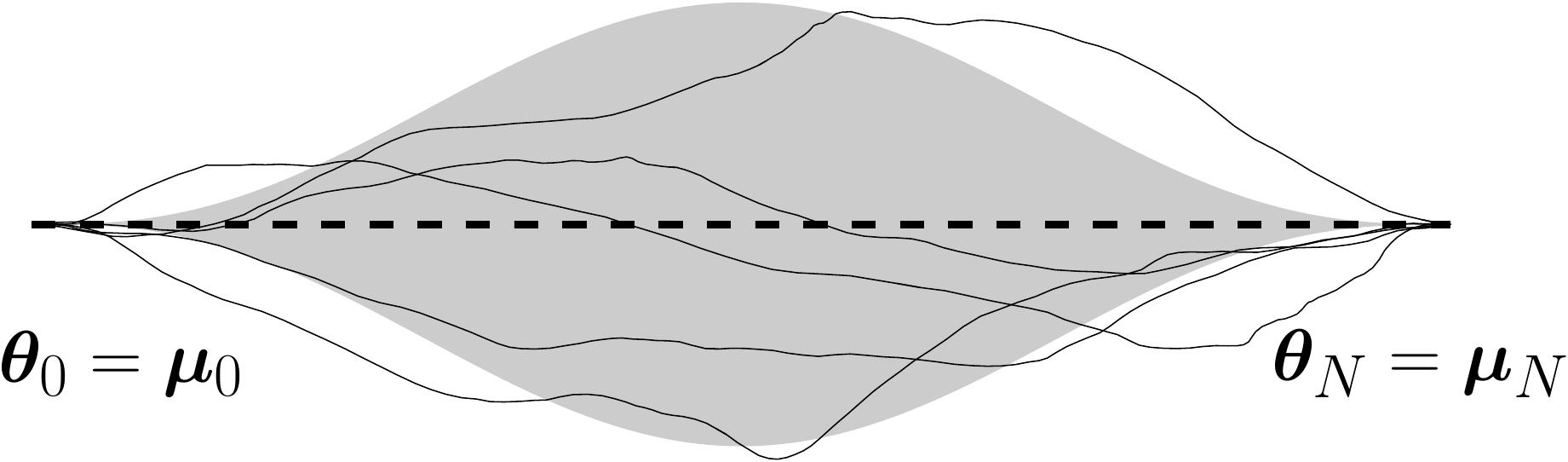}
	\caption{An example GP prior for trajectories. The dashed line is the mean trajectory $\bm{\mu}(t)$ and the shaded area indicates the covariance. The 5 solid lines are sample trajectories $\bm{\theta}(t)$ from the GP prior.}
	\label{fig:gptraj}
\end{figure}

The GP defines a prior on the space of trajectories :
\begin{align}
	p(\bm{\theta}) &\propto \exp \bigg\{ - \frac{1}{2} \parallel \bm{\theta} - \bm{\mu} \parallel^{2}_{\bm{\mathcal{K}}} \bigg\} \label{eq:prob_gp}
\end{align}
where $\parallel \bm{\theta} - \bm{\mu} \parallel^{2}_{\bm{\mathcal{K}}} \doteq (\bm{\theta} - \bm{\mu})^\top \bm{\mathcal{K}}^{-1} (\bm{\theta} - \bm{\mu})$ is the Mahalanobis distance. Figure~\ref{fig:gptraj} shows an example GP prior  for trajectories. Intuitively this prior encourages \emph{smoothness} encoded by the kernel $\bm{\mathcal{K}}$ and directly applies on the function space of trajectories. The negative log of this distribution serves as the prior cost functional in the objective (see Section~\ref{sec:costs}) and penalizes the deviation of the trajectory from the mean defined by the prior.

%------------------------------------------------------------------------------------------------------------------------------------------------%
\subsection{A Gauss-Markov model}\label{sec:ltvsde}

Similar to previous work~\citep{sarkka2013spatiotemporal,barfoot2014batch}, we use a structured kernel   generated by a linear time-varying stochastic differential equation (LTV-SDE)
\begin{equation}
	\dot{\bm{\bm{\theta}}}(t) = \mathbf{A}(t)\bm{\theta}(t) + \mathbf{u}(t) + \mathbf{F}(t)\mathbf{w}(t), \label{eq:LVT-SDE}
\end{equation}
where $\mathbf{u}(t)$ is the known system control input, $\mathbf{A}(t)$ and $\mathbf{F}(t)$ are time-varying matrices of the system, and $\mathbf{w}(t)$ is generated by a white noise process. The white noise process is itself a zero-mean GP
\begin{equation}
	\mathbf{w}(t) \sim \mathcal{GP}(\mathbf{0}, \mathbf{Q}_C\delta(t-t')). 
\end{equation}
$\mathbf{Q}_C$ is the power-spectral density matrix and $\delta(t-t')$ is the Dirac delta function. The solution to the initial value problem of this LTV-SDE is
\begin{equation}\label{eq:ltvsde-solution}
\begin{small}
	\bm{\theta}(t) = \bm\Phi(t,t_0) \bm{\theta}_0 + \int_{t_0}^{t} \bm\Phi(t,s)(\mathbf{u}(s) + \mathbf{F}(s)\mathbf{w}(s)) \diff s,
\end{small}
\end{equation}
where $\mathbf{\Phi}(t, s)$ is the state transition matrix, which transfers
state from time $s$ to time $t$. The mean and covariance functions of the GP defined by this LTV-SDE are calculated by taking the first and second moments respectively on Eq.~\eqref{eq:ltvsde-solution},
\begin{align}
	&\,\,\widetilde{\bm{\mu}}(t) = \mathbf{\Phi}(t,t_0) \bm{\mu}_0 + \int_{t_0}^{t} \mathbf{\Phi}(t,s) \mathbf{u}(s) \diff s, \\
	\widetilde{\bm{\mathcal{K}}}&(t,t') =  \mathbf{\Phi}(t,t_0) \bm{\mathcal{K}}_0 \mathbf{\Phi}(t',t_0)^\top\nonumber \\
	&+ \int_{t_0}^{\min(t,t')} \mathbf{\Phi}(t,s) \mathbf{F}(s) \mathbf{Q}_C \mathbf{F}(s)^\top \mathbf{\Phi}(t',s)^\top \diff s.
\end{align}
$\bm{\mu}_0$ and $\bm{\mathcal{K}}_0$ are the initial mean and covariance of the start state respectively.

The desired prior of trajectories between a given start state $\bm{\theta}_0$ and goal state $\bm{\theta}_N$  for a finite set of support states, as described in Section~\ref{sec:gpprior}, can be found by conditioning this GP with a fictitious observation on the goal state with mean $\bm{\mu}_N$ and covariance $\bm{\mathcal{K}}_N$. Specifically
\begin{small}
\begin{align}
	{\bm{\mu}}&= \widetilde{\bm{\mu}} + \widetilde{\bm{\mathcal{K}}}(t_N,\bm t)^\top (\widetilde{\bm{\mathcal{K}}}(t_N,t_N) + \bm{\mathcal{K}}_N )^{-1} (\bm{\theta}_N - {\bm{\mu}}_N) \label{eq:prior_mean}\\
	{\bm{\mathcal{K}}}& = \widetilde{\bm{\mathcal{K}}} - \widetilde{\bm{\mathcal{K}}}(t_N,\bm t)^\top (\widetilde{\bm{\mathcal{K}}}(t_N,t_N) + \bm{\mathcal{K}}_N )^{-1} \widetilde{\bm{\mathcal{K}}}(t_N,\bm t), \label{eq:prior_k}
\end{align}
\end{small}
where $\widetilde{\bm{\mathcal{K}}}(t_N,\bm t) = [ \begin{matrix} \widetilde{\bm{\mathcal{K}}}(t_N,t_0) & \dots & \widetilde{\bm{\mathcal{K}}}(t_N,t_N) \end{matrix} ]$ (see Appendix A for proof).

This particular construction of the prior leads to a Gauss-Markov model that generates a GP with an exactly sparse tridiagonal precision matrix (inverse kernel) that can be factored as:
\begin{equation} \label{eq:Kinv}
	\bm{\mathcal{K}}^{-1} = \mathbf B^\top \mathbf Q^{-1} \mathbf B
\end{equation}
with,
\begin{equation}\label{eq:Ainv}
\small
	\mathbf B = \left[ \begin{matrix}
	\mathbf{I} & \bm 0 & \dots & \bm 0 & \bm 0\\
	-\mathbf\Phi(t_1,t_0) & \mathbf{I} & \dots & \bm 0 & \bm 0\\
	\bm 0 & -\mathbf\Phi(t_2,t_1) & \ddots & \vdots & \vdots \\
	\vdots & \vdots & \ddots & \mathbf{I} & \bm 0 \\
	\bm 0 & \bm 0 & \dots &-\mathbf\Phi(t_N,t_{N-1}) & \mathbf{I} \\
	\bm 0 & \bm 0 & \dots & \bm 0 & \mathbf{I} \end{matrix} \right],
\end{equation}
which has a band diagonal structure and $\mathbf Q^{-1}$ is block diagonal such that
\begin{align}
	\mathbf Q^{-1} &= \operatorname{diag}(\bm{\mathcal{K}}_0^{-1}, \mathbf Q_{0,1}^{-1}, \dots , \mathbf Q_{N-1,N}^{-1}, \bm{\mathcal{K}}_N^{-1}), \label{eq:Q}\\
	\mathbf Q_{a,b} &= \int_{t_a}^{t_b} \mathbf \Phi(b,s) \mathbf F(s) \mathbf{Q}_c \mathbf F(s)^\top \mathbf \Phi(b,s)^\top \diff s \label{eq:Qab}
\end{align}
(see Appendix A for proof). This sparse structure is useful for fast GP interpolation (Section \ref{sec:gpintp}) and efficient optimization (Section~\ref{sec:gpmp} and~\ref{sec:pinf}).

An interesting observation here is that this choice of kernel can be viewed as a generalization of CHOMP~\citep{zucker2013chomp}. For instance, if the identity and zero blocks in the precision matrix are scalars, the state transition matrix $\mathbf{\Phi}$ is a unit scalar, and $\mathbf Q^{-1}$ is an identity matrix, $\bm{\mathcal{K}}^{-1}$ reduces to the matrix $A$ formed by finite differencing in CHOMP. In this context, it means that CHOMP considers a trajectory of positions in configuration space, that is generated by a deterministic differential equation (since $\mathbf Q^{-1}$ is identity).

The linear model in Eq.~\eqref{eq:LVT-SDE} is sufficient to model kinematics for the robot manipulators considered in the scope of this work, however our framework can be extended to consider non-linear models following~\cite{anderson2015batch}.

\begin{figure}[t]
\begin{centering}
{\includegraphics[width=0.95\columnwidth]{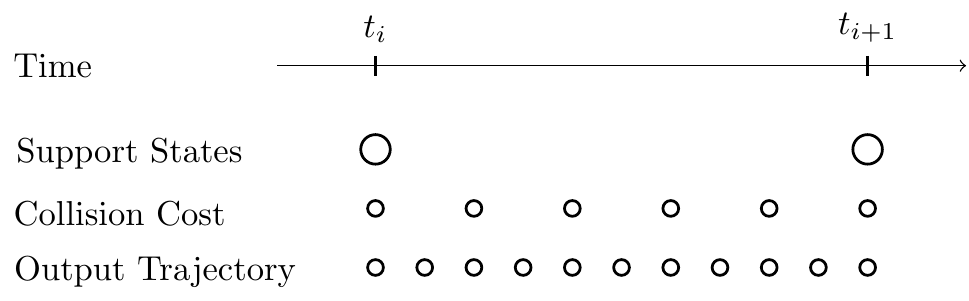}}
\par\end{centering}
\protect\caption{An example that shows the trajectory at different resolutions. Support states parameterize the trajectory, collision cost checking is performed at a higher resolution during optimization and the output trajectory can be up-sampled further for execution. 
\label{fig:inter_time_stamps}}
\end{figure}

\begin{figure*}[!t]
	\begin{centering}
		\begin{subfigure}[b]{0.32\textwidth}
			\centering
			\includegraphics[width=1\linewidth]{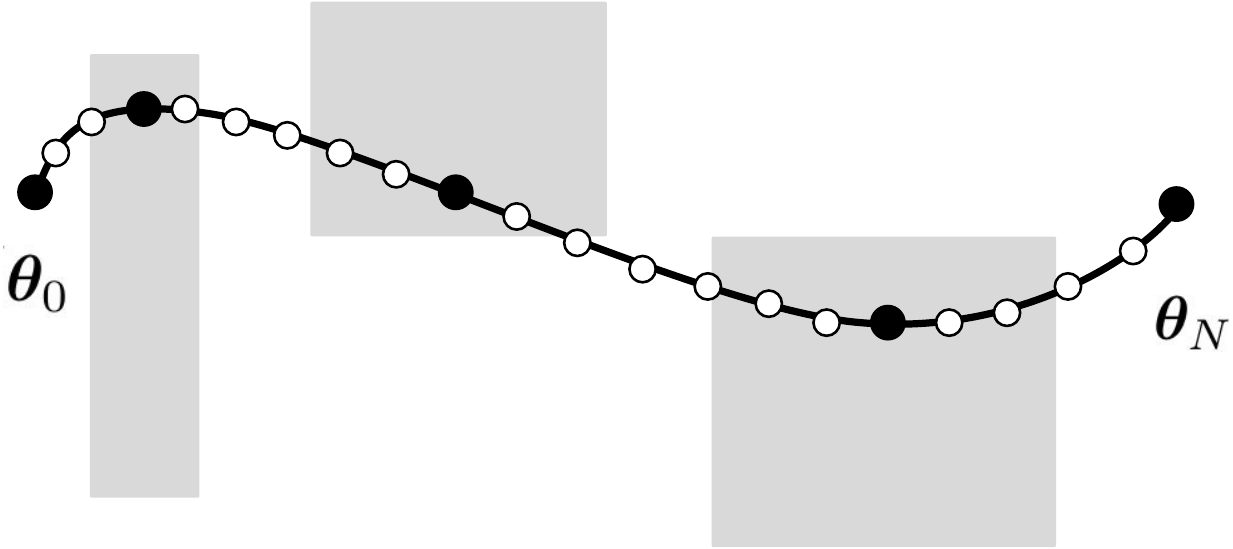}
			\caption{Interpolated trajectory}
		\end{subfigure}
		\hfill
		\begin{subfigure}[b]{0.32\textwidth}
			\centering
			\includegraphics[width=1\linewidth]{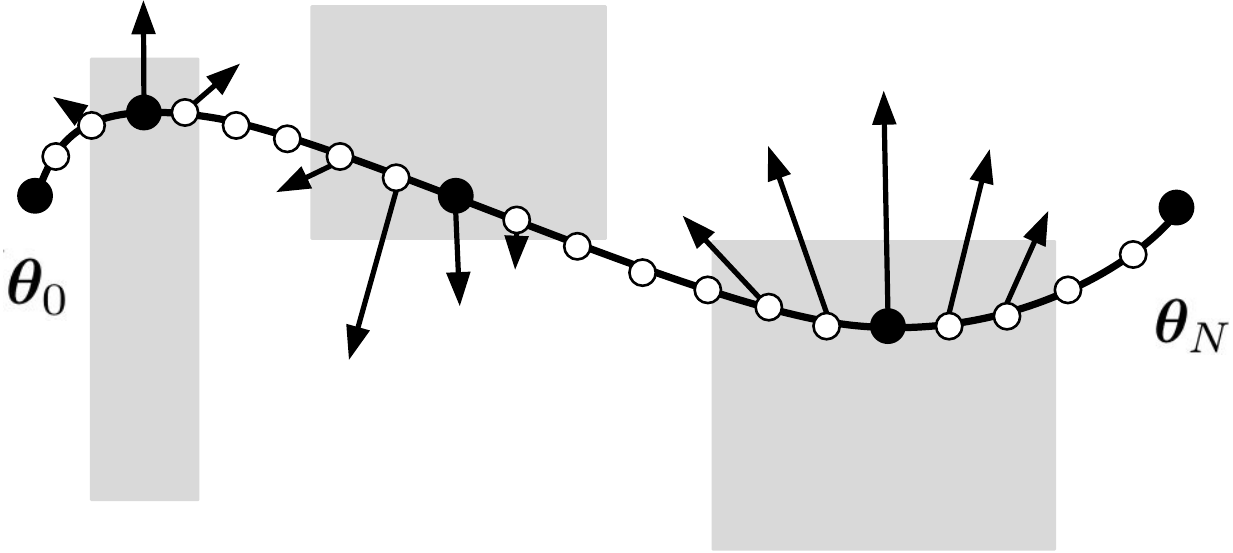}
			\caption{Gradient on interpolated trajectory}
		\end{subfigure}	
		\hfill
		\begin{subfigure}[b]{0.32\textwidth}
			\centering
			\includegraphics[width=1\linewidth]{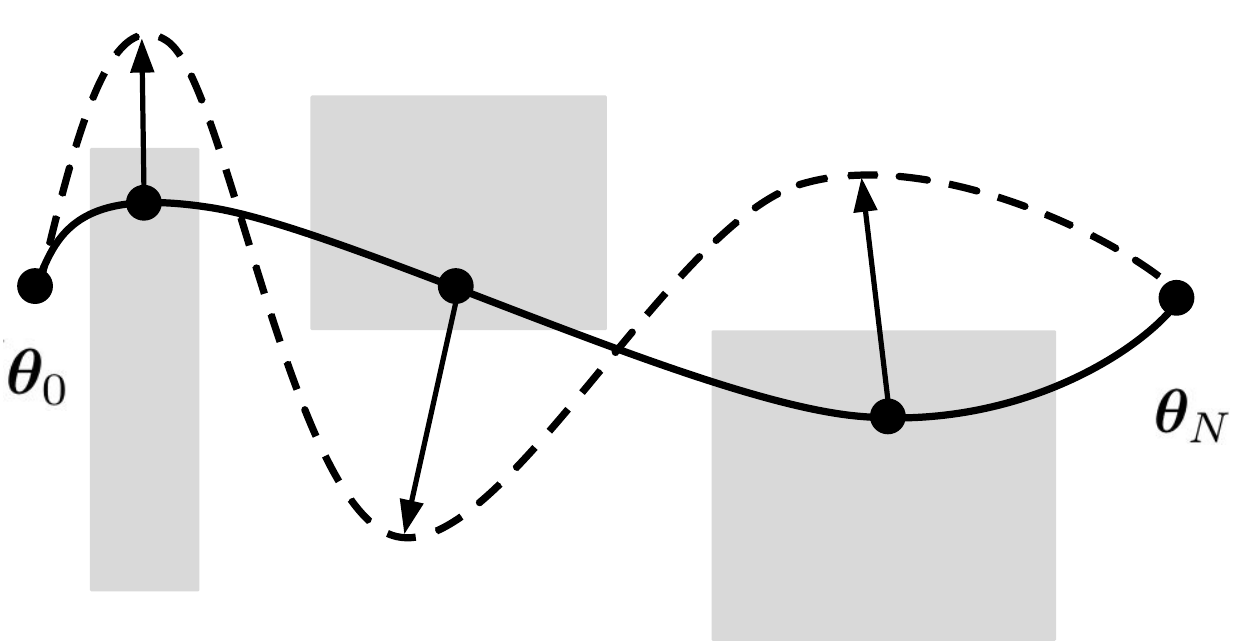}
			\caption{Gradient on support states}
		\end{subfigure}
	\end{centering}
	\caption{An example showing how GP interpolation is used during optimization. (a) shows the current iteration of the trajectory (black curve) parameterized by a sparse set of support states (black circles). GP regression is used to densely up-sample the trajectory with interpolated states (white circles). Then, in (b) cost is evaluated on all states and their gradients are illustrated by the arrows. Finally, in (c) the cost and gradient information is propagated to just the support states illustrated by the larger arrows such that only the support states are updated that parameterize the new trajectory (dotted black curve).}
	\label{fig:gpintp}
\end{figure*}

%------------------------------------------------------------------------------------------------------------------------------------------------%
\subsection{Gaussian process interpolation}\label{sec:gpintp}

One of the benefits of Gaussian processes is that they can be parameterized by only a sparse set of \emph{support states}, but the trajectory can be queried at \emph{any} time of interest through Gaussian process interpolation. The reduced parameterization makes each iteration of trajectory optimization efficient. Given the choice of the structured prior from the previous subsection, rich collision costs between the support states can be evaluated by performing dense GP interpolation between the support states quickly and efficiently. This cost can then be used to update the support states in a meaningful manner, reducing the computational effort. A much denser resolution of interpolation (Figure~\ref{fig:inter_time_stamps}) can also be useful in practice to feed the trajectory to a controller on a real robot.

The process of updating a trajectory with GP interpolation is explained through an example illustrated in Figure~\ref{fig:gpintp}. At each iteration of optimization, the trajectory with a sparse set of support states can be densely interpolated with a large number of states, and the collision cost can be evaluated on all the states (both support and interpolated). Next, collision costs at the interpolated states are propagated and accumulated to the nearby support states (the exact process to do this is explained in Section~\ref{sec:fast_updaterule} and~\ref{sec:fg}). Finally, the trajectory is updated by only updating the support states given the accumulated cost information.

Following~\citet{sarkka2013spatiotemporal,barfoot2014batch,Yan17ras}, we show how to exploit the structured prior to perform fast GP interpolation. The posterior mean of the trajectory at any time $\tau$ can be found in terms of the current trajectory $\bm{\theta}$ at time points $\bm t$~\citep{rasmussen2006gaussian} by conditioning on the support states that parameterize trajectory:
\begin{align}\label{eq:interp}
	\bm{\theta}(\tau) = \widetilde{\bm \mu}(\tau) + \widetilde{\bm{\mathcal{K}}}(\tau, {\bm t})\widetilde{\bm{\mathcal{K}}}^{-1} (\bm \theta - \widetilde{\bm \mu})
\end{align}
i.e. performing Gaussian process regression. Although the interpolation in Eq.~\eqref{eq:interp} na\"{\i}vely requires $O(N)$ operations, $\bm{\theta}(\tau)$ can be computed in $O(1)$ by leveraging the structure of the 
sparse GP prior generated by the Gauss-Markov model introduced in Section~\ref{sec:gp}. This implies that $\bm{\theta}(\tau)$ at $\tau, t_i < \tau < t_{i+1}$ can be expressed as a linear combination of \emph{only} the adjacent function values $\bm{\theta}_{i}$ and $\bm{\theta}_{i+1}$ and is efficiently computed as
\begin{equation}\label{eq:interpolation}
\small
	\bm{\theta}(\tau) = \widetilde{\bm{\mu}}(\tau) + \mathbf{\Lambda}(\tau) ({\bm{\theta}}_{i} - \widetilde{\bm{\mu}}_{i}) + \mathbf{\Psi}(\tau) ({\bm{\theta}}_{i+1} - \widetilde{\bm{\mu}}_{i+1})
\end{equation}
where
\begin{align*}
	\mathbf{\Lambda}(\tau) &= \mathbf{\Phi}(\tau, t_{i}) - \mathbf{\Psi}(\tau) \mathbf{\Phi}(t_{i+1}, t_{i}) \\
	\mathbf{\Psi}(\tau) &= \mathbf{Q}_{i,\tau} \mathbf{\Phi}(t_{i+1},\tau)^\top \mathbf{Q}_{i,i+1}^{-1}
\end{align*}
is derived by substituting
\begin{equation*}
	\widetilde{\bm{\mathcal{K}}}(\tau) \widetilde{\bm{\mathcal{K}}}^{-1} = [\ \bm 0 \; \dots \; \bm 0 \; \bm\Lambda(\tau) \; \bm\Psi(\tau) \; \bm 0 \; \dots \; \bm 0 \ ]
\end{equation*}
in Eq.~\eqref{eq:interp} with only the $(i)^{th}$ and $(i+1)^{th}$ block columns being non-zero.

This provides an elegant way to do fast GP interpolation on the trajectory that exploits the structure of the problem. In Section~\ref{sec:fast_updaterule} and~\ref{sec:fg} we show how this is utilized to perform efficient optimization.

%%%%%%%%%%%%%%%%%%%%%%%%%%%%%%%%%%%%%%%%%%%%%%%%%%%%%%%%%%%%%%%%%%%%%%%%%%%%%%%%%%%%%%%%%%%%%%%%%%%%%%%%%%%%%%%%

\section{Gaussian process motion planning}\label{sec:gpmp}

We now describe the Gaussian Process Motion Planner (\textbf{GPMP}),  which combines the Gaussian process representation with a gradient descent-based optimization algorithm for motion planning.

%------------------------------------------------------------------------------------------------------------------------------------------------%
\subsection{Cost functionals}\label{sec:costs}
Following the problem definition in Eq.~\eqref{eq:optimization} we design the objective functional as
\begin{equation} \label{eq:chomp_cost}
	\mathcal{F}[\bm{\theta}(t)] = \mathcal{F}_{obs}[\bm{\theta}(t)] + \lambda \mathcal{F}_{gp}[\bm{\theta}(t)]
\end{equation}
where $\mathcal{F}_{gp}$ is the GP prior cost functional (the negative natural logarithm of prior distribution) from Eq.~\eqref{eq:prob_gp}
\begin{equation} \label{eq:gp_cost}
	\mathcal{F}_{gp}[\bm{\theta}(t)] = \mathcal{F}_{gp}[\bm{\theta}] = \frac{1}{2}\| \bm{\theta} - \bm\mu \|^2_{\bm{\mathcal{K}}}
\end{equation}
penalizing the deviation of the parameterized trajectory from the prior mean, $\mathcal{F}_{obs}$ is the obstacle cost functional that penalizes collision with obstacles and $\lambda$ is the trade-off between the two functionals.

As discussed in Section~\ref{sec:ltvsde} the GP smoothness prior can be considered a generalization to the one used in practical applications of CHOMP constructed through finite dynamics. In contrast to CHOMP, we also consider our trajectory to be augmented by velocities and acceleration. This allows us to keep the state Markovian in the prior model (Section~\ref{sec:ltvsde}), is useful in computation of the obstacle cost gradient (Section~\ref{sec:gpmp_opt}), and also allows us to stretch or squeeze the trajectory in space while keeping the states on the trajectory temporally equidistant~\citep{byravan2014space}.

The obstacle cost functional $\mathcal{F}_{obs}$  is also similar to the one used in CHOMP~\citep{zucker2013chomp}. This functional computes the arc-length parameterized line integral of the workspace obstacle cost of each body point as it passes through the workspace, and integrates over all body points:
\begin{equation}\label{eq:obs_cont}
	\mathcal{F}_{obs}[\bm{\theta}(t)] = \int_{t_0}^{t_{N}} \int_\mathcal{B} c(x) \| \dot x \| \diff u \diff t
\end{equation}
where $c(\cdot):\mathbb{R}^3 \to \mathbb{R}$ is the workspace cost function that penalizes the set of points $\mathcal{B} \subset \mathbb{R}^3$ on the robot body when they are in or around an obstacle, and $x$ is the forward kinematics that maps robot configuration to workspace (see~\cite{zucker2013chomp} for details).

In practice, the cost functional can be approximately evaluated on the discrete support state parameterization of the trajectory i.e. $\mathcal{F}_{obs}[\bm{\theta}(t)] = \mathcal{F}_{obs}[\bm{\theta}]$, the obstacle cost is calculated using a precomputed signed distance field (see Section~\ref{sec:gpmp_imp}), and the inner integral is replaced with a summation over a finite number of body points that well approximate the robot's physical body.

%------------------------------------------------------------------------------------------------------------------------------------------------%
\subsection{Optimization}\label{sec:gpmp_opt}

We adopt an iterative, gradient-based approach to minimize the non-convex objective functional in Eq.~\eqref{eq:chomp_cost}. In each iteration, we form an approximation to the cost functional via a Taylor series expansion around the current parameterized trajectory $\bm{\theta}$:
\begin{equation} \label{eq:taylor}
	\mathcal{F}[\bm{\theta} + \delta \bm{\theta}] \approx \mathcal{F}[\bm{\theta}]+ \bar\nabla\mathcal{F}[\bm{\theta}] \delta \bm{\theta}
\end{equation}
We next minimize the approximate cost while constraining the trajectory to be close to the previous one. Then the optimal perturbation $\delta\bm{\theta}^*$ to the trajectory is:
\begin{align} \label{eq:optimal_pertubation}
	\delta\bm{\theta}^* = \argmin_{\delta\bm{\theta}} \Big\{ \mathcal{F}[\bm{\theta}] + \bar\nabla\mathcal{F}[\bm{\theta}] \delta \bm{\theta} + \frac{\eta}{2} \| \delta \bm{\theta} \|^2_{\bm{\mathcal{K}}} \Big\}
\end{align}
where $\eta$ is the regularization constant. Differentiating the right-hand side and setting the result to zero we obtain the update rule for each iteration:
\begin{gather}
	\bar\nabla\mathcal{F}[\bm{\theta}] + \eta\bm{\mathcal{K}}^{-1}\delta\bm{\theta}^* = 0 \quad \implies \quad \delta\bm{\theta}^* = - \frac{1}{\eta}\bm{\mathcal{K}}\bar\nabla \mathcal{F}[\bm{\theta}] \nonumber \\
	\bm{\theta} \leftarrow \bm{\theta} + \delta \bm{\theta}^* = \bm{\theta} - \frac{1}{\eta}\bm{\mathcal{K}}\bar\nabla \mathcal{F}[\bm{\theta}] \label{eq:update}
\end{gather}
To compute the update rule we need to find the gradient of the cost functional at the current trajectory
\begin{equation} \label{eq:cost_grad}
	\bar\nabla \mathcal{F}[\bm{\theta}] =  \bar\nabla\mathcal{F}_{obs}[\bm{\theta}] + \lambda \bar\nabla\mathcal F_{gp}[\bm{\theta}],
\end{equation}
which requires computing the gradients of the GP and obstacle cost functional. The gradient of the GP prior cost can be computed by taking the derivative of Eq.~\eqref{eq:gp_cost} with respect to the current trajectory 
\begin{align}\label{eq:gp_grad}
	\mathcal{F}_{gp}[\bm{\theta}] &=\frac{1}{2} (\bm{\theta} - \bm\mu)^\top \bm{\mathcal{K}}^{-1}(\bm{\theta} - \bm\mu) \nonumber \\
	\bar\nabla\mathcal F_{gp}[\bm{\theta}] &= \bm{\mathcal{K}}^{-1} (\bm{\theta} - \bm\mu)
\end{align}
The gradient of the obstacle cost functional can be computed from the Euler-Lagrange equation~\citep{courant1966methods} in which a functional of the form $\mathcal{F}[\bm{\theta}(t)] = \int v(\bm{\theta}(t)) \diff t$ yields a gradient
\begin{equation}\label{eq:ele}
	\bar\nabla \mathcal{F}[\bm{\theta}(t)] = \frac{\partial v}{\partial \bm{\theta}(t)} - \frac{\diff}{\diff t} \frac{\partial v}{\partial \bm{\dot \theta}(t)}
\end{equation}
Applying Eq.~\eqref{eq:ele} to find the gradient of Eq.~\eqref{eq:obs_cont} in the workspace and then mapping it back to the configuration space via the kinematic Jacobian $J$, and following the proof by \cite{quinlan1994real}, we compute the gradient with respect to configuration position, velocity, and acceleration at any time point $t_i$ as
\begin{equation}\label{eq:obs_grad}
\small
	\bar\nabla \mathcal{F}_{obs}[\bm{\theta}_i] =
	\left[ \begin{array}{c}
	\vspace{0.1cm} \int_\mathcal{B} J^\top ||\dot x|| \big[ (I - \hat{\dot x} \hat{\dot x}^\top) \nabla c - c\kappa \big] \diff u \\
	\vspace{0.1cm} \int_\mathcal{B} J^\top c \; \hat{\dot x} \diff u \\
	0
	\end{array} \right]\hspace{-.85mm}
\end{equation}
where $\kappa = ||\dot x||^{-2} (I - \hat{\dot x} \hat{\dot x}^\top) \ddot x$ is the curvature vector along the workspace trajectory traced by a body point, $\dot x$, $\ddot x$ are the velocity and acceleration respectively, of that body point determined by forward kinematics and the Hessian, and $\hat{\dot x} = \dot x/||\dot x||$ is the normalized velocity vector. Due to the augmented state, the velocity and acceleration can be obtained through the Jacobian and Hessian directly from the state. This is in contrast to CHOMP, which approximates the velocity and acceleration through finite differencing. The gradients at each time point are stacked together into a single vector $\mathbf{g} = \bar\nabla \mathcal{F}_{obs}[\bm{\theta}]$. We plug the cost gradients back into the update rule in Eq.~\eqref{eq:update} to get the update
\begin{equation}\label{eq:augchomp_update}
	\bm{\theta} \gets \bm{\theta} - \frac{1}{\eta} \bm{\mathcal{K}} \bigg( \lambda \bm{\mathcal{K}}^{-1}(\bm{\theta} - \bm \mu) + \mathbf{g} \bigg)
\end{equation}
This update rule can be interpreted as a generalization of the update rule for CHOMP with an augmented trajectory and a generalized prior. 

%------------------------------------------------------------------------------------------------------------------------------------------------%
\subsection{Compact trajectory representations and faster updates via GP interpolation}\label{sec:fast_updaterule}

In this section, we show that the finite number of states used to parameterize smooth trajectories can be very sparse in practice. Through GP interpolation, we can up-sample the trajectory to any desired resolution, calculate costs and gradients at this resolution, and then project the gradients back  to just the sparse set of support states. To interpolate $n_{ip}$ states between two support states at $t_i$ and $t_{i+1}$, we define two aggregated matrices using Eq.~\eqref{eq:interpolation},
\begin{align*}
	\mathbf\Lambda_{i} &= \left[ \begin{array}{ccccc}
	\mathbf\Lambda_{i,1}^\top & \hdots& \mathbf\Lambda_{i,j}^\top & \hdots & \mathbf\Lambda_{i,n_{ip}}^\top 
	\end{array} \right]^\top \\
	\mathbf\Psi_{i} &= \left[ \begin{array}{ccccc}
	\mathbf\Psi_{i,1}^\top & \hdots& \mathbf\Psi_{i,j}^\top & \hdots & \mathbf\Psi_{i,n_{ip}}^\top 
	\end{array} \right]^\top
\end{align*}
If we want to up-sample a sparse trajectory $\bm{\theta}$ by interpolating $n_{ip}$ states between every support state, we can quickly compute the new trajectory $\bm{\theta}_{up}$ as
\begin{equation}\label{eq:upsample}
	\bm{\theta}_{up} = \mathbf{M} (\bm{\theta}- \bm \mu) + \bm \mu_{up}
\end{equation}
where $\bm{\mu}_{up}$ corresponds to the prior mean with respect to the up sampled trajectory, and
\begin{equation}
	\hspace{-2mm}\scriptsize{
	\mathbf M = \left[ \begin{array}{cccccccc}\hspace{-1mm}
	\mathbf I & \mathbf 0 & \mathbf 0 & \dots & \dots & \dots & \mathbf 0 & \mathbf 0	\\
	\mathbf \Lambda_0 & \mathbf \Psi_0 & \mathbf 0 & \dots & \dots & \dots & \mathbf 0 & \mathbf 0\\
	\mathbf 0 & \mathbf I & \mathbf 0 & \dots & \dots & \dots & \mathbf 0 & \mathbf 0\\
	\mathbf 0 & \mathbf \Lambda_1 & \mathbf \Psi_1 & \dots & \dots & \dots & \mathbf 0 & \mathbf 0\\
	\vdots & \vdots & \ddots &&&& \vdots& \vdots \\
	\mathbf 0 & \mathbf 0 & \dots & \mathbf I & \mathbf 0 & \dots & \mathbf 0 & \mathbf 0\\
	\mathbf 0 & \mathbf 0 & \dots & \mathbf \Lambda_i & \mathbf \Psi_i & \dots & \mathbf 0 & \mathbf 0\\
	\mathbf 0 & \mathbf 0 & \dots &  \mathbf 0 & \mathbf I& \dots & \mathbf 0 & \mathbf 0\\
	\vdots &\vdots  &&  & & \ddots& \vdots& \vdots \\
	\mathbf 0 & \mathbf 0 & \dots &  \dots & \dots& \dots & \mathbf \Lambda_{N-1} & \mathbf \Psi_{N-1} \\
	\mathbf 0 & \mathbf 0 & \dots &  \dots & \dots& \dots & \mathbf 0 & \mathbf I \\
	\end{array}\hspace{-1mm} \right]}\hspace{-2mm}
\end{equation}
is a tall matrix that up-samples a sparse trajectory $\bm{\theta}$ with only $N+1$ support states to trajectory $\bm{\theta}_{up}$ with $(N+1) + N \times n_{ip}$ states.
The fast, high-temporal-resolution interpolation is also useful in practice if we want to feed the planned trajectory into a controller.

The efficient update rule is defined analogous to Eq.~\eqref{eq:augchomp_update} except on a sparse parametrization of the trajectory
\begin{align}\label{eq:gpmp_update}
 	\bm{\theta} &\gets \bm{\theta} - \frac{1}{\eta} \bm{\mathcal{K}} \bigg( \lambda \bm{\mathcal{K}}^{-1}(\bm{\theta} - \bm\mu) + \mathbf M^\top \mathbf g_{up} \bigg)
\end{align}
where the obstacle gradient over the sparse trajectory is found by chain rule using Eq.~\eqref{eq:upsample} and the obstacle gradient, $\mathbf g_{up}$ over the up-sampled trajectory. In other words, the above equation calculates the obstacle gradient for \emph{all} states (interpolated and support) and then projects them back onto just the support states using $\mathbf M^\top$. Cost information between support states is still utilized to perform the optimization, however only a sparse parameterization is necessary making the remainder of the update more efficient.

GPMP demonstrates how a continuous-time representation of the trajectory using GPs can generalize CHOMP and improve performance through sparse parameterization. However, the gradient-based optimization scheme has two drawbacks: first, convergence is slow due to the large number of iterations required to get a feasible solution; and, second, the gradients can be costly to calculate (See Figure~\ref{fig:timing}). We improve upon GPMP and address these concerns in the next section.

%%%%%%%%%%%%%%%%%%%%%%%%%%%%%%%%%%%%%%%%%%%%%%%%%%%%%%%%%%%%%%%%%%%%%%%%%%%%%%%%%%%%%%%%%%%%%%%%%%%%%%%%%%%%%%%%

\section{Motion planning as probabilistic inference}\label{sec:pinf}

To fully evoke the power of GPs, we view motion planning as probabilistic inference. A similar view has been explored before by Toussaint et al.~\citep{toussaint2009robot,toussaint2010bayesian}. Unlike this previous work, which uses message passing to perform inference,  we exploit the duality between inference and optimization and borrow ideas from the SLAM community for a more efficient approach. In particular, we use tools from the Smoothing and Mapping (SAM) framework~\citep{dellaert2006square} that performs inference on factor graphs by solving a nonlinear least squares problem~\citep{kschischang2001factor}. This approach exploits the sparsity of the underlying problem to obtain quadratic convergence.

The probabilistic inference view of motion planning provides several advantages:
\begin{enumerate}
\item The duality between inference and least squares optimization allows us to perform inference very efficiently, so motion planning is extremely \emph{fast}.
\item Inference tools from other areas of robotics, like the incremental algorithms based on the Bayes tree data structure~\citep{kaess2011isam2}, can be exploited and used in the context of planning. These tools can help speed up \emph{replanning}.
\item Inference can provide a deeper understanding of the connections between different areas of robotics, such as planning and control~\citep{Mukadam-ICRA-17}, estimation and planning~\citep{Mukadam-RSS-17,Mukadam-AURO-18}, and learning from demonstration and planning~\citep{pmlr-v78-rana17a,Rana-IROS-18}.
\end{enumerate}
In this section, we first develop the \textbf{GPMP2} algorithm, which is more efficient compared to GPMP. In Section~\ref{sec:replanning}, we show how Bayes trees can be used to develop a more efficient algorithm for replanning. Finally, we discuss theoretical connections to other areas in Section~\ref{sec:discussion}.

%------------------------------------------------------------------------------------------------------------------------------------------------%
\subsection{Maximum a posteriori inference}

To formulate this problem as inference, we seek to find a trajectory parameterized by $\bm{\theta}$ given desired events $\mathbf{e}$. For example, binary events $e_i$ at $t_i$ might signify that the trajectory is collision-free if all $e_i = 0$ (i.e. $\mathbf{e} = \mathbf{0}$) and in collision if any $e_i = 1$. In general, the motion planning problem can be formulated with any set of desired events, but we will primarily focus on the collision-free events in this paper.

The posterior density of $\bm{\theta}$ given $\mathbf{e}$ can be computed by Bayes rule from a prior and likelihood
\begin{align}
p(\bm{\theta} | \mathbf{e}) &=  p(\bm{\theta}) p(\mathbf{e} | \bm{\theta}) / p(\mathbf{e})  \\
&\propto p(\bm{\theta}) p(\mathbf{e} | \bm{\theta}),
\end{align}
where $p(\bm{\theta})$ is the {prior} on $\bm{\theta}$ that encourages {smooth} trajectories, and $p(\mathbf{e} | \bm{\theta})$ is the {likelihood} that the trajectory $\bm\theta$ is collision free.
The optimal trajectory $\bm{\theta}$ is found by the
\emph{maximum a posteriori} (MAP) estimator, which chooses the trajectory that maximizes the posterior  $p(\bm{\theta} | \mathbf{e})$
\begin{align}
\bm{\theta}^* &= \argmax_{\bm{\theta}}  p(\bm{\theta} | \mathbf{e}) \\
&= \argmax_{\bm{\theta}}  p(\bm{\theta}) l(\bm{\theta} ; \mathbf{e}), \label{eq:map}
\end{align}
where $l(\bm{\theta} ; \mathbf{e})$ is the likelihood of states $\bm{\theta}$ given events $\mathbf{e}$ on the whole trajectory%
\begin{equation}
l(\bm{\theta} ; \mathbf{e}) \propto p(\mathbf{e} | \bm{\theta}). 
\end{equation}
We use the same GP prior as in Section~\ref{sec:gp}
\begin{align}\label{eq:prob_gp2}
	p(\bm{\theta}) &\propto \exp \bigg\{ - \frac{1}{2} \parallel \bm{\theta} - \bm{\mu} \parallel^{2}_{\bm{\mathcal{K}}} \bigg\}.
\end{align}
The collision free likelihood is defined as a distribution in the exponential family
\begin{equation}\label{eq:obs_likelihood}
	l(\bm{\theta} ; \mathbf{e}) = \exp \bigg\{ - \frac{1}{2} \parallel \bm{h}(\bm{\theta}) \parallel^{2}_{\mathbf{\Sigma}_{obs}} \bigg\}
\end{equation}
where 
$\bm{h}(\bm{\theta})$ is a vector-valued \emph{obstacle cost} for the trajectory, and $\mathbf{\Sigma}_{obs}$ is a diagonal matrix and the hyperparameter of the distribution. The specific obstacle cost  used in our implementation is defined in Section \ref{sec:fg}.

%------------------------------------------------------------------------------------------------------------------------------------------------%
\subsection{Factor graph formulation}\label{sec:fg}

Given the Markovian structure of the trajectory and sparsity of inverse kernel matrix, the posterior distribution can be further factored such that MAP inference can be equivalently viewed as performing inference on a \emph{factor graph}~\citep{kschischang2001factor}. 

A factor graph $G = \{\mathbf{\Theta}, \mathcal{F}, \mathcal{E} \}$ is a bipartite graph, which represents a factored function, where $\mathbf{\Theta} = \{\bm{\theta}_0, \dots , \bm{\theta}_N\}$ are a set of variable nodes, $\mathcal{F} \doteq \{ f_0, \dots , f_M \}$ are a set of factor nodes, and $\mathcal{E}$ are edges connecting the two type of nodes.

In our problems, the factorization of the posterior distribution can be written as
\begin{equation}
p(\bm{\theta} | \mathbf{e}) \propto \prod_{m=1}^{M} f_m(\mathbf{\Theta}_m),
\end{equation}
where $f_m$ are factors on variable subsets $\mathbf{\Theta}_m$.

Given the tridiagonal inverse kernel matrix defined by Eq.~\eqref{eq:Kinv}-\eqref{eq:Q}, we factor the prior
\begin{equation}
p(\bm{\theta}) \propto f^{p}_0 (\bm{\theta}_{0}) f^{p}_N (\bm{\theta}_{N}) \prod_{i=0}^{N-1} f^{gp}_i(\bm{\theta}_{i}, \bm{\theta}_{i+1}),
\end{equation}
where $f^{p}_0 (\bm{\theta}_{0})$ and $f^{p}_N (\bm{\theta}_{N})$ define the prior distributions on start and goal states respectively
\begin{equation}
f^{p}_i (\bm{\theta}_{i}) = \exp \bigg\{- \frac{1}{2} \| \bm{\theta}_{i} - \bm{\mu}_{i} \|^{2}_{\bm{\mathcal{K}}_i} \bigg\}, i=0 \text{ or } N
\end{equation}
where $\bm{\mathcal{K}}_0$ and $\bm{\mathcal{K}}_N$ are covariance matrices on start and goal states respectively, and $\bm{\mu}_{0}$ and $\bm{\mu}_{N}$ are prior (known) start and goal states respectively. The GP prior factor is
\begin{align}\label{eq:gp_prior_factor}
\small
	& f^{gp}_i (\bm{\theta}_{i}, \bm{\theta}_{i+1}) =  \\
	& \hspace{3mm} \exp \bigg\{\hspace{-1mm} - \frac{1}{2} \| \mathbf{\Phi}(t_{i+1}, t_{i})\bm{\theta}_{i} - \bm{\theta}_{i+1} +  \mathbf{u}_{i,i+1} \|^{2}_{\mathbf{Q}_{i,i+1}} \bigg\} \hspace{-1mm} \nonumber
\end{align}
where $\mathbf{u}_{a,b} = \int_{t_{a}}^{t_{b}} \mathbf{\Phi}(b,s) \mathbf{u}(s) \diff s$, $\mathbf{\Phi}(t_{i+1}, t_{i})$ is the state transition matrix, and $\mathbf{Q}_{i,i+1}$ is defined by Eq.~\eqref{eq:Q} (see \cite{Yan17ras} for details). 

\begin{figure}[t]
	\begin{centering}
		{\includegraphics[width=0.98\columnwidth]{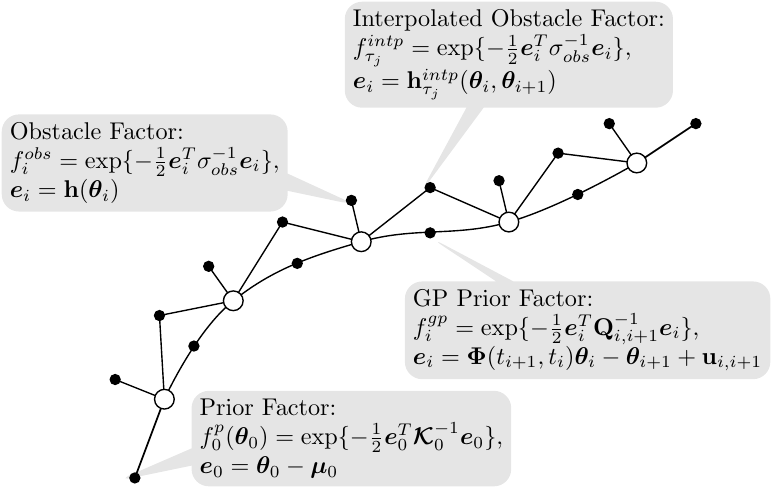}}
		\par\end{centering}
	\protect\caption{A factor graph of an example trajectory optimization problem showing support states (white circles) and four kinds of factors (black dots), namely prior factors on start and goal states, GP prior factors that connect consecutive support states, obstacle factors on each state, and interpolated obstacle factors between consecutive support states (only one shown here for clarity, any number of them may be present in practice).
		\label{fig:factor_graph}}
\end{figure}

To factor the collision-free likelihood $l(\bm{\theta};\mathbf{e})$, we define two types of obstacle cost factors: regular obstacle factors $f^{obs}_i$ and interpolated obstacle factors $f^{intp}_{\tau_j}$. The $l(\bm{\theta};\mathbf{e})$ is the product of all obstacle factors
\begin{equation}\label{eq:obs_likelihood_decomp}
l(\bm{\theta};\mathbf{e}) = \prod_{i=0}^{N} \bigg\{ f^{obs}_i (\bm{\theta}_i) \prod_{j=1}^{n_{ip}} f^{intp}_{\tau_j} (\bm{\theta}_i, \bm{\theta}_{i+1}) \bigg\},
\end{equation}
where $n_{ip}$ is the number of \emph{interpolated} states defined between each nearby support state pair $\bm{\theta}_i$ and $\bm{\theta}_{i+1}$, and $\tau_j$ is the time to perform interpolation which satisfies $t_i < \tau_j < t_{i+1}$.

The regular obstacle factor describes the obstacle cost on a single state variable and is a \emph{unary} factor defined as
\begin{align}\label{eq:obs_factor}
	f^{obs}_i (\bm{\theta}_{i}) = \exp \bigg\{ - \frac{1}{2} \parallel \mathbf{h}(\bm{\theta}_i) \parallel^{2}_{\bm{\sigma}_{obs}} \bigg\},
\end{align}
where $\mathbf{h}(\bm{\theta}_i)$ is a $M$-dimensional vector-valued obstacle cost function for a single state, and $\bm{\sigma}_{obs}$ is a $M \times M$ hyperparameter matrix.

The interpolated obstacle factor describes the obstacle cost at $\tau_j$, which is not on any support state and needs be interpolated from the support states. Since the Gauss-Markov model we choose enables fast interpolation from adjacent states, we can interpolate a state at any $\tau_j$ from $\bm{\theta}_i$ and $\bm{\theta}_{i+1}$ by Eq.~\eqref{eq:interpolation}, which satisfies $t_i < \tau_j < t_{i+1}$. This allows us to derive a \emph{binary} interpolated obstacle factor that relates the cost at an interpolated point to the adjacent two trajectory states
\begin{align}\label{eq:intp_factor}
	f^{intp}_{\tau_j} (\bm{\theta}_{i}, \bm{\theta}_{i+1}) &= \exp \bigg\{ - \frac{1}{2} \parallel \mathbf{h}(\bm{\theta}(\tau_j)) \parallel^{2}_{\bm{\sigma}_{obs}} \bigg\} \\ &= \exp \bigg\{ - \frac{1}{2} \parallel \mathbf{h}^{intp}_{\tau_j}(\bm{\theta}_i, \bm{\theta}_{i+1}) \parallel^{2}_{\bm{\sigma}_{obs}} \bigg\} \nonumber.
\end{align}
In other words, $\bm{\theta}(\tau_j)$ is a function of $\bm{\theta}_i$ and $\bm{\theta}_{i+1}$ (see Eq.~\eqref{eq:interpolation}). Just like in GPMP, here too the interpolated obstacle factor incorporates the obstacle information at all $\tau$ in the factor graph and is utilized to meaningfully update the sparse set of support states. 

An example factor graph that combines all of the factors described above is illustrated in Figure \ref{fig:factor_graph}. Note that if there are enough support states to densely cover the trajectory, interpolated obstacle factors are not needed. But to fully utilize the power of the continuous-time trajectory representation and to maximize performance, the use of sparse support states along with interpolated obstacle factor is encouraged.

Given the factorized obstacle likelihood in Eq.~\eqref{eq:obs_likelihood_decomp}-\eqref{eq:intp_factor}, we can retrieve the vector-valued obstacle cost function of the trajectory defined in Eq.~\eqref{eq:obs_likelihood} by simply stacking all the vector-valued obstacle cost functions on all regular and interpolated states into a single vector
\begin{align}
\bm{h}(\bm{\theta}) = \big[ &\mathbf{h}(\bm{\theta}_0); \mathbf{h}^{intp}_{\tau_1}(\bm{\theta}_0, \bm{\theta}_1); \hdots; \mathbf{h}^{intp}_{\tau_{n_{ip}}}(\bm{\theta}_0, \bm{\theta}_1); \label{eq:vector_obs_cost} \\ 
&\mathbf{h}(\bm{\theta}_1); \mathbf{h}^{intp}_{\tau_1}(\bm{\theta}_1, \bm{\theta}_2); \hdots; \mathbf{h}^{intp}_{\tau_{n_{ip}}}(\bm{\theta}_1, \bm{\theta}_2); \nonumber \\
&\hdots \nonumber \\
&\mathbf{h}(\bm{\theta}_{N-1}); \mathbf{h}^{intp}_{\tau_1}(\bm{\theta}_{N-1}, \bm{\theta}_N); \hdots; \mathbf{h}^{intp}_{\tau_{n_{ip}}}(\bm{\theta}_{N-1}, \bm{\theta}_N); \nonumber \\ 
&\mathbf{h}(\bm{\theta}_N)], \nonumber
\end{align}
where all $\mathbf{h}$ are obstacle cost functions from regular obstacle factors defined in Eq.~\eqref{eq:obs_factor}, and all $ \mathbf{h}^{intp}$ are obstacle cost functions from interpolated obstacle factors defined in Eq.~\eqref{eq:intp_factor}. Since there are a total of $N~+~1$ regular obstacle factors on support states, and $n_{ip}$ interpolated factors between each support state pair, the total dimensionality of $\bm{h}(\bm{\theta})$ is $M \times (N+1 + N \times n_{ip})$. The hyperparameter matrix $\mathbf{\Sigma}_{obs}$ in Eq.~\eqref{eq:obs_likelihood} is then defined by
\begin{equation}\label{eq:hyper_obs_matrix}
\mathbf{\Sigma}_{obs} = \begin{bmatrix} \bm{\sigma}_{obs} & & \\ & \ddots & \\ & & \bm{\sigma}_{obs} \end{bmatrix},
\end{equation}
which has size $M \times (N+1 + N \times n_{ip})$ by $M \times (N+1 + N \times n_{ip})$.

In our framework, the obstacle cost function $\mathbf{h}$ can be any nonlinear function, and the construction of $\mathbf{h}$, $M$, and $\bm{\sigma}_{obs}$ are flexible as long as $l(\bm{\theta};\mathbf{e})$ gives the collision-free likelihood. Effectively $\mathbf{h}(\bm{\theta}_i)$ should have a larger value when a robot collides with obstacles at $\bm{\theta}_i$, and a smaller value when the robot is collision-free. Our implementation of $\mathbf{h}$, definition of $M$, and guideline for the hyperparameter $\bm{\sigma}_{obs}$ is discussed in Section~\ref{sec:cost_function}.

%------------------------------------------------------------------------------------------------------------------------------------------------%
\subsection{Computing the MAP trajectory}\label{sec:MAP}

To solve the MAP inference problem in Eq.~\eqref{eq:map}, we first illustrate the duality between inference and optimization by performing minimization on the negative log of the posterior distribution
\begin{align}
	\bm{\theta}^*
	&= \argmax_{\bm{\theta}} p(\bm{\theta}) l(\bm{\theta} ; \mathbf{e}) \nonumber  \\
	&= \argmin_{\bm{\theta}} \bigg\{ - \log \Big( p(\bm{\theta}) l(\bm{\theta} ; \mathbf{e}) \Big) \bigg\} \nonumber\\
	&= \argmin_{\bm{\theta}} \bigg\{ \frac{1}{2} \parallel \bm{\theta} - \bm{\mu}
	\parallel^{2}_{\bm{\mathcal{K}}} + \frac{1}{2} \parallel \bm{h}(\bm{\theta})
	\parallel^{2}_{\mathbf{\Sigma}_{obs}} \bigg\} \label{eq:least_square}
\end{align}
where Eq.~\eqref{eq:least_square} follows from Eq.~\eqref{eq:prob_gp2} and Eq.~\eqref{eq:obs_likelihood}. This duality connects the two different perspectives on motion planning problems such that the terms in Eq. \eqref{eq:least_square} can be viewed as `cost' to be minimized, or information to be maximized. The apparent construction of the posterior now becomes clear as we have a nonlinear least squares optimization problem, which has been well studied and for which many numerical tools are available. Iterative approaches, like Gauss-Newton or Levenberg-Marquardt repeatedly resolve a quadratic approximation of Eq.~\eqref{eq:least_square} until convergence. 

Linearizing the nonlinear obstacle cost function around the current trajectory
${\bm{\theta}}$
\begin{gather}
\bm{h}(\bm{\theta} + \diff \bm{\theta}) \approx \bm{h}({\bm{\theta}}) + \mathbf{H}\diff \bm{\theta} \\
\mathbf{H} = \frac{\diff \bm{h}}{\diff \bm{\theta}} \Bigr|_{{\bm{\theta}}}
\end{gather}
where $\mathbf{H}$ is the Jacobian matrix of $\bm{h}(\bm{\theta})$, we convert Eq.~\eqref{eq:least_square} to a linear least squares problem
\begin{equation}\label{eq:linear_least_square}
\small
\hspace{-2mm}	\delta \bm{\theta}^* = \argmin_{\delta \bm{\theta}} \bigg\{  \frac{1}{2} \|\bm{\theta} + \delta\bm{\theta} - \bm{\mu} \parallel^{2}_{\bm{\mathcal{K}}} + \frac{1}{2} \parallel \bm{h}({\bm{\theta}}) + \mathbf{H} \delta\bm{\theta} \parallel^{2}_{\mathbf{\Sigma}_{obs}} \hspace{-2mm} \bigg\}.\hspace{-2mm}
\end{equation}
The optimal perturbation $\delta\bm{\theta}^*$ results from solving the following linear system
\begin{equation}\label{eq:linear_system}
\small
	(\bm{\mathcal{K}}^{-1} + \mathbf{H}^\top \mathbf{\Sigma}_{obs}^{-1} \mathbf{H}) \delta \bm{\theta}^* = -\bm{\mathcal{K}}^{-1}({\bm{\theta}} - \bm{\mu}) - \mathbf{H}^\top \mathbf{\Sigma}_{obs}^{-1} \bm{h}({\bm{\theta}})
\end{equation}
Once the linear system is solved, the iteration
\begin{equation}\label{eq:iter}
	{\bm{\theta}} \leftarrow {\bm{\theta}} + \delta \bm{\theta}^*
\end{equation}
is applied until convergence criteria are met. Eq.~\eqref{eq:iter} serves as the update rule for GPMP2.

If the linear system in Eq.~\eqref{eq:linear_system} is sparse, then $\delta\bm{\theta}^* $ can be solved efficiently by exploiting the sparse Cholesky decomposition followed by forward-backward passes~\citep{golub2012matrix}. Fortunately, this is the case: we have selected a Gaussian process prior with a block tridiagonal precision matrix $\bm{\mathcal{K}}^{-1}$ (Section~\ref{sec:ltvsde}) and $\mathbf{H}^\top \mathbf{\Sigma}_{obs}^{-1} \mathbf{H}$ is also block tridiagonal (see proof in Appendix B). The structure exploiting iteration combined with the quadratic convergence rate of nonlinear least squares optimization method we employ (Gauss-Newton or Levenberg-Marquardt) makes GPMP2 more efficient and faster compared to GPMP.

%%%%%%%%%%%%%%%%%%%%%%%%%%%%%%%%%%%%%%%%%%%%%%%%%%%%%%%%%%%%%%%%%%%%%%%%%%%%%%%%%%%%%%%%%%%%%%%%%%%%%%%%%%%%%%%%

\section{Incremental inference for fast replanning}\label{sec:replanning}

We have described how to formulate motion planning problem as probabilistic inference on factor graphs, results in fast planning through least squares optimization. In this section, we show that this perspective also gives us the flexibility to use other inference and optimization tools on factor graphs. In particular, we describe how factor graphs can be used to perform \emph{incremental} updates to solve \emph{replanning} problems efficiently.

The replanning problem can be defined as: given a solved motion planning problem, resolve the problem with partially changed conditions. Replanning problems are commonly encountered in the real world, when, for example: (i) the goal position for the end-effector has changed during middle of the execution; (ii) the robot receives updated estimation about its current state; or (iii) new information about the environment is available. Since replanning is performed online, possibly in dynamic environments, fast real-time replanning is critical to ensuring safety.

A na{\"i}ve way to solve this problem is to literally replan by re-optimizing from scratch. However, this is potentially too slow for real-time settings. Furthermore, if the majority of the problem is left unchanged, resolving the entire problem duplicates work and should be avoided to improve efficiency.

Here we adopt an incremental approach to updating the current solution given new or updated information. We use the \emph{Bayes tree}~\citep{kaess2011bayes,kaess2011isam2} data structure to perform incremental inference on factor graphs.\footnote{Given that the trajectories are represented by GPs, the incremental updates of the factor graphs can also be viewed as incremental GP regression~\citep{Yan17ras}.} 
We first give a brief overview of the Bayes tree and its relation to factor graphs, and then we give a more detailed example to show how to use a Bayes tree to perform incremental inference for a replanning problem.

\subsection{Bayes tree as incremental inference}\label{sec:bayes_tree}

\begin{figure*}[t]
	\begin{centering}
		\begin{subfigure}[b]{0.32\textwidth}
			\centering
			\includegraphics[width=1\linewidth]{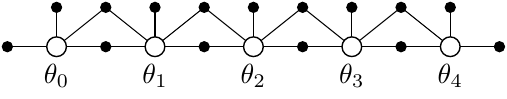}
			\caption{Factor graph}
		\end{subfigure}
		\hfill
		\begin{subfigure}[b]{0.32\textwidth}
			\centering
			\includegraphics[width=1\linewidth]{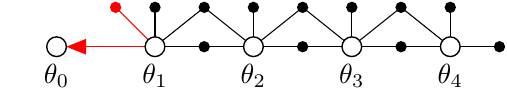}
			\caption{Eliminating $\theta_0$}
		\end{subfigure}	
		\hfill
		\begin{subfigure}[b]{0.32\textwidth}
			\centering
			\includegraphics[width=1\linewidth]{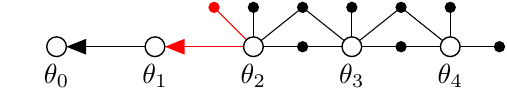}
			\caption{Eliminating $\theta_1$}
		\end{subfigure}
	\end{centering}
	\begin{centering}
		\begin{subfigure}[b]{0.32\textwidth}
			\centering
			\includegraphics[width=1\linewidth]{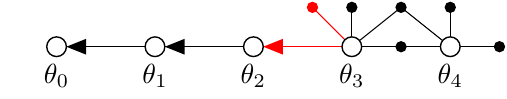}
			\caption{Eliminating $\theta_2$}
		\end{subfigure}
		\hfill
		\begin{subfigure}[b]{0.32\textwidth}
			\centering
			\includegraphics[width=1\linewidth]{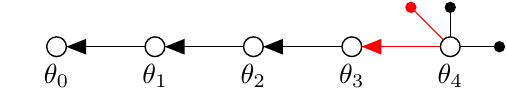}
			\caption{Eliminating $\theta_3$}
		\end{subfigure}	
		\hfill
		\begin{subfigure}[b]{0.32\textwidth}
			\centering
			\includegraphics[width=1\linewidth]{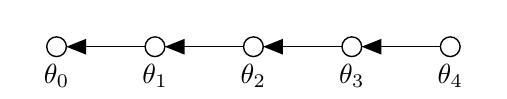}
			\caption{Eliminating $\theta_4$ and final Bayes net}
		\end{subfigure}
	\end{centering}
	\caption{Example of applying variable elimination on a planning factor graph. Red arrows/factors indicate the parts that change in Bayes net/factor graph respectively.}
\label{fig:elimination_example}
\end{figure*}

Before introducing Bayes tree, we first briefly review the \emph{variable elimination} process in smoothing and mapping~\citep{dellaert2006square}, which converts a factor graph to a \emph{Bayes net}. More details about the elimination algorithm can be found in~\citep{dellaert2006square,kaess2011bayes}.
Given a factor graph over variables $\Theta$, and a variable elimination ordering, we would like to convert the probabilistic density over all variables $p(\Theta)$ to the form
\begin{equation}
p(\Theta) = \prod_j p(\theta_j | S_j),\label{eq:bayes_net}
\end{equation}
by Algorithm~\ref{alg:elimination}, where $S_j \subset \Theta$ is the separator of $\theta_j$. 
Note that different variable elimination orderings will generate different Bayes nets. Therefore, we want to select an optimal ordering to produce a Bayes net which is as sparse as possible. Given the chain-like factor graph structure in planning problems, as shown in Figure~\ref{fig:factor_graph}, we can simply select the ordering from start state $\theta_0$ to goal state $\theta_N$, which gives us a chain-like Bayes net structure (which is equivalent to $S_j = \{\theta_{j+1}\}$) and the density can be factorized as
\begin{equation}
p(\theta) = p(\theta_N) \prod_{j=0}^{N-1} p(\theta_{j} | \theta_{j+1}).\label{eq:bayes_net_gpmp2}
\end{equation}
An example of applying variable elimination on a planning factor graph of 5 states is shown in Figure~\ref{fig:elimination_example}.

\begin{algorithm}[!t]
\caption{Eliminate variable $\theta_j$ from factor graph}\label{alg:elimination}
\DontPrintSemicolon
1. Remove all factors $f_i$ connected to $\theta_j$ from factor graph, define $S_j = \{$all variables involved in all $f_i\}$\;
3. $f_j(\theta_j, S_j) = \prod_i f_i(\Theta_i)$\;
4. Factorize $f_j(\theta_j, S_j) = p(\theta_j|S_j)f_{new}(S_j)$\;
5. Add $p(\theta_j|S_j)$ in Bayes net, and add $f_{new}(S_j)$ back in factor graph.
\end{algorithm}

A Bayes tree is a tree-structured graphical model that is derived from a Bayes net. The Bayes net generated from the variable elimination algorithm is proved to be chordal~\citep{kaess2011isam2}, so we can always produce junction trees~\citep{cowell2006probabilistic} from such Bayes nets. A Bayes tree is similar to a junction tree but is directed, which represent the conditional relations in factored probabilistic density.

\begin{figure}[t]
	\begin{centering}
		{\includegraphics[width=0.9\columnwidth]{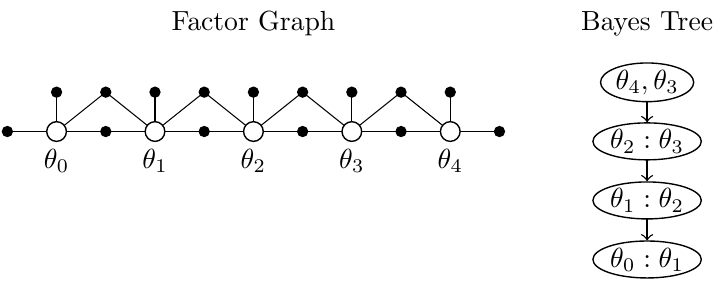}}
		\par\end{centering}
	\protect\caption{Example of a Bayes Tree with its corresponding factor graph. 
		\label{fig:bayes_tree_example}}
\end{figure}

For a Bayes net that results from Algorithm~\ref{alg:elimination}, we extract all cliques $C_k$ and build the Bayes tree by defining all node as cliques $C_k$. For each node of Bayes tree, we define a conditional density $p(F_k|S_k)$, where $S_k$ is the \emph{separator} as intersection $S_k = C_k \cap \Pi_k$ between $C_k$ and $C_k$'s parent $\Pi_k$. The \emph{frontal variable} is defined by $F_k = C_k \backslash S_k$. We can also write the clique as $C_k = F_k : S_k$. The joint density of the Bayes tree is defined by
\begin{equation}
p(\Theta) = \prod_k p_{C_k}(F_k | S_k).\label{eq:bayes_tree}
\end{equation}
For a detailed algorithm to convert an arbitrary chordal Bayes net to a Bayes tree please refer to~\citet{kaess2011isam2}, but since our planning Bayes net has a simple chain structure, as shown in Eq.~\eqref{eq:bayes_net_gpmp2} and Figure~\ref{fig:elimination_example}, all the cliques are $C_k = \{\theta_k, \theta_{k+1}\}$ and the joint density is
\begin{equation}
p(\theta) = p_{C_{N-1}}(\theta_N, \theta_{N-1}) \prod_{k=0}^{N-2} p_{C_k}(\theta_{k} | \theta_{k+1}),\label{eq:bayes_tree_gpmp2}
\end{equation}
and the corresponding Bayes tree is shown in Figure~\ref{fig:bayes_tree_example}.

Since a Bayes tree has nice tree structure, the inference is performed from root to leaves. If variable $\theta_i$ is affected by new factors, only cliques contain $\theta_i$ or cliques between cliques contain $\theta_i$ to root are affected, and need to be re-factorized, remaining cliques can be left unchanged. This makes performing incremental inference with minimal computation possible, since we only have to re-factorize parts of cliques that have changed. The closer the affected cliques are to the root (in planning cases the affected state $\theta_i$ is closer to the goal state $\theta_N$), the less computation is needed for incremental inference. Readers are encouraged to refer to~\citet{kaess2011isam2} for more details about how incremental inference is performed with a Bayes tree.

\subsection{Replanning using Bayes tree}\label{sec:replansub}

\begin{figure}[t]
	\begin{centering}
		{\includegraphics[width=0.9\columnwidth]{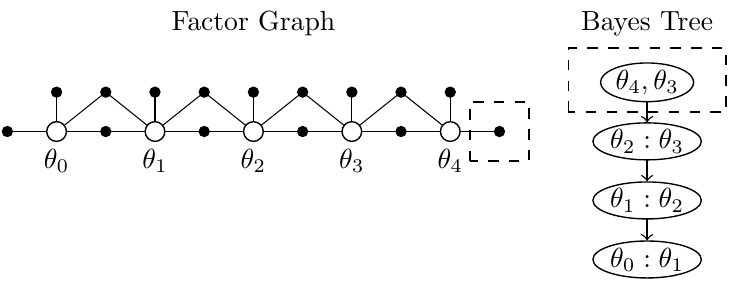}}
		\\[1ex]
		{\includegraphics[width=0.9\columnwidth]{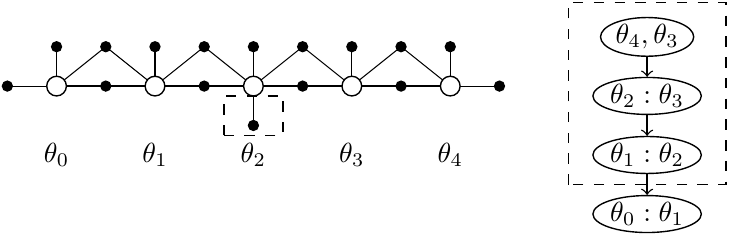}}
		\par\end{centering}
	\protect\caption{Replanning examples using Bayes Trees. Dashed boxes indicate parts of the factor graphs and Bayes Trees that are affected and changed while performing replanning.
		\label{fig:replanning_bayes_tree_example}}
\end{figure}

Two replanning examples with Bayes tree are shown in Figure~\ref{fig:replanning_bayes_tree_example}. The first example shows replanning when the goal configuration changes causing an update to the prior factor on the goal state. When the Bayes tree is updated with the new goal, only the root node of the tree is changed. The second example shows a replanning problem, given an observation of the current configuration (e.g. from perception during execution) that is added as a prior factor at $\bm{\theta}_2$ where the estimation was taken. When the Bayes Tree is updated, the parts of the tree that change correspond to the parts of the trajectory that get updated. 

In our implementation, we use the iSAM2 incremental solver~\citep{kaess2011isam2} within the GPMP2 framework to solve the replanning problem. We call this incremental variant of GPMP2, \textbf{iGPMP2}. A replanning scenario typically has the following steps. First, the original batch problem is solved with GPMP2. Then, we collect the additional information to form factors that need to be added or replaced within the factor graph. Finally, we run Algorithm~\ref{alg:isam_replan} to update the Bayes Tree inside iSAM2, to get a newly updated optimal solution.

\begin{algorithm}[!t]
\caption{Replanning using iSAM2}\label{alg:isam_replan}
\DontPrintSemicolon
\SetAlgoLined
\SetKwInOut{Input}{Input}
\SetKwInOut{Output}{Output}
\Input{new factors $f_{new}$, replaced factors $f_{replace}$}
\Output{updated optimal trajectory $\bm{\theta}^*$}
\textit{Initialization} :\;
add factors $f_{add} = \emptyset$, remove factors $f_{remove} = \emptyset$\;
\textit{iSAM2 update} :\;
$f_{add} = f_{new}$\;
\If{($f_{replace} \neq \emptyset$)}{
	$f_{add} = f_{add} + f_{replace}$\;
	$f_{remove} = \ $\textit{findOldFactors}($f_{replace}$)\;
}
iSAM2.\textit{updateBayesTree}($f_{add}$, $f_{remove}$)\;
\textbf{return} iSAM2.\textit{getCurrentEstimation}()\;
\end{algorithm}

%%%%%%%%%%%%%%%%%%%%%%%%%%%%%%%%%%%%%%%%%%%%%%%%%%%%%%%%%%%%%%%%%%%%%%%%%%%%%%%%%%%%%%%%%%%%%%%%%%%%%%%%%%%%%%%%

\section{Implementation details}\label{sec:details}

GPMP is implemented on top of the CHOMP~\citep{zucker2013chomp} code since it uses an identical framework, albeit with several augmentations. To implement GPMP2 and iGPMP2 algorithms, we used the GTSAM \citep{frank2012factor} library. Our implementation is available as a single open source C++ library, \texttt{gpmp2}.\footnote{Available at \url{https://github.com/gtrll/gpmp2}} We have also released a ROS interface as part of the PIPER~\citep{mukadam2017piper} package. In this section we describe the implementation details of our algorithms.

\subsection{GPMP}\label{sec:gpmp_imp}

\subsubsection{GP prior:}
GPMP employs a constant-acceleration (i.e. jerk-minimizing) prior to generate a trajectory with a Markovian state comprising of configuration position, velocity and acceleration, by following the LTV-SDE in Eq.~\eqref{eq:LVT-SDE} with parameters
\begin{equation}
	\mathbf{A}(t) = \begin{bmatrix} \mathbf{0} & \mathbf{I} & \mathbf{0} \\ \mathbf{0} & \mathbf{0} & \mathbf{I} \\ \mathbf{0} & \mathbf{0} & \mathbf{0} \end{bmatrix}, \mathbf{u}(t) = \mathbf{0}, 
	\mathbf{F}(t) = \begin{bmatrix} \mathbf{0} \\ \mathbf{0} \\ \mathbf{I} \end{bmatrix} 
\end{equation}
and given $\Delta t_i = t_{i+1} - t_{i}$,
\begin{gather}
	\mathbf{\Phi}(t,s) = \begin{bmatrix}
	\mathbf{I} & (t-s) \mathbf{I} & \frac{1}{2}(t-s)^2 \mathbf{I} \\ 
	\mathbf{0} & \mathbf{I} & (t-s)\mathbf{I} \\
	\mathbf{0} & \mathbf{0} & \mathbf{I} \end{bmatrix} \\ 
	\mathbf{Q}_{i,i+1} = \begin{bmatrix}
	\frac{1}{2}\Delta	t_i^5	\mathbf{Q}_C & \frac{1}{8}\Delta t_i^4 \mathbf{Q}_C & \frac{1}{6}\Delta t_i^3 \mathbf{Q}_C \\
	\frac{1}{8}\Delta	t_i^4	\mathbf{Q}_C & \frac{1}{3}\Delta t_i^3 \mathbf{Q}_C & \frac{1}{2}\Delta t_i^2 \mathbf{Q}_C \\
	\frac{1}{6}\Delta	t_i^3	\mathbf{Q}_C & \frac{1}{2}\Delta t_i^2 \mathbf{Q}_C & \Delta t_i \mathbf{Q}_C \end{bmatrix}
\end{gather}
This prior is centered around a zero jerk trajectory and encourages smoothness by attempting to minimize jerk during optimization. 

\subsubsection{Obstacle avoidance and constraints:}
\begin{figure}
\begin{centering}
{\includegraphics[width=0.6\columnwidth]{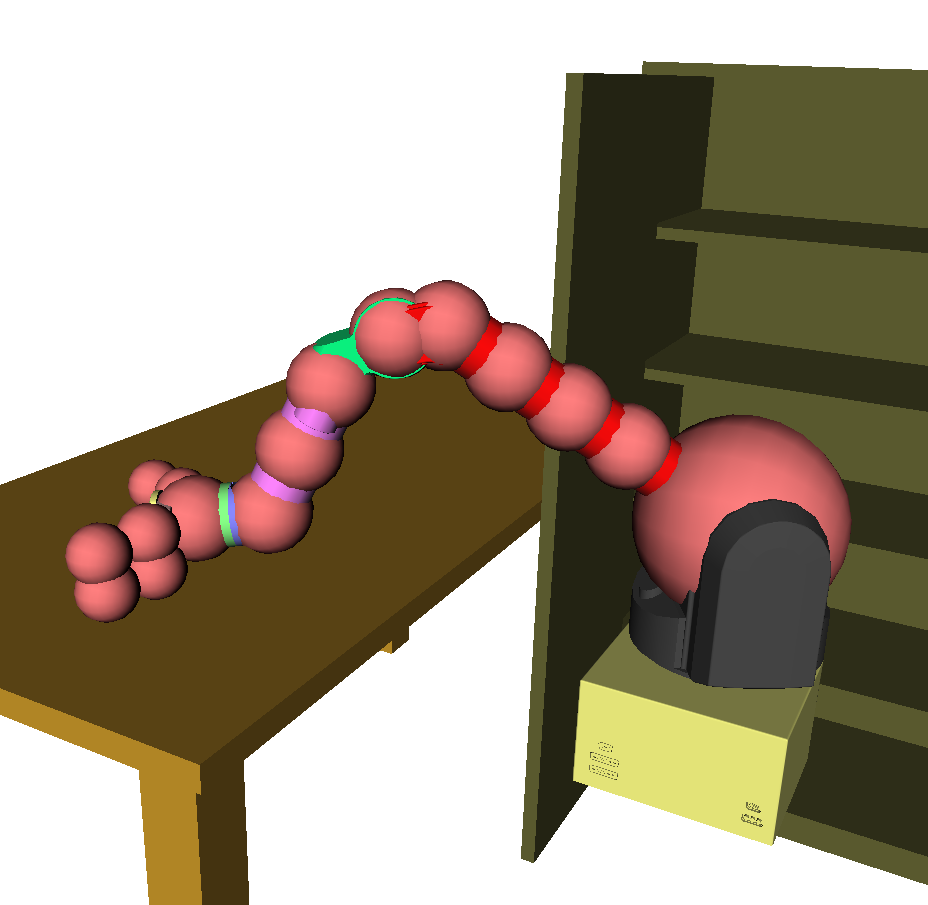}}
\par\end{centering}
\protect\caption{
The WAM arm is represented by multiple spheres (pink), which are used during collision cost calculation.
\label{fig:sphere}}
\end{figure}

To quickly calculate the collision cost for an arbitrary shape of the robot's physical body, GPMP represents the robot with a set of spheres, as in~\citet{zucker2013chomp} (shown in Figure ~\ref{fig:sphere}). This leads to a more tractable approximation to finding the signed distance from the robot surface to obstacles.
GPMP uses the same obstacle cost function as CHOMP (see Eq.~\eqref{eq:obs_cont}) where the cost is summed over the sphere set on the robot body calculated using a precomputed \emph{signed distance field} (SDF). 
Constraints are also handled in the same manner as CHOMP. Joint limits are enforced by smoothly projecting joint violations using the technique similar to projecting the obstacle gradient in Eq.~\eqref{eq:gpmp_update}. Along each point on the up-sampled trajectory the violations are calculated via $L_1$ projections to bring inside the limits (see~\cite{zucker2013chomp} for details). Then they are collected into a violation trajectory, $\bm{\theta}^v_{up}$ to be projected:
\begin{equation}\label{eq:jl_gp,p}
	\bm{\theta} = \bm{\theta} + \mathcal{K} \mathbf{M}^\top \bm{\theta}^v_{up}.
\end{equation}

%------------------------------------------------------------------------------------------------------------------------------------------------%
\subsection{GPMP2 and iGPMP2}\label{sec:gpmp2_imp}
GPMP2 uses the Levenberg-Marquardt algorithm to solve the nonlinear least squares optimization problem, with the initial damping parameter set as $0.01$. The optimization is stopped if a maximum of 100 iterations is reached, or if the relative decrease in error is smaller than $10^{-4}$. iGPMP2 uses the iSAM2 \citep{kaess2011isam2} incremental optimizer with default settings.

\subsubsection{GP prior:}\label{sec:const_gp_prior}
We use a constant-velocity prior in GPMP2 with the Markovian state comprising of configuration position and velocity.  Note that, unlike GPMP, we did not include acceleration since it was not needed for any gradients and an acceleration-minimizing prior for optimization was sufficient for the tasks we consider in this work. Ideally a jerk-minimizing trajectory would be beneficial to use on faster moving systems like quadrotors. GPMP2 scales only quadratically in computation with the size of the state. So even if the same prior as GPMP was used, GPMP2 would still be faster given its quadratic convergence rate.

The trajectory is similarly generated by following the LTV-SDE in Eq.~\eqref{eq:LVT-SDE} with
\begin{equation}
	\mathbf{A}(t) = \begin{bmatrix} \mathbf{0} & \mathbf{I} \\ 
	\mathbf{0} & \mathbf{0} \end{bmatrix}, \mathbf{u}(t) = \mathbf{0}, 
	\mathbf{F}(t) = \begin{bmatrix} \mathbf{0} \\ \mathbf{I} \end{bmatrix}
\end{equation}
and given $\Delta t_i = t_{i+1} - t_{i}$,
\begin{equation}
\small
	\mathbf{\Phi}(t,s) = \begin{bmatrix} \mathbf{I} & (t-s)\mathbf{I} \\ 
	\mathbf{0} & \mathbf{I} \end{bmatrix},
	\mathbf{Q}_{i,i+1} = \begin{bmatrix} \frac{1}{3} \Delta t_i^3 \mathbf{Q}_C &
	\frac{1}{2} \Delta t_i^2 \mathbf{Q}_C \\ 
	\frac{1}{2} \Delta t_i^2 \mathbf{Q}_C &
	\Delta t_i \mathbf{Q}_C \end{bmatrix}
\end{equation}
Analogously this prior is centered around a zero-acceleration trajectory.

\subsubsection{Collision-free likelihood:}\label{sec:cost_function}

Similar to GPMP and CHOMP, the robot body is represented by a set of spheres as shown in Figure~\ref{fig:sphere}, and the obstacle cost function for any configuration $\bm{\theta}_i$ is then completed by computing the hinge loss for each sphere $S_j$ ($j=1,\dots,M$) and collecting them into a single vector,
\begin{equation} \label{eq:imp_obs_cost}
	\mathbf{h}(\bm \theta_i) = [ \mathbf c( \mathbf{d}( \mathbf{x}(\bm\theta_i,S_j)))] \Bigr|_{1 \leq j \leq M}
\end{equation}
where $\mathbf{x}$ is the forward kinematics, $\mathbf{d}$ is the signed distance function, $\mathbf c$ is the hinge loss function, and $M$ is the number of spheres that represent the robot model.

Forward kinematics $\mathbf{x}(\bm\theta_i,S_j)$ maps any configuration $\bm{\theta}_i$ to the 3D workspace, to find the center position of any sphere $S_j$. Given a sphere and its center position, we calculate $\mathbf{d}(x)$, the \emph{signed distance} from the sphere at $x$ to the closest obstacle surface in the workspace. The sphere shape makes the surface-to-surface distance easy to calculate, since it is equal to the distance from sphere center to closest obstacle surface minus the sphere radius. Using a precomputed signed distance field (SDF), stored in a voxel grid with a desired resolution, the signed distance of any position in 3D space is queried by trilinear interpolation on the voxel grid. The hinge loss function\footnote{The hinge loss function is not differentiable at $d = \epsilon$, so in our implementation we set $\diff \mathbf{c}(d) / \diff d = -0.5$ when $d = \epsilon$.} is defined as
\begin{equation}
	\mathbf{c}(d) =
	\begin{cases} 
	      \hfill -d + \epsilon \hfill & \text{if} \ d \leqslant \epsilon  \\
	      \hfill 0 \hfill & \text{if} \ d > \epsilon \\
	\end{cases}	\label{eq:hinge_loss}
\end{equation}
where $d$ is the signed distance, and $\epsilon$ is a `safety distance' indicating the boundary of the `danger area' near obstacle surfaces. By adding a non-zero obstacle cost, even if the robot is not in collision but rather too close to the obstacles, $\epsilon$ enables the robot to stay a minimum distance away from obstacles. The remaining parameter $\bm{\sigma}_{obs}$ needed to fully implement the likelihood in Eq.~\eqref{eq:obs_factor} and Eq.~\eqref{eq:intp_factor} is defined by an isotropic diagonal matrix
\begin{equation} \label{eq:hyper_obs_matrix_small}
\bm{\Sigma}_{obs} = \sigma_{obs}^2 \mathbf{I},
\end{equation}
where $\sigma_{obs}$ is the `obstacle cost weight' parameter.

\begin{figure}
\begin{centering}
{\includegraphics[width=0.6\columnwidth]{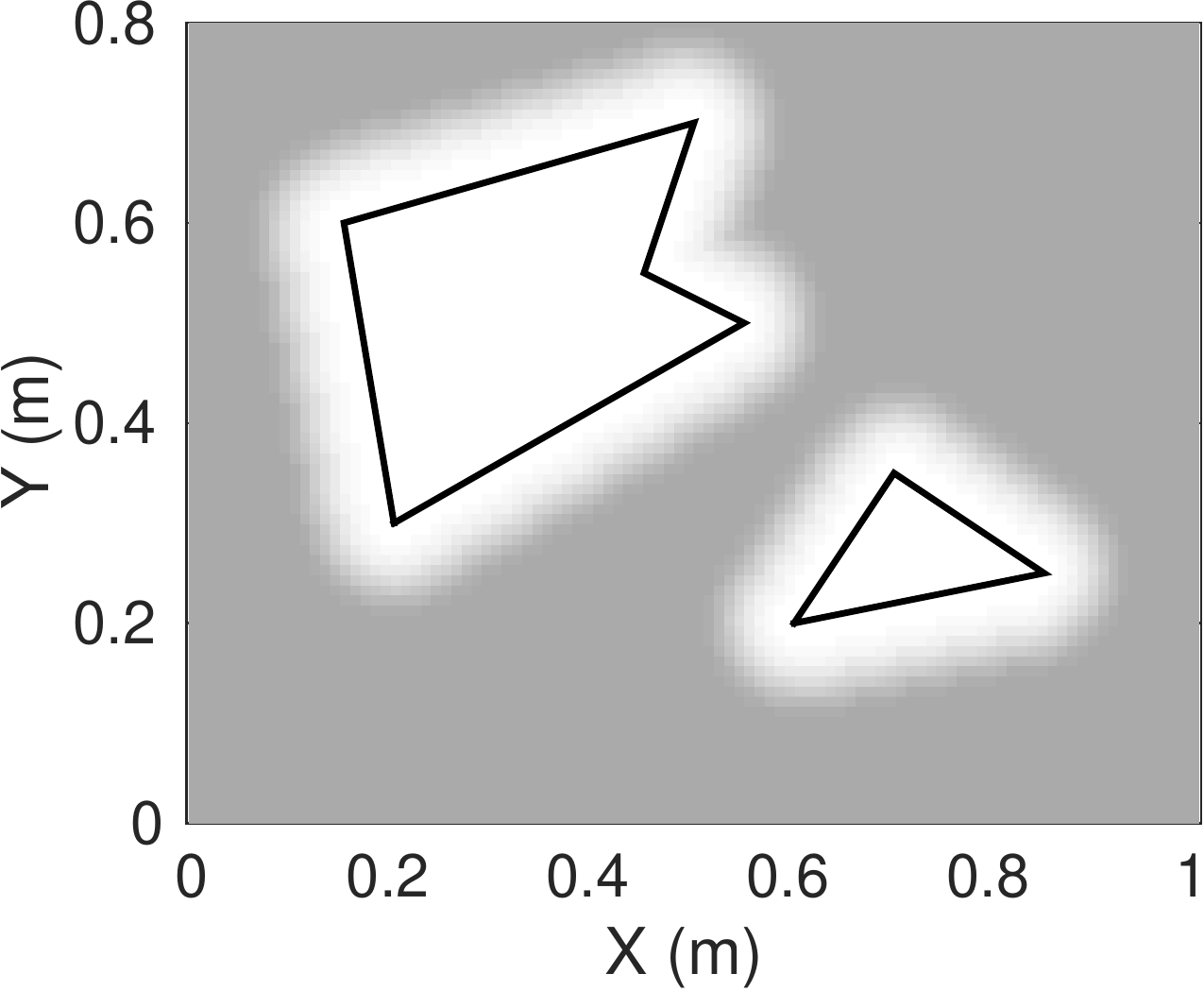}}
\par\end{centering}
\protect\caption{
The likelihood function $\mathbf{h}$ in a 2D space with two obstacles and $\epsilon = 0.1$m. Obstacles are marked by black lines and darker area has higher likelihood for no-collision.
\label{fig:hinge_loss}}
\end{figure}

Figure~\ref{fig:hinge_loss} visualizes a 2D example of the collision-free likelihood defined by the obstacle cost function in Eq.~\eqref{eq:imp_obs_cost}. The darker region shows a free configuration space where the likelihood of no-collision is high. The small area beyond the boundary of the obstacles is lighter, implying `safety marginals' defined by $\epsilon$.

Note that the obstacle cost function used here is different from the one used in GPMP and CHOMP, where $\mathbf{c}$ is instead a smooth function (necessary for gradient calculation) and is multiplied with the norm of the workspace velocity (see Eq.~\eqref{eq:obs_cont}). This arc-length parameterization helps in making the trajectory avoid obstacles rather than speeding through them, while minimizing cost. The GP prior we use for GPMP2 helps us achieve the same purpose, by incorporating cost on large accelerations. The choice of cost function in Eq.~\eqref{eq:hinge_loss} serves as a good approximation for the tasks we consider and is also less computationally expensive.

\subsubsection{Motion constraints:}\label{sec:constrain}

Motion constraints exist in real-world planning problems and should be considered during trajectory optimization. Examples include the constrained start and goal states as well as constraints on any other states along the trajectory. Since we are solving unconstrained least square problems, there is no way to enforce direct equality or inequality constraints during inference. In our framework, these constraints are instead handled in a `soft' way, by treating them as prior knowledge on the trajectory states with very small uncertainties. Although the constraints are not exact, this has not been an issue in practice in any of our evaluations.

Additional \emph{equality} motion constraints, such as end-effector rotation constraints (e.g. holding a cup filled with water upright) written as $\mathbf{f}(\bm{\theta}_{c}) = \mathbf{0}$, where $\bm{\theta}_{c}$ is the set of states involved, can be incorporated into a likelihood,
\begin{equation}
	L_{constraint}(\bm{\theta}) \propto \exp \bigg\{ - \frac{1}{2} \parallel 
	\mathbf{f}(\bm{\theta}_{c})
	\parallel^{2}_{\mathbf{\Sigma}_{c}}
	 \bigg\}\label{eq:likelihood_eq_constr}
\end{equation}
where, $\mathbf{\Sigma}_{c} = \sigma_c^2 \mathbf{I}$, $\sigma_c$ is an arbitrary variance for this constraint, indicating how `tight' the constraint is.

To prevent joint-limit (and velocity-limit) violations, we add \emph{inequality} soft constraint factors to the factor graph. Similar to obstacle factors, the inequality motion constraint factor uses a hinge loss function to enforce soft constraints at both the maximum $\bm\theta_{\text{max}}^{d}$ and the minimum $\bm\theta_{\text{min}}^{d}$ values, with some given safety margin $\epsilon$ on each dimension $d = \{1, \dots, D\}$
\begin{equation}
	\mathbf{c}(\bm\theta_i^{d}) =
	\begin{cases} 
	      \hfill -\bm\theta_i^{d} +\bm\theta_{\text{min}}^{d} - \epsilon \hfill & \text{if} \ \bm\theta_i^{d} < \bm\theta_{\text{min}}^{d} + \epsilon  \\
	      \hfill \bm{0} \hfill & \text{if} \ \bm\theta_{\text{min}}^{d} + \epsilon \leq \bm\theta_i^{d} \leq \bm\theta_{\text{max}}^{d} - \epsilon \\
	      \hfill \bm\theta_i^{d} -\bm\theta_{\text{max}}^{d} + \epsilon  & \text{if} \ \bm\theta_i^{d} > \bm\theta_{\text{max}}^{d} - \epsilon  \\
	\end{cases}	\label{eq:hinge_loss_joint}
\end{equation}
This factor has a vector valued cost function $\mathbf{f}(\bm \theta_i) = [ \mathbf{c}(\bm \theta_i^{d}) ] \Bigr|_{1 \leq d \leq D}$ and the same likelihood as the equality constraint factor in Eq.~\eqref{eq:likelihood_eq_constr}. At the final iteration we also detect limit violations and clamp to the maximum or minimum values.

%%%%%%%%%%%%%%%%%%%%%%%%%%%%%%%%%%%%%%%%%%%%%%%%%%%%%%%%%%%%%%%%%%%%%%%%%%%%%%%%%%%%%%%%%%%%%%%%%%%%%%%%%%%%%%%%
\section{Evaluation}\label{sec:evaluation}

\begin{figure}[!t]
\centering
{\includegraphics[width=0.32\columnwidth]{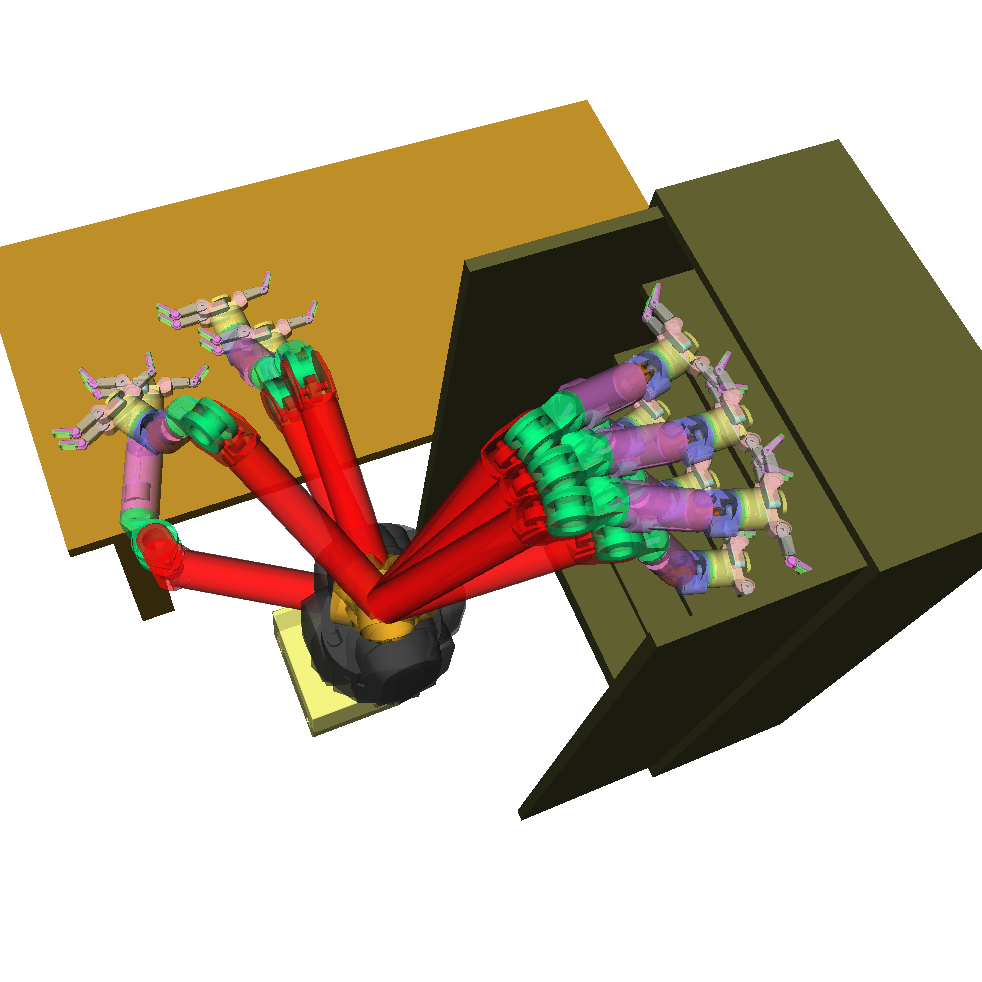}}
{\includegraphics[width=0.32\columnwidth]{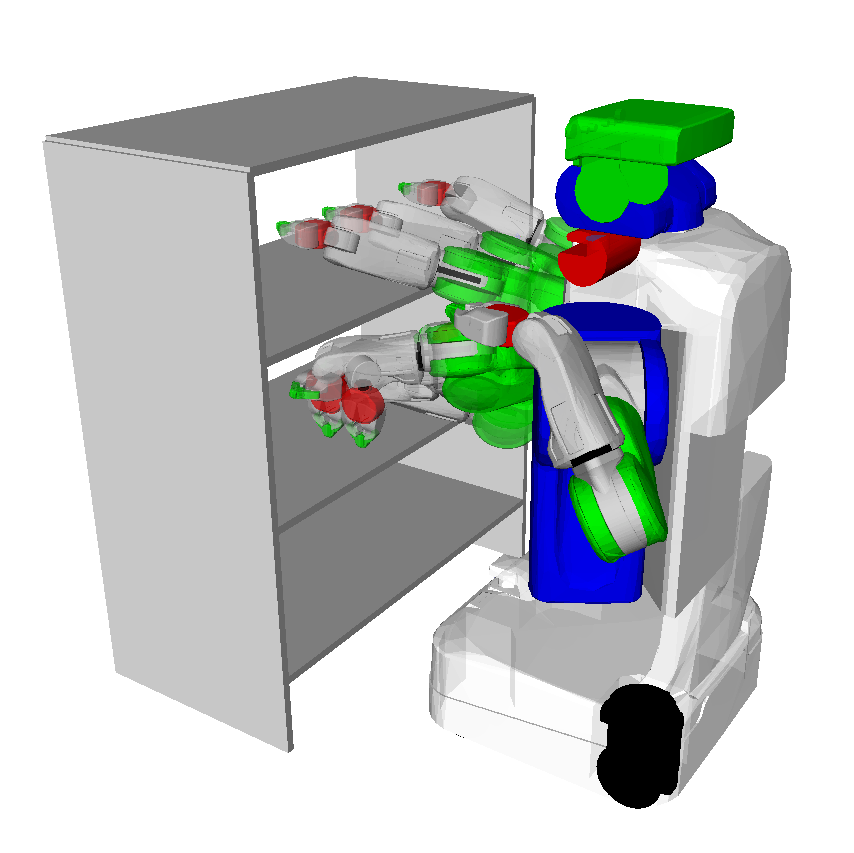}}
{\includegraphics[width=0.32\columnwidth]{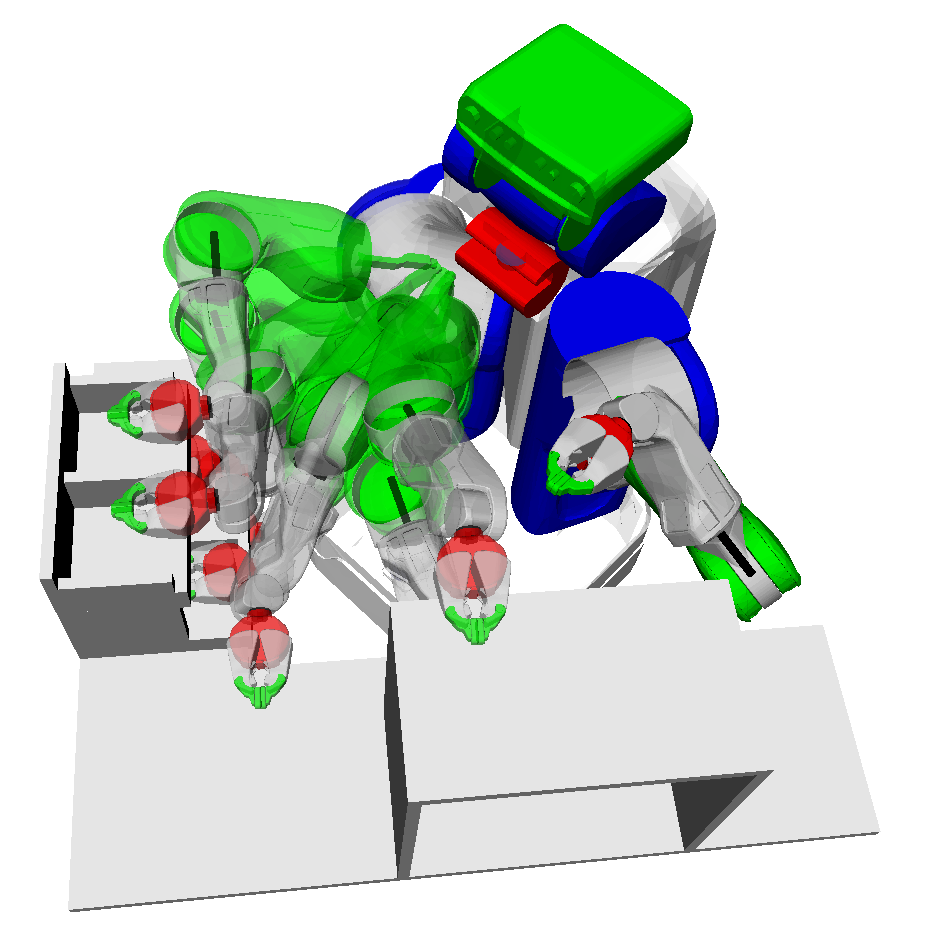}}
\caption{Environments used for evaluation with robot start and goal configurations showing the WAM dataset (left), and a subset of the PR2 dataset (\textit{bookshelves} (center) and \textit{industrial} (right)).}
\label{fig:environments}
\end{figure}

We conducted our experiments\footnote{A video of experiments is available at \url{https://youtu.be/mVA8qhGf7So}.} on two datasets with different start and goal configurations. We used: (1) the 7-DOF WAM arm dataset~\citep{Mukadam-ICRA-16} consisting of 24 unique planning problems in the \textit{lab} environment; and (2) the PR2's 7-DOF right arm dataset~\citep{schulman2014motion} consisting of a total of 198 unique planning problems in four different environments (Figure~\ref{fig:environments}). Finally, we validated successful trajectories on a real 7-DOF WAM arm in an environment identical to the simulation (Figure~\ref{fig:intro}).

%------------------------------------------------------------------------------------------------------------------------------------------------%

\subsection{Batch planning benchmark}\label{sec:evaluation_batch}

\subsubsection{Setup:}

We benchmarked our algorithms, GPMP and GPMP2,  both with interpolation (GPMP2-intp) during optimization and without interpolation (GPMP2-no-intp) against trajectory optimizations algorithms - TrajOpt \citep{schulman2014motion} and CHOMP \citep{zucker2013chomp}, and against sampling based algorithms - RRT-Connect~\citep{kuffner2000rrt} and LBKPIECE~\citep{csucan2009kinodynamic} available within the OMPL implementation~\citep{sucan2012open}. All benchmarks were run on a single thread of a 3.4GHz Intel Core i7 CPU.

For trajectory optimizers, GPMP2, TrajOpt and CHOMP were initialized by a constant-velocity straight line trajectory in configuration space and GPMP was initialized by an acceleration-smooth straight line. For the WAM dataset all initialized trajectories were parameterized by 101 temporally equidistant states. GPMP2-intp and GPMP use interpolation so we initialized them with 11 support states and $n_{ip}=\;$9 (101 states effectively). Since trajectory tasks are shorter in the PR2 dataset, we used 61 temporally equidistant states to initialize the trajectories and for GPMP2-intp and GPMP we used 11 support states and $n_{ip}=\;$5 (61 states effectively). 

To keep comparisons fair we also compared against TrajOpt using only 11 states (TrajOpt-11) in both datasets since it uses continuous-time collision checking and can usually find a successful trajectory with fewer states. Although TrajOpt is faster when using fewer states, post-processing on the resulting trajectory is needed to make it executable and keep it smooth. It is interesting to note that since the continuous time-collision checking is performed only linearly, after the trajectory is post-processed it may not offer any collision-free guarantees. GPMP and GPMP2 avoid this problem when using fewer states by up-sampling the trajectory with GP interpolation and checking for collision at the interpolated points. This up-sampled trajectory remains smooth and can be used directly during execution.

For sampling-based planners no post processing or smoothing step was applied and they were used with default settings.

All algorithms were allowed to run for a maximum of 10 seconds on any problem and marked successful if a feasible solution is found in that time. GPMP, CHOMP, RRT-Connect and LBKPIECE are stopped if a collision free trajectory is found before the max time (for GPMP and CHOMP collision checking is started after optimizing for at least 10 iterations). GPMP2 and TrajOpt are stopped when convergence is reached before the max time (we observed this was always the case) and feasibility is evaluated post-optimization.

\begin{figure}
\centering
\begin{subfigure}[b]{0.17\textwidth}
\centering
\includegraphics[width=1\linewidth]{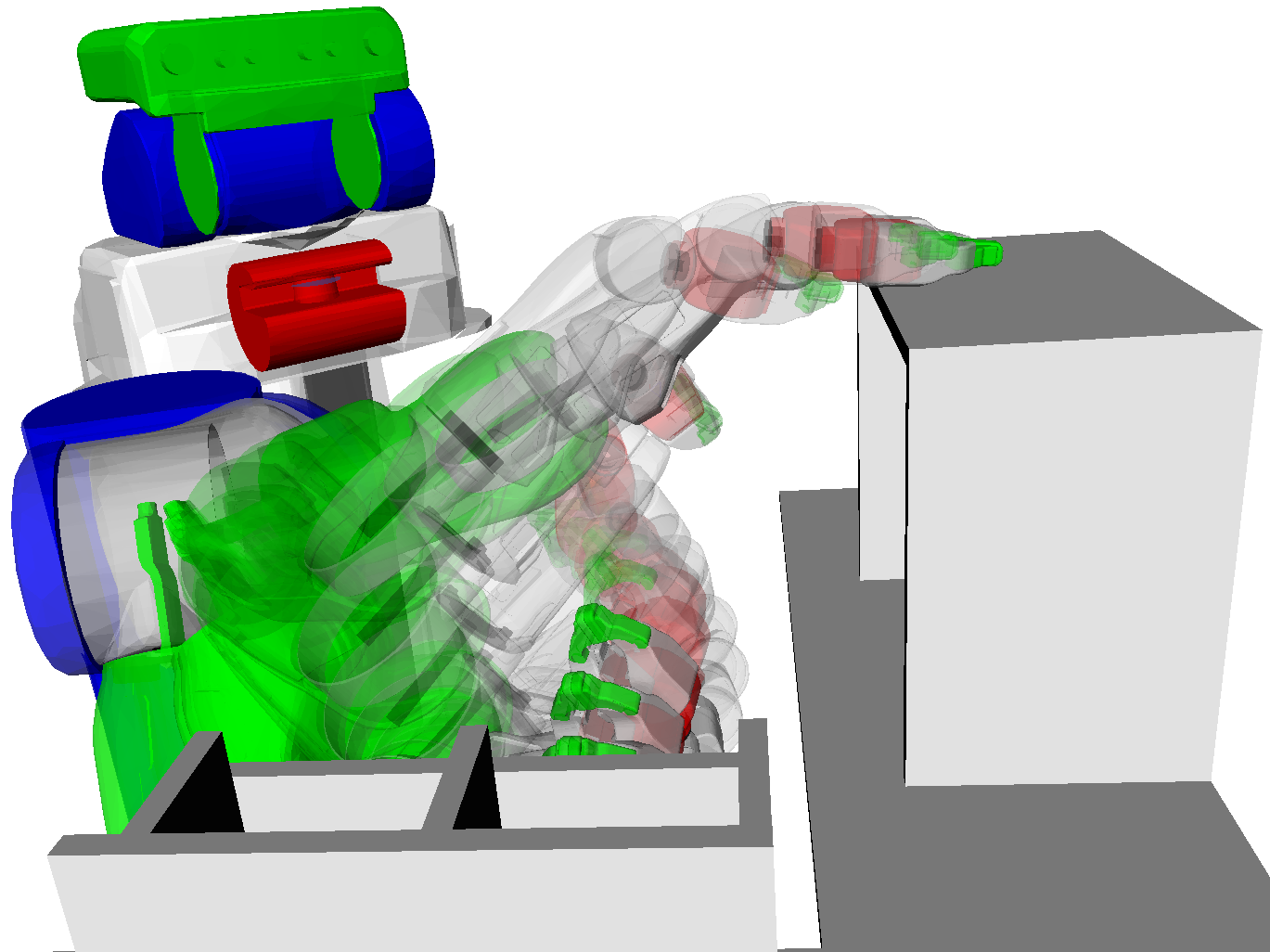}
\caption{$\sigma_{obs}=0.005$}
\end{subfigure}
\quad
\begin{subfigure}[b]{0.17\textwidth}
\centering
\includegraphics[width=1\linewidth]{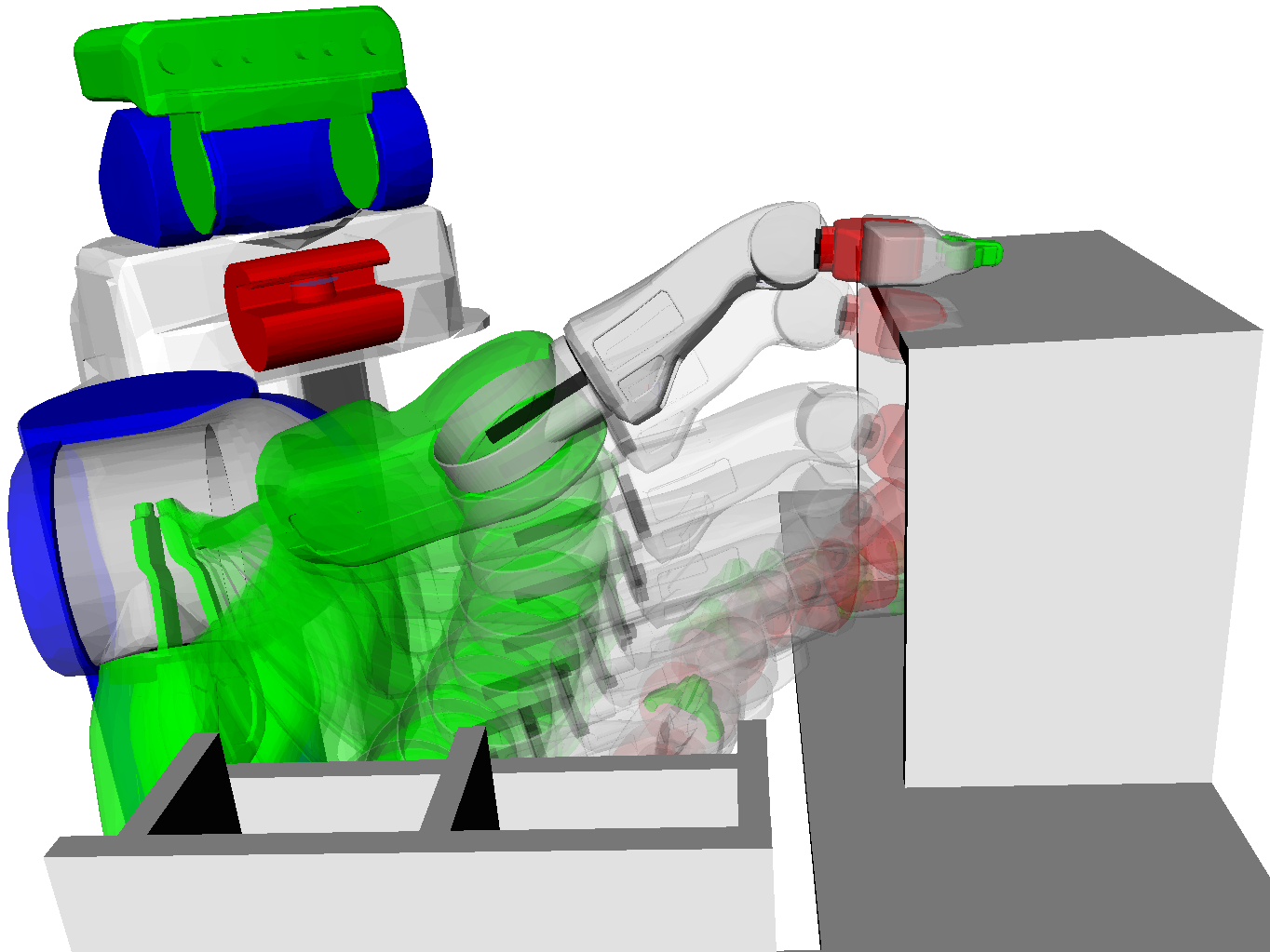}
\caption{$\sigma_{obs}=0.05$}
\end{subfigure}
\protect\caption{
(a) shows a successful trajectory with a good selection of $\sigma_{obs}$; (b) shows failure, where the trajectory collides with the top part of the shelf, when $\sigma_{obs}$ is too large.
\label{fig:success_fail_batch}}
\end{figure}

\begin{table*}[!t]
\begin{center}

\begin{subtable}{\textwidth}
\caption{Results for 24 planning problems on the 7-DOF WAM arm.}
\label{table:results_wam}
\centering
\scalebox{0.85}{
\begin{tabular}{lllllllll}
\toprule
& \bf{GPMP2-intp} & \bf{GPMP2-no-intp} & \bf{TrajOpt-101} & \bf{TrajOpt-11} & \bf{GPMP} & \bf{CHOMP} & \bf{RRT-Connect} & \bf{LBKPIECE} \\ 
\midrule
Success (\%)  & 91.7 & \bf{100.0} & 91.7 & 20.8 & 95.8 & 75 & 91.7 & 62.5 \\
Avg. Time (s) & 0.121 & 0.384 & 0.313 & \bf{0.027} & 0.3 & 0.695 & 1.87 & 6.89 \\
Max Time (s)  & 0.367 & 0.587 & 0.443 & \bf{0.033} & 0.554 & 2.868 & 5.18 & 9.97 \\
\bottomrule
\end{tabular}}
\end{subtable}

\par\bigskip

\begin{subtable}{\textwidth}
\caption{Results for 198 planning problems on PR2's 7-DOF right arm.}
\label{table:results_pr2}
\centering
\scalebox{0.85}{
\begin{tabular}{lllllllll}
\toprule
& \bf{GPMP2-intp} & \bf{GPMP2-no-intp} & \bf{TrajOpt-61 } & \bf{TrajOpt-11} & \bf{GPMP} & \bf{CHOMP} & \bf{RRT-Connect} & \bf{LBKPIECE} \\ 
\midrule
Success (\%) & 79.3 & 78.8 & 68.7 & 77.8 & 36.9 & 59.1 & \bf{82.3} & 33.8 \\
Avg. Time (s)  & \bf{0.11} & 0.196 & 0.958 & 0.191 & 1.7 & 2.38 & 3.02 & 7.12 \\
Max Time (s) & \bf{0.476} & 0.581 & 4.39 & 0.803 & 9.08 & 9.81 & 9.33 & 9.95 \\
\bottomrule
\end{tabular}}
\end{subtable}

\end{center}
\end{table*}

\subsubsection{Parameters:}
For both GPMP and GPMP2, $\mathbf{Q}_C$ controls the uncertainty in the prior distribution. A higher value means the trajectories will have a lower cost on deviating from the mean and the distribution covers a wider area of the configuration space. Thus a higher value is preferable in problems with more difficult navigation constraints. However, a very high value might result in noisy trajectories since the weight on the smoothness cost becomes relatively low. A reverse effect will be seen with a smaller value. This parameter can be set based on the problem and the prior model used (for example, constant velocity or constant acceleration). In our benchmarks, for GPMP we set $\mathbf{Q}_C = 100$ for the WAM dataset and $\mathbf{Q}_C = 50$ for the PR2 dataset and for GPMP2 we set $\mathbf{Q}_C = 1$ for both datasets.

Another common parameter, `safety distance,' $\epsilon$ is selected to be about double the minimum distance to any obstacle allowed in the scene and should be adjusted based on the robot, environment, and the obstacle cost function used. In our benchmarks we set $\epsilon = 0.2\text{m}$ for both GPMP and GPMP2 for the WAM dataset, and $\epsilon = 0.05\text{m}$ for GPMP and $\epsilon = 0.08\text{m}$ for GPMP2 for the PR2 dataset.

For GPMP2 the `obstacle cost weight' $\sigma_{obs}$ acts like a weight term that balances smoothness and collision-free requirements on the optimized trajectory and is  set based on the application. Smaller $\sigma_{obs}$ puts more weight on obstacle avoidance and vice versa. Figure \ref{fig:success_fail_batch} shows an example of an optimized trajectory for PR2 with different settings of $\sigma_{obs}$. In our experiments we found that the range $[0.001, 0.02]$ works well for $\sigma_{obs}$ and larger robot arms should use larger $\sigma_{obs}$. In the benchmarks we set $\sigma_{obs} = 0.02\text{m}$ for the WAM dataset and $\sigma_{obs} = 0.005$ for the PR2 dataset.

\subsubsection{Analysis:}

The benchmark results for the WAM dataset are summarized in Table \ref{table:results_wam}\footnote{Parameters for benchmark on the WAM dataset: For GPMP and CHOMP, $\lambda = 0.005$, $\eta = 1$. For CHOMP, $\epsilon = 0.2$. For TrajOpt, coeffs $= 20$, dist\_pen $= 0.05$.} and for the PR2 dataset are summarized in Table \ref{table:results_pr2}\footnote{Parameters for benchmark on the PR2 dataset: For CHOMP, $\epsilon = 0.05$. All remaining parameters are the same from the WAM dataset.}. Average time and maximum time include only successful runs.

Evaluating motion planning algorithms is a challenging task. The algorithms here use different techniques to formulate and solve the motion planning problem, and exhibit performance that depends on initial conditions as well as a range of parameter settings that can change based on the nature of the planning problem. Therefore, in our experiments we have tuned each algorithm to the settings close to default ones that worked best for each dataset. However, we still observe that TrajOpt-11 performs poorly on the WAM dataset (possibly due to using too few states on the trajectory) while GPMP performs poorly on the PR2 dataset (possibly due to the different initialization of the trajectory, and also the start and end configurations in the dataset being very close to the obstacles).

\begin{table*}[!t]
	\centering
	\caption{Average number of optimization iterations on successful runs.}
	\label{table:nr_iter}
	\begin{tabular}{llllll}
		\toprule
		& \bf{CHOMP}& \bf{GPMP-no-intp} & \bf{GPMP-intp} & \bf{GPMP2-no-intp} & \bf{GPMP2-intp} \\
		\midrule
		WAM & 26.4 & 11.5 & 12.0 & 23.6 & 13.0 \\
		PR2 & 46.6 & 32.2 & 19.1 & 26.4 & 24.4\\
		\bottomrule 
	\end{tabular}
\end{table*}

From the results in Table \ref{table:results_wam} and \ref{table:results_pr2} we see that GPMP2 perform consistently well compared to other algorithms on these datasets. Using interpolation during optimization (GPMP2-intp) achieves $30-50\%$ speedup of average and maximum runtime when compared to not using interpolation (GPMP2-no-intp). On the WAM dataset TrajOpt-11 has the lowest runtime but is able to solve only $20\%$ of the problems, while GPMP2-intp has the second lowest runtime with a much higher success rate. GPMP2-no-intp has the highest success rate. On the relatively harder PR2 dataset, GPMP2-intp has the lowest runtime and is twice as fast with a slightly higher success rate compared to TrajOpt-11. GPMP2-intp has the second highest success rate and is slightly behind RRT-Connect but is 30 times faster. The timing for RRT-Connect would further increase if a post processing or smoothing step was applied.

\begin{figure}[t]
	\begin{centering}
		\includegraphics[width=1\columnwidth]{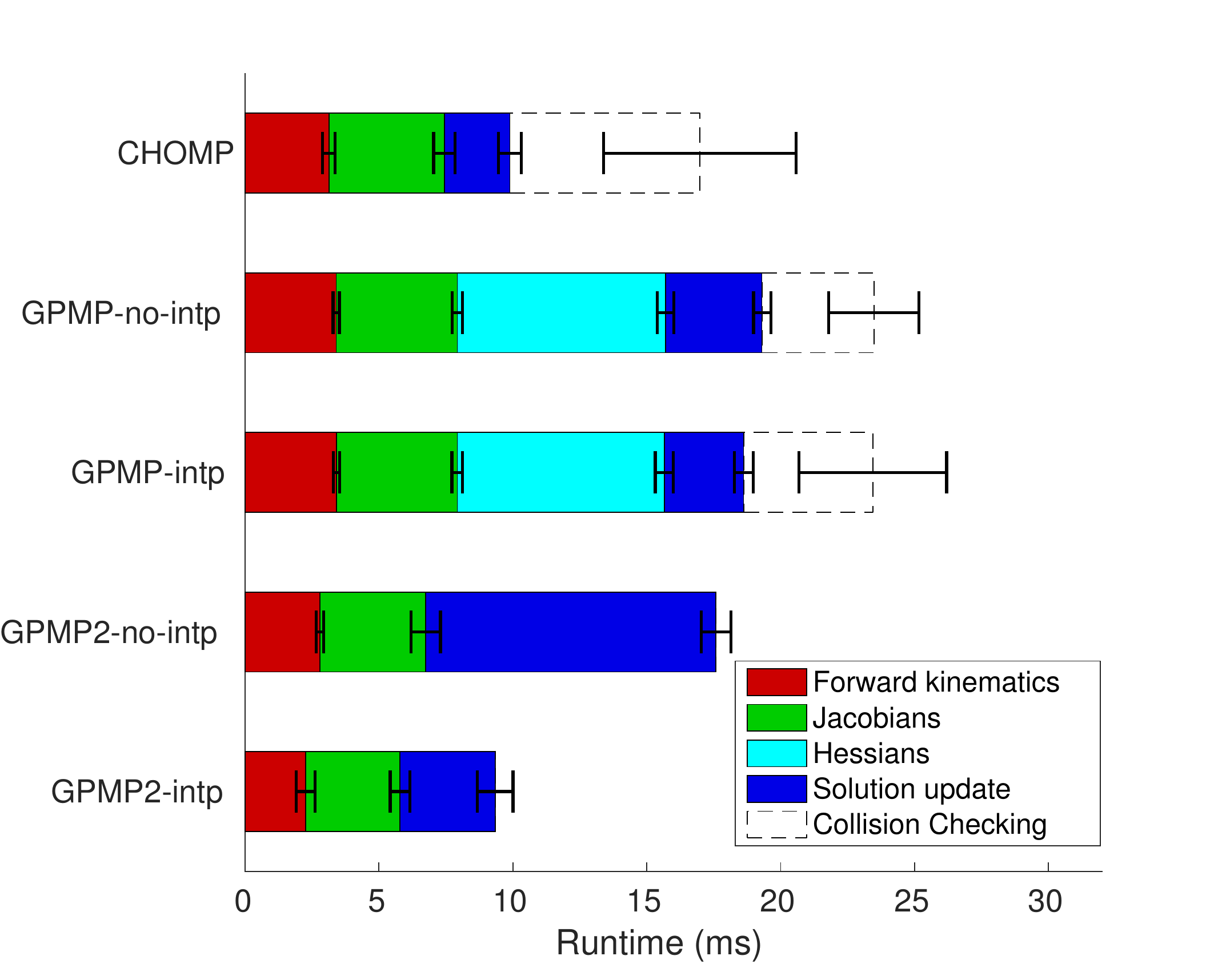}
	\end{centering}
	\caption{Breakdown of average timing per task {\bf per iteration} on all problems in the WAM dataset is shown for CHOMP, GPMP-no-intp, GPMP-intp, GPMP2-no-intp and GPMP2-intp. }
	\label{fig:timing}
\end{figure}

As seen from the max run times, GPMP2 always converges well before the maximum time limit and all the failure cases are due to infeasible local minima. Solutions like, random restarts (that are commonly employed) or GPMP-GRAPH~\citep{Huang-ICRA-17}, an extension to our approach that uses graph-based trajectories, can help contend with this issue.

To understand how the GP representation and the inference framework result in performance boost we compare timing breakdowns during any iteration for CHOMP, GPMP and GPMP2. Figure~\ref{fig:timing} shows the breakdown of average timing per task per iteration on the WAM dataset where the solution update portion (dark blue) incorporates the optimization costs. Table~\ref{table:nr_iter} shows average number of optimization iterations for successful runs in both the WAM and the PR2 datasets. We see that compared to CHOMP, GPMP is more expensive per iteration primarily from the computation of the Hessian, that is needed to find the acceleration in workspace (CHOMP approximates the acceleration with finite differencing). However, due to the GP representation and gradients on the augmented trajectory, GPMP is able to take larger update steps and hence converge faster with fewer iterations. GPMP2 on the other hand takes advantage of quadratic convergence while also benefiting from the GP representation and the inference framework. GP interpolation further reduces the runtime per iteration, especially for GPMP2. The dashed bars in Figure~\ref{fig:timing} represent computational costs due to collision checking during optimization at a finer resolution, on top of the computational cost incurred to evaluate gradient information. This was necessary to determine convergence, since the CHOMP solution can jump in and out of feasibility between iterations~\citep{zucker2013chomp}. GPMP also incurs this cost since it too exhibits this behavior due to its similar construction. Note that the total computational time in Table~\ref{table:results_wam} reflects the total iteration time as shown in Figure~\ref{fig:timing} plus time before and after the iterations including setup and communication time.

%------------------------------------------------------------------------------------------------------------------------------------------------%
\subsection{Incremental planning benchmark}\label{sec:evaluation_isam}

We evaluate our incremental motion planner iGPMP2 by benchmarking it against GPMP2 on replanning problems with the WAM and PR2 datasets.

For each problem in this benchmark, we have a planned trajectory from a start configuration to an originally assigned goal configuration. Then, at the middle time-step of the trajectory a new goal configuration is assigned. The replanning problem entails finding a trajectory to the newly assigned goal. This requires two changes to the factor graph: a new goal factor at the end of the trajectory to ensure that the trajectory reaches the new location in configuration, and a fixed state factor at the middle time step to enforce constraint of current state.

A total of 72 and 54 replanning problems are prepared for the WAM and the PR2 datasets, respectively. GP interpolation is used and all parameters are the same as the batch benchmarks. The benchmark results are shown in Table \ref{table:replan_wam} and Table \ref{table:replan_pr2}. 
We see from the results that iGPMP2 provides an order of magnitude speed-up, but suffers loss in success rate compared to GPMP2.

GPMP2 reinitializes the trajectory as a constant-velocity straight line from the middle state to the new goal and replans from scratch. However, iGPMP2 can use the solution to the old goal and the updated Bayes Tree as the initialization to incrementally update the trajectory, thus finding the solution much faster. 
There are three possible explanations why iGPMP2's success rate suffers as compared to the GPMP2's. First, iGPMP2 uses the original trajectory as initialization, which may be a poor choice if the goal has moved significantly. Second, in iSAM2 not every factor is relinearized and updated in Bayes tree for efficiency, which may lead to a poor linear approximation. Finally,  GPMP2 uses Levenberg-Marquardt for optimization that provides appropriate step damping, helping to improve the results, but iGPMP2 does not use similar step damping in its current implementation.
Since relinearizing factors or reinitializing variables will update the corresponding cliques of the Bayes tree and break its incremental nature, this results in runtime similar to batch optimization, and should not be done frequently. A good heuristic is to only perform relinearization/reinitialization when a planning failure is detected. We leave the task of designing a better solution to overcome this issue as future work.

\begin{figure}[!t]
	\begin{centering}
		{\includegraphics[width=0.48\columnwidth]{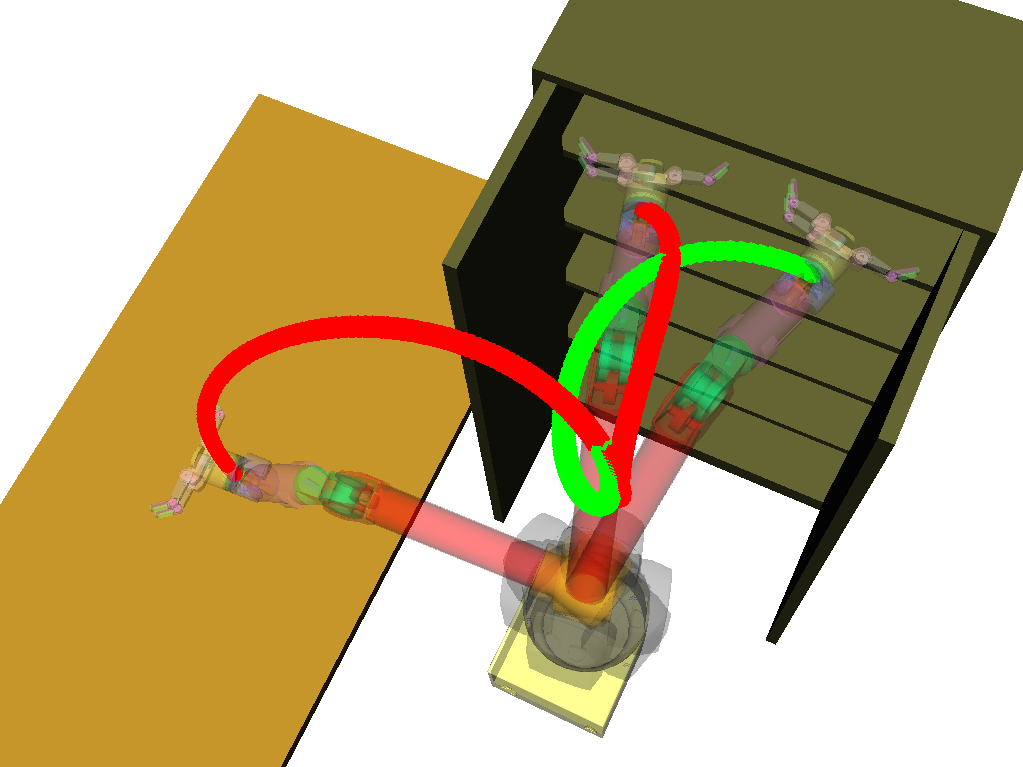}}
		\hfill
		{\includegraphics[width=0.48\columnwidth]{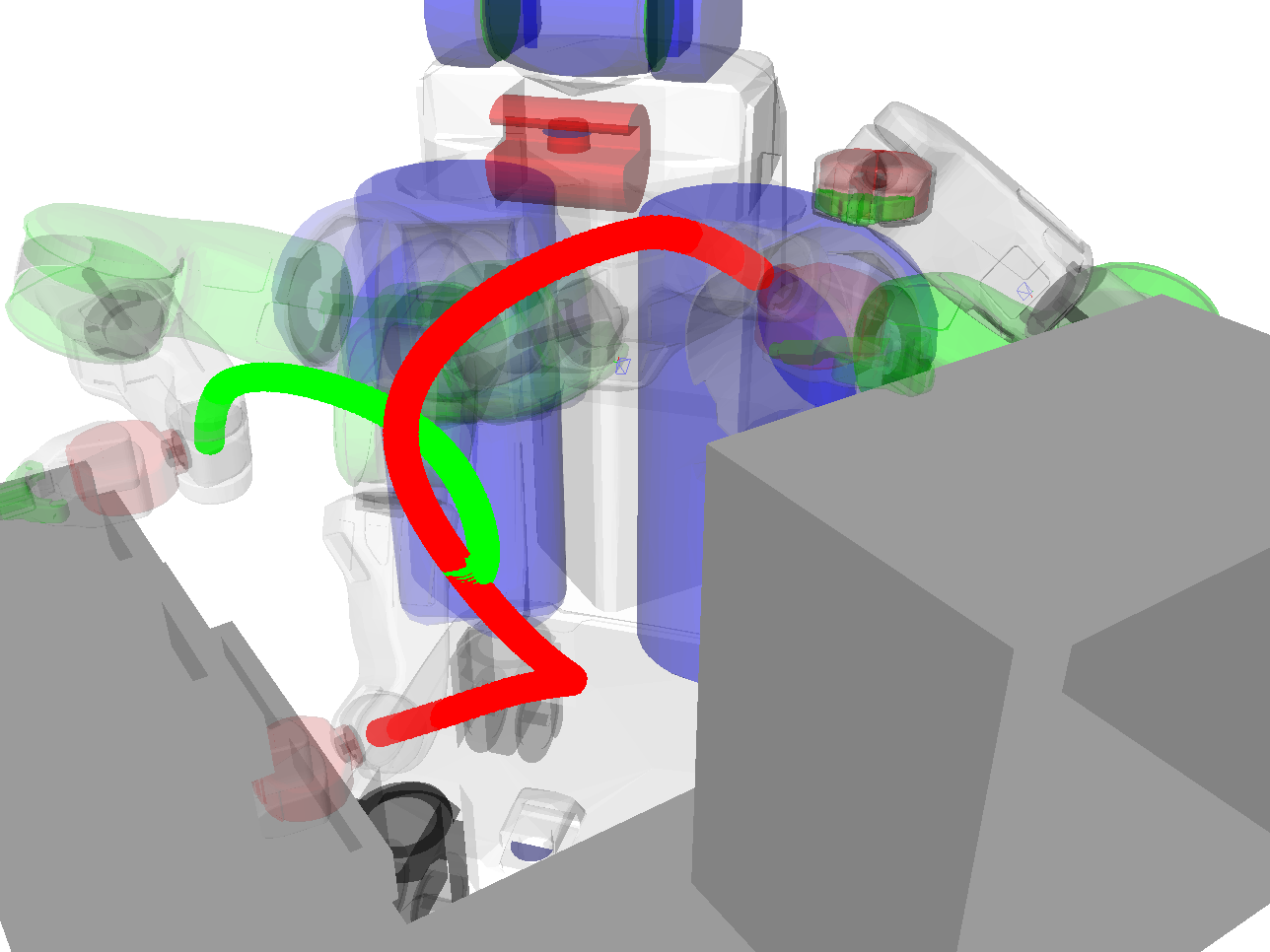}}
		\par\end{centering}
	\protect\caption{
		Example iGPMP2 results on the WAM and PR2 \textit{industrial}.
		Red lines show originally planned end-effector trajectories, 
		and green lines show replanned end-effector trajectories.
		Best viewed in color.
		\label{fig:replan_traj_examples}}\vspace{-4mm}
\end{figure}

\begin{table}[!t]
    \begin{subtable}{1\linewidth}
      \centering
        \caption{Results for 72 replanning problems on WAM.}
        \label{table:replan_wam}
    \begin{tabular}{lll}
	\toprule
	& \bf{iGPMP2} & \bf{GPMP2} \\
	\midrule
	Success (\%)      & \bf{100.0} & \bf{100.0} \\
	Avg. Time (ms) & \bf{8.07} & 65.68 \\
	Max Time (ms) & \bf{12.65} & 148.31 \\
	\bottomrule 
	\end{tabular}
    \end{subtable}
    
\par\bigskip    
    
    \begin{subtable}{1\linewidth}
      \centering
        \caption{Results for 54 replanning problems on PR2.}
        \label{table:replan_pr2}
    \begin{tabular}{lll}
	\toprule
	& \bf{iGPMP2} & \bf{GPMP2} \\
	\midrule
	Success (\%)      & 66.7 & \bf{88.9} \\
	Avg. Time (ms) & \bf{6.17} & 27.30 \\
	Max Time (ms) & \bf{7.37} & 87.95 \\
	\bottomrule 
	\end{tabular}
    \end{subtable} 
\end{table}

To maximize performance and overcome the deficiencies of iGPMP2, the rule of thumb when using iGPMP2 for replanning is to keep the difference between replanning problems and existing solutions to a minimum. This will lead to better initialization and reduced effect of linearization errors, and thus will improve iGPMP2's success rate. We verify this with the PR2 benchmark, where a smaller distance between original goal configuration and new goal configuration means a smaller difference between the replanning problem and the existing solution. We use $L2$ distance $\parallel \bm{\theta}_o - \bm{\theta}_r \parallel_2$ to quantify the distance between the original goal $\bm{\theta}_o$ and the new goal $\bm{\theta}_r$. From the PR2 benchmark, 27 problems have $\parallel \bm{\theta}_o - \bm{\theta}_r \parallel_2 < 2.0$, where iGPMP2 has 81.5\% success rate. On the other hand, we see that for the remaining 27 problems where $\parallel \bm{\theta}_o - \bm{\theta}_r \parallel_2 \geq 2.0$, iGPMP2 only has 51.9\% success rate.

Examples of successfully replanned trajectories generated using iGPMP2 are shown in Figure \ref{fig:replan_traj_examples}. The use of the fixed state factor at the middle time step helps make a smooth transition between original trajectories and replanned trajectories, which is critical if the trajectory is being executed on a real robot.

%%%%%%%%%%%%%%%%%%%%%%%%%%%%%%%%%%%%%%%%%%%%%%%%%%%%%%%%%%%%%%%%%%%%%%%%%%%%%%%%%%%%%%%%%%%%%%%%%%%%%%%%%%%%%%%%
\section{Discussion}\label{sec:discussion}

\subsection{Comparisons with related work}

GPMP can be viewed as a generalization on CHOMP where the trajectory is a sample from a GP and is augmented with velocities and accelerations. Both GPMP and GPMP2 use the GP representation for a continuous-time trajectory, GP interpolation, and signed distance fields for collision checking. However, with GPMP2 we fully embrace the probabilistic view of motion planning. In contrast to similar views on motion planning~\citep{toussaint2009robot,toussaint2006probabilistic} that use message passing, we instead solve the inference problem as nonlinear least squares. This allows us to use solvers with quadratic convergence rates that exploit the sparse structure of our problem, leading to a much faster algorithm compared to GPMP (and CHOMP) that only has linear convergence and is encumbered by the slow gradient computation. The update step in GPMP2 involves only linearization and the Cholesky decomposition to solve the linear system.

TrajOpt \citep{schulman2013finding,schulman2014motion} formulates the motion planning problem as constrained optimization, which allows the use of hard constraints on obstacles but also makes the optimization problem much more difficult and, as a consequence, slower to solve. Benchmark results in Section \ref{sec:evaluation_batch} show that our approach is faster than TrajOpt even when it uses a small number of states to represent the trajectory. TrajOpt performs continuous-time collision checking and can, therefore, solve problems with only a few states, in theory. However, the trajectory does not have a continuous-time representation and therefore must perform collision checking by approximating the convex-hull of obstacles and a straight line between states. This may not work in practice since a trajectory with few states would need to be post-processed to make it executable. Furthermore, depending on the post-processing method, collision-free guarantees may not exist for the final trajectory. Representing trajectories in continuous-time with GPs and using GP interpolation to up-sample them, allows our algorithms to circumvent this problem.

Unlike sampling based methods, our algorithms do not guarantee probabilistic completeness. However, from the benchmarks we see that GPMP2 is efficient at finding locally optimal trajectories that are feasible from na{\"i}ve straight line initialization that may be in collision. We note that trajectory optimization is prone to local minima and this strategy may not work on harder planning problems like mazes where sampling based methods excel. Recent work however, has begun to push the boundaries in trajectory optimization based planning. GPMP-GRAPH~\citep{Huang-ICRA-17}, an extension of our work, employs graph-based trajectories to explore exponential number of initializations simultaneously rather than trying them one at a time. Results show that it can quickly find feasible solutions even in mazes. Depending on the problem and time budget, multiple random initializations can also be a viable approach (since GPMP2 is fast), or GPMP2 can also be used on top of a path returned from a sampling based method to generate a time parameterized trajectory that is smooth.

Finally, our framework allows us to solve replanning problems very quickly, something that none of the above trajectory optimization approaches can provide. We are able to achieve this through incremental inference on a factor graph. On simpler replanning problems like changing goals, multi-query planners like PRM~\citep{kavraki1996probabilistic} can be useful but are time consuming since a large initial exploration of the space is necessary to build the first graph, a majority of which may not be needed. Solving these types of problems fast is very useful in real-time real-world applications.

%------------------------------------------------------------------------------------------------------------------------------------------------%
\subsection{Limitations \& future work}

A drawback of iterative methods for solving nonlinear least square problems is that they offer no global optimality guarantees. However, given that our objective is to satisfy \emph{smoothness} and to be \emph{collision-free}, a globally optimal solution is not strictly necessary. Many of the prior approaches to motion planning face similar issues of getting stuck in local minima. Random restarts is a commonly used method to combat this, however our approach allows for a more principled way~\citep{Huang-ICRA-17} in which this problem can be tackled.

The main drawback of our proposed approach is that it is limited in its ability to handle motion constraints like nonlinear inequality constraints. Sequential quadratic programming (SQP) can be used to solve problems with such constraints, and has been used before in motion planning~\citep{schulman2013finding,schulman2014motion}. We believe that SQP can be integrated into our trajectory optimizer, although this remains future work.

%%%%%%%%%%%%%%%%%%%%%%%%%%%%%%%%%%%%%%%%%%%%%%%%%%%%%%%%%%%%%%%%%%%%%%%%%%%%%%%%%%%%%%%%%%%%%%%%%%%%%%%%%%%%%%%%

\section{Conclusion}\label{sec:conclusion}

We use Gaussian processes to reason about continuous-time trajectories in the context of motion planning as trajectory optimization. Using GP interpolation we can query the trajectory at any time of interest such that the initial trajectory can be parameterized by only a few support states. The up-sampled trajectory is used during optimization to propagate the cost information back to the support states such that only they are updated. By formulating motion planning as probabilistic inference on factor graphs we also perform fast structure exploiting nonlinear least square optimization.

We benchmark our algorithms against several state-of-the-art trajectory optimization and sampling based algorithms on 7-DOF arm planning problems on two datasets in multiple environments and show that our approach, GPMP2 is consistently faster, often several times faster, than its nearest competitors.

Finally, by performing incremental inference on factor graphs we solve replanning problems with iGPMP2 incrementally in an order of magnitude faster than resolving from scratch with GPMP2. This property is unique to our motion planning algorithm and highly useful for planning in real-time real-world applications.

%%%%%%%%%%%%%%%%%%%%%%%%%%%%%%%%%%%%%%%%%%%%%%%%%%%%%%%%%%%%%%%%%%%%%%%%%%%%%%%%%%%%%%%%%%%%%%%%%%%%%%%%%%%%%%%%

\begin{acks}
This work was partially supported by National Institute of Food and Agriculture, U.S. Department of Agriculture, under award number 2014-67021-22556, NSF NRI award number 1637758, and NSF CRII award number 1464219.
\end{acks}

%%%%%%%%%%%%%%%%%%%%%%%%%%%%%%%%%%%%%%%%%%%%%%%%%%%%%%%%%%%%%%%%%%%%%%%%%%%%%%%%%%%%%%%%%%%%%%%%%%%%%%%%%%%%%%%%

\bibliographystyle{SageH}
\bibliography{ref}

\begin{thebibliography}{69}
\providecommand{\natexlab}[1]{#1}
\providecommand{\url}[1]{\texttt{#1}}
\providecommand{\urlprefix}{URL }
\expandafter\ifx\csname urlstyle\endcsname\relax
  \providecommand{\doi}[1]{DOI:\discretionary{}{}{}#1}\else
  \providecommand{\doi}{DOI:\discretionary{}{}{}\begingroup
  \urlstyle{rm}\Url}\fi

\bibitem[{Anderson and Barfoot(2013)}]{Anderson13icra}
Anderson S and Barfoot TD (2013) Towards relative continuous-time {SLAM}.
\newblock In: \emph{IEEE Intl. Conf. on Robotics and Automation (ICRA)}. IEEE.

\bibitem[{Anderson et~al.(2015)Anderson, Barfoot, Tong and
  S{\"a}rkk{\"a}}]{anderson2015batch}
Anderson S, Barfoot TD, Tong CH and S{\"a}rkk{\"a} S (2015) Batch nonlinear
  continuous-time trajectory estimation as exactly sparse gaussian process
  regression.
\newblock \emph{Autonomous Robots} 39(3): 221--238.

\bibitem[{Anderson et~al.(2014)Anderson, Dellaert and Barfoot}]{Anderson14icra}
Anderson S, Dellaert F and Barfoot T (2014) A hierarchical wavelet
  decomposition for continuous-time {SLAM}.
\newblock In: \emph{IEEE Intl. Conf. on Robotics and Automation (ICRA)}.

\bibitem[{Attias(2003)}]{attias2003planning}
Attias H (2003) Planning by probabilistic inference.
\newblock In: \emph{AISTATS}.

\bibitem[{Barfoot et~al.(2014)Barfoot, Tong and Sarkka}]{barfoot2014batch}
Barfoot T, Tong CH and Sarkka S (2014) Batch continuous-time trajectory
  estimation as exactly sparse {G}aussian process regression.
\newblock \emph{Proceedings of Robotics: Science and Systems, Berkeley, USA} .

\bibitem[{Bibby and Reid(2010)}]{Bibby10icra}
Bibby C and Reid I (2010) A hybrid {SLAM} representation for dynamic marine
  environments.
\newblock In: \emph{IEEE Intl. Conf. on Robotics and Automation (ICRA)}. IEEE,
  pp. 257--264.

\bibitem[{Bosse and Zlot(2009)}]{Bosse09icra}
Bosse M and Zlot R (2009) Continuous {3D} scan-matching with a spinning {2D}
  laser.
\newblock In: \emph{IEEE Intl. Conf. on Robotics and Automation (ICRA)}. IEEE,
  pp. 4312--4319.

\bibitem[{Byravan et~al.(2014)Byravan, Boots, Srinivasa and
  Fox}]{byravan2014space}
Byravan A, Boots B, Srinivasa SS and Fox D (2014) Space-time functional
  gradient optimization for motion planning.
\newblock In: \emph{Robotics and Automation (ICRA), 2014 IEEE International
  Conference on}. IEEE, pp. 6499--6506.

\bibitem[{Courant and Hilbert(1966)}]{courant1966methods}
Courant R and Hilbert D (1966) \emph{Methods of mathematical physics},
  volume~1.
\newblock CUP Archive.

\bibitem[{Cowell et~al.(2006)Cowell, Dawid, Lauritzen and
  Spiegelhalter}]{cowell2006probabilistic}
Cowell RG, Dawid P, Lauritzen SL and Spiegelhalter DJ (2006)
  \emph{Probabilistic networks and expert systems: Exact computational methods
  for Bayesian networks}.
\newblock Springer Science \& Business Media.

\bibitem[{Deisenroth and Rasmussen(2011)}]{deisenroth2011pilco}
Deisenroth M and Rasmussen CE (2011) {PILCO}: A model-based and data-efficient
  approach to policy search.
\newblock In: \emph{Proceedings of the 28th International Conference on machine
  learning (ICML-11)}. pp. 465--472.

\bibitem[{Dellaert(2012)}]{frank2012factor}
Dellaert F (2012) Factor graphs and {GTSAM}: a hands-on introduction.
\newblock Technical report, Georgia Tech Technical Report,
  GT-RIM-CP\&R-2012-002.

\bibitem[{Dellaert and Kaess(2006)}]{dellaert2006square}
Dellaert F and Kaess M (2006) Square root {SAM}: Simultaneous localization and
  mapping via square root information smoothing.
\newblock \emph{The International Journal of Robotics Research} 25(12):
  1181--1203.

\bibitem[{Dong and Barfoot(2014)}]{Dong2014fsr}
Dong H and Barfoot TD (2014) Lighting-invariant visual odometry using lidar
  intensity imagery and pose interpolation.
\newblock In: \emph{Field and Service Robotics (FSR)}. Springer, pp. 327--342.

\bibitem[{Dong et~al.(2016)Dong, Mukadam, Dellaert and Boots}]{Dong-RSS-16}
Dong J, Mukadam M, Dellaert F and Boots B (2016) Motion planning as
  probabilistic inference using {G}aussian processes and factor graphs.
\newblock In: \emph{Proceedings of Robotics: Science and Systems (RSS)}.

\bibitem[{Elbanhawi et~al.(2015)Elbanhawi, Simic and Jazar}]{Elbanhawi15itits}
Elbanhawi M, Simic M and Jazar R (2015) Randomized bidirectional {B}-{S}pline
  parameterization motion planning.
\newblock \emph{Intelligent Transportation Systems, IEEE Transactions on}
  PP(99): 1--1.

\bibitem[{Furgale et~al.(2013)Furgale, Rehder and Siegwart}]{Furgale13iros}
Furgale P, Rehder J and Siegwart R (2013) Unified temporal and spatial
  calibration for multi-sensor systems.
\newblock In: \emph{IEEE/RSJ Intl. Conf. on Intelligent Robots and Systems
  (IROS)}. IEEE.

\bibitem[{Furgale et~al.(2015)Furgale, Tong, Barfoot and
  Sibley}]{Furgale15ijrr}
Furgale P, Tong CH, Barfoot TD and Sibley G (2015) Continuous-time batch
  trajectory estimation using temporal basis functions.
\newblock \emph{Intl. J. of Robotics Research} 34(14): 1688--1710.

\bibitem[{Gammell et~al.(2015)Gammell, Srinivasa and
  Barfoot}]{gammell2015batch}
Gammell JD, Srinivasa SS and Barfoot TD (2015) Batch informed trees (bit*):
  Sampling-based optimal planning via the heuristically guided search of
  implicit random geometric graphs.
\newblock In: \emph{2015 IEEE International Conference on Robotics and
  Automation (ICRA)}. IEEE, pp. 3067--3074.

\bibitem[{Golub and Van~Loan(2012)}]{golub2012matrix}
Golub GH and Van~Loan CF (2012) \emph{Matrix computations}, volume~3.
\newblock JHU Press.

\bibitem[{He et~al.(2013)He, Martin and Zucker}]{he2013multigrid}
He K, Martin E and Zucker M (2013) Multigrid {CHOMP} with local smoothing.
\newblock In: \emph{Proc. of 13th IEEE-RAS Int. Conference on Humanoid Robots
  (Humanoids)}.

\bibitem[{Huang et~al.(2017)Huang, Mukadam, Liu and Boots}]{Huang-ICRA-17}
Huang E, Mukadam M, Liu Z and Boots B (2017) Motion planning with graph-based
  trajectories and {G}aussian process inference.
\newblock In: \emph{Proceedings of the 2017 IEEE Conference on Robotics and
  Automation (ICRA)}.

\bibitem[{Kaess et~al.(2011{\natexlab{a}})Kaess, Ila, Roberts and
  Dellaert}]{kaess2011bayes}
Kaess M, Ila V, Roberts R and Dellaert F (2011{\natexlab{a}}) The {B}ayes tree:
  An algorithmic foundation for probabilistic robot mapping.
\newblock In: \emph{Algorithmic Foundations of Robotics IX}. Springer, pp.
  157--173.

\bibitem[{Kaess et~al.(2011{\natexlab{b}})Kaess, Johannsson, Roberts, Ila,
  Leonard and Dellaert}]{kaess2011isam2}
Kaess M, Johannsson H, Roberts R, Ila V, Leonard JJ and Dellaert F
  (2011{\natexlab{b}}) i{SAM}2: Incremental smoothing and mapping using the
  {B}ayes tree.
\newblock \emph{The International Journal of Robotics Research} :
  0278364911430419.

\bibitem[{Kaess et~al.(2008)Kaess, Ranganathan and Dellaert}]{kaess2008isam}
Kaess M, Ranganathan A and Dellaert F (2008) i{SAM}: Incremental smoothing and
  mapping.
\newblock \emph{Robotics, IEEE Transactions on} 24(6): 1365--1378.

\bibitem[{Kalakrishnan et~al.(2011)Kalakrishnan, Chitta, Theodorou, Pastor and
  Schaal}]{kalakrishnan2011stomp}
Kalakrishnan M, Chitta S, Theodorou E, Pastor P and Schaal S (2011) {STOMP}:
  Stochastic trajectory optimization for motion planning.
\newblock In: \emph{Robotics and Automation (ICRA), 2011 IEEE International
  Conference on}. IEEE, pp. 4569--4574.

\bibitem[{Karaman and Frazzoli(2010)}]{karaman2010incremental}
Karaman S and Frazzoli E (2010) Incremental sampling-based algorithms for
  optimal motion planning.
\newblock \emph{Robotics Science and Systems VI} 104.

\bibitem[{Kavraki et~al.(1996)Kavraki, Svestka, Latombe and
  Overmars}]{kavraki1996probabilistic}
Kavraki LE, Svestka P, Latombe JC and Overmars MH (1996) Probabilistic roadmaps
  for path planning in high-dimensional configuration spaces.
\newblock \emph{Robotics and Automation, IEEE Transactions on} 12(4): 566--580.

\bibitem[{Kersting et~al.(2007)Kersting, Plagemann, Pfaff and
  Burgard}]{kersting2007most}
Kersting K, Plagemann C, Pfaff P and Burgard W (2007) Most likely
  heteroscedastic {G}aussian process regression.
\newblock In: \emph{Proceedings of the 24th international conference on Machine
  learning}. ACM, pp. 393--400.

\bibitem[{Ko and Fox(2009)}]{ko2009gp}
Ko J and Fox D (2009) {GP}-{B}ayes{F}ilters: {B}ayesian filtering using
  {G}aussian process prediction and observation models.
\newblock \emph{Autonomous Robots} 27(1): 75--90.

\bibitem[{Koenig et~al.(2003)Koenig, Tovey and Smirnov}]{koenig2003performance}
Koenig S, Tovey C and Smirnov Y (2003) Performance bounds for planning in
  unknown terrain.
\newblock \emph{Artificial Intelligence} 147(1): 253--279.

\bibitem[{Kschischang et~al.(2001)Kschischang, Frey and
  Loeliger}]{kschischang2001factor}
Kschischang FR, Frey BJ and Loeliger HA (2001) Factor graphs and the
  sum-product algorithm.
\newblock \emph{Information Theory, IEEE Transactions on} 47(2): 498--519.

\bibitem[{Kuffner and LaValle(2000)}]{kuffner2000rrt}
Kuffner JJ and LaValle SM (2000) {RRT}-connect: An efficient approach to
  single-query path planning.
\newblock In: \emph{Robotics and Automation, 2000. Proceedings. ICRA'00. IEEE
  International Conference on}, volume~2. IEEE, pp. 995--1001.

\bibitem[{LaValle(2006)}]{lavalle2006planning}
LaValle SM (2006) \emph{Planning algorithms}.
\newblock Cambridge university press.

\bibitem[{Leutenegger et~al.(2015)Leutenegger, Lynen, Bosse, Siegwart and
  Furgale}]{Leutenegger15ijrr}
Leutenegger S, Lynen S, Bosse M, Siegwart R and Furgale P (2015) Keyframe-based
  visual--inertial odometry using nonlinear optimization.
\newblock \emph{Intl. J. of Robotics Research} 34(3): 314--334.

\bibitem[{Levine and Koltun(2013)}]{levine2013variational}
Levine S and Koltun V (2013) Variational policy search via trajectory
  optimization.
\newblock In: \emph{Advances in Neural Information Processing Systems}. pp.
  207--215.

\bibitem[{Li et~al.(2013)Li, Kim and Mourikis}]{Li13icra}
Li M, Kim BH and Mourikis AI (2013) Real-time motion tracking on a cellphone
  using inertial sensing and a rolling-shutter camera.
\newblock In: \emph{IEEE Intl. Conf. on Robotics and Automation (ICRA)}. IEEE.

\bibitem[{Likhachev et~al.(2005)Likhachev, Ferguson, Gordon, Stentz and
  Thrun}]{likhachev2005anytime}
Likhachev M, Ferguson DI, Gordon GJ, Stentz A and Thrun S (2005) Anytime
  dynamic {A}*: An anytime, replanning algorithm.
\newblock In: \emph{ICAPS}. pp. 262--271.

\bibitem[{Marinho et~al.(2016)Marinho, Dragan, Byravan, Boots, Gordon and
  Srinivasa}]{Marinho-RSS-16}
Marinho Z, Dragan A, Byravan A, Boots B, Gordon GJ and Srinivasa S (2016)
  Functional gradient motion planning in reproducing kernel {H}ilbert spaces.
\newblock In: \emph{Proceedings of Robotics: Science and Systems (RSS)}.

\bibitem[{Mukadam(2017)}]{mukadam2017piper}
Mukadam M (2017) {PIPER}.
\newblock \emph{[Online] Available at \url{https://github.com/gtrll/piper}} .

\bibitem[{Mukadam et~al.(2017{\natexlab{a}})Mukadam, Cheng, Yan and
  Boots}]{Mukadam-ICRA-17}
Mukadam M, Cheng CA, Yan X and Boots B (2017{\natexlab{a}}) Approximately
  optimal continuous-time motion planning and control via probabilistic
  inference.
\newblock In: \emph{Proceedings of the 2017 IEEE Conference on Robotics and
  Automation (ICRA)}.

\bibitem[{Mukadam et~al.(2017{\natexlab{b}})Mukadam, Dong, Dellaert and
  Boots}]{Mukadam-RSS-17}
Mukadam M, Dong J, Dellaert F and Boots B (2017{\natexlab{b}}) Simultaneous
  trajectory estimation and planning via probabilistic inference.
\newblock In: \emph{Proceedings of Robotics: Science and Systems (RSS)}.

\bibitem[{Mukadam et~al.(2018)Mukadam, Dong, Dellaert and
  Boots}]{Mukadam-AURO-18}
Mukadam M, Dong J, Dellaert F and Boots B (2018) {STEAP}: simultaneous
  trajectory estimation and planning.
\newblock In: \emph{Autonomous Robots (AURO)}.

\bibitem[{Mukadam et~al.(2016)Mukadam, Yan and Boots}]{Mukadam-ICRA-16}
Mukadam M, Yan X and Boots B (2016) {G}aussian process motion planning.
\newblock In: \emph{2016 IEEE International Conference on Robotics and
  Automation (ICRA)}. pp. 9--15.

\bibitem[{Nguyen-Tuong et~al.(2008)Nguyen-Tuong, Peters, Seeger and
  Sch{\"o}lkopf}]{nguyen2008learning}
Nguyen-Tuong D, Peters J, Seeger M and Sch{\"o}lkopf B (2008) Learning inverse
  dynamics: a comparison.
\newblock In: \emph{European Symposium on Artificial Neural Networks},
  EPFL-CONF-175477.

\bibitem[{Park et~al.(2012)Park, Pan and Manocha}]{park2012itomp}
Park C, Pan J and Manocha D (2012) {ITOMP}: Incremental trajectory optimization
  for real-time replanning in dynamic environments.
\newblock In: \emph{ICAPS}.

\bibitem[{Park et~al.(2013)Park, Pan and Manocha}]{park2013real}
Park C, Pan J and Manocha D (2013) Real-time optimization-based planning in
  dynamic environments using {GPU}s.
\newblock In: \emph{Robotics and Automation (ICRA), 2013 IEEE International
  Conference on}. IEEE, pp. 4090--4097.

\bibitem[{Patron-Perez et~al.(2015)Patron-Perez, Lovegrove and
  Sibley}]{Patron15ijcv}
Patron-Perez A, Lovegrove S and Sibley G (2015) A spline-based trajectory
  representation for sensor fusion and rolling shutter cameras.
\newblock \emph{International Journal of Computer Vision} 113(3): 208--219.

\bibitem[{Quinlan(1994)}]{quinlan1994real}
Quinlan S (1994) \emph{Real-time modification of collision-free paths}.
\newblock PhD Thesis, Stanford University.

\bibitem[{Rana et~al.(2017)Rana, Mukadam, Ahmadzadeh, Chernova and
  Boots}]{pmlr-v78-rana17a}
Rana MA, Mukadam M, Ahmadzadeh SR, Chernova S and Boots B (2017) Towards robust
  skill generalization: Unifying learning from demonstration and motion
  planning.
\newblock In: \emph{Proceedings of the 1st Annual Conference on Robot
  Learning}, volume~78. PMLR, pp. 109--118.

\bibitem[{Rana et~al.(2018)Rana, Mukadam, Ahmadzadeh, Chernova and
  Boots}]{Rana-IROS-18}
Rana MA, Mukadam M, Ahmadzadeh SR, Chernova S and Boots B (2018) Learning
  generalizable robot skills from demonstrations in cluttered environments.
\newblock In: \emph{Proceedings of the International Conference on Intelligent
  Robots and Systems (IROS)}.

\bibitem[{Rasmussen(2006)}]{rasmussen2006gaussian}
Rasmussen CE (2006) \emph{Gaussian processes for machine learning}.
\newblock Citeseer.

\bibitem[{Ratliff et~al.(2009)Ratliff, Zucker, Bagnell and
  Srinivasa}]{ratliff2009chomp}
Ratliff N, Zucker M, Bagnell JA and Srinivasa S (2009) {CHOMP}: Gradient
  optimization techniques for efficient motion planning.
\newblock In: \emph{Robotics and Automation, 2009. ICRA'09. IEEE International
  Conference on}. IEEE, pp. 489--494.

\bibitem[{Rawlik et~al.(2012)Rawlik, Toussaint and
  Vijayakumar}]{rawlik2012stochastic}
Rawlik K, Toussaint M and Vijayakumar S (2012) On stochastic optimal control
  and reinforcement learning by approximate inference.
\newblock \emph{Proceedings of Robotics: Science and Systems} .

\bibitem[{Sarkka et~al.(2013)Sarkka, Solin and
  Hartikainen}]{sarkka2013spatiotemporal}
Sarkka S, Solin A and Hartikainen J (2013) Spatiotemporal learning via
  infinite-dimensional {B}ayesian filtering and smoothing: A look at {G}aussian
  process regression through {K}alman filtering.
\newblock \emph{IEEE Signal Processing Magazine} 30(4): 51--61.

\bibitem[{Schulman et~al.(2014)Schulman, Duan, Ho, Lee, Awwal, Bradlow, Pan,
  Patil, Goldberg and Abbeel}]{schulman2014motion}
Schulman J, Duan Y, Ho J, Lee A, Awwal I, Bradlow H, Pan J, Patil S, Goldberg K
  and Abbeel P (2014) Motion planning with sequential convex optimization and
  convex collision checking.
\newblock \emph{The International Journal of Robotics Research} 33(9):
  1251--1270.

\bibitem[{Schulman et~al.(2013)Schulman, Ho, Lee, Awwal, Bradlow and
  Abbeel}]{schulman2013finding}
Schulman J, Ho J, Lee A, Awwal I, Bradlow H and Abbeel P (2013) Finding locally
  optimal, collision-free trajectories with sequential convex optimization.
\newblock In: \emph{Robotics: Science and Systems}, volume~9. Citeseer, pp.
  1--10.

\bibitem[{Sturm et~al.(2009)Sturm, Plagemann and Burgard}]{sturm2009body}
Sturm J, Plagemann C and Burgard W (2009) Body schema learning for robotic
  manipulators from visual self-perception.
\newblock \emph{Journal of Physiology-Paris} 103(3): 220--231.

\bibitem[{{\c{S}}ucan and Kavraki(2009)}]{csucan2009kinodynamic}
{\c{S}}ucan IA and Kavraki LE (2009) Kinodynamic motion planning by
  interior-exterior cell exploration.
\newblock In: \emph{Algorithmic Foundation of Robotics VIII}. Springer, pp.
  449--464.

\bibitem[{Sucan et~al.(2012)Sucan, Moll and Kavraki}]{sucan2012open}
Sucan IA, Moll M and Kavraki LE (2012) The open motion planning library.
\newblock \emph{IEEE Robotics \& Automation Magazine} 19(4): 72--82.

\bibitem[{Tay and Laugier(2008)}]{tay2008modelling}
Tay MKC and Laugier C (2008) Modelling smooth paths using {G}aussian processes.
\newblock In: \emph{Field and Service Robotics}. Springer, pp. 381--390.

\bibitem[{Theodorou et~al.(2010)Theodorou, Tassa and
  Todorov}]{theodorou2010stochastic}
Theodorou E, Tassa Y and Todorov E (2010) Stochastic differential dynamic
  programming.
\newblock In: \emph{American Control Conference (ACC), 2010}. IEEE, pp.
  1125--1132.

\bibitem[{Tong et~al.(2012)Tong, Furgale and Barfoot}]{tong2012gaussian}
Tong CH, Furgale P and Barfoot TD (2012) {G}aussian process {G}auss-{N}ewton:
  Non-parametric state estimation.
\newblock In: \emph{Computer and Robot Vision (CRV), 2012 Ninth Conference on}.
  IEEE, pp. 206--213.

\bibitem[{Toussaint(2009)}]{toussaint2009robot}
Toussaint M (2009) Robot trajectory optimization using approximate inference.
\newblock In: \emph{Proceedings of the 26th annual international conference on
  machine learning}. ACM, pp. 1049--1056.

\bibitem[{Toussaint and Goerick(2010)}]{toussaint2010bayesian}
Toussaint M and Goerick C (2010) A {B}ayesian view on motor control and
  planning.
\newblock In: \emph{From Motor Learning to Interaction Learning in Robots}.
  Springer, pp. 227--252.

\bibitem[{Toussaint and Storkey(2006)}]{toussaint2006probabilistic}
Toussaint M and Storkey A (2006) Probabilistic inference for solving discrete
  and continuous state {M}arkov decision processes.
\newblock In: \emph{Proceedings of the 23rd international conference on Machine
  learning}. ACM, pp. 945--952.

\bibitem[{Vijayakumar et~al.(2005)Vijayakumar, D'souza and
  Schaal}]{vijayakumar2005incremental}
Vijayakumar S, D'souza A and Schaal S (2005) Incremental online learning in
  high dimensions.
\newblock \emph{Neural computation} 17(12): 2602--2634.

\bibitem[{Yan et~al.(2017)Yan, Indelman and Boots}]{Yan17ras}
Yan X, Indelman V and Boots B (2017) Incremental sparse {GP} regression for
  continuous-time trajectory estimation and mapping.
\newblock In: \emph{Robotics and Autonomous Systems}, volume~87. pp. 120--132.

\bibitem[{Zucker et~al.(2013)Zucker, Ratliff, Dragan, Pivtoraiko, Klingensmith,
  Dellin, Bagnell and Srinivasa}]{zucker2013chomp}
Zucker M, Ratliff N, Dragan AD, Pivtoraiko M, Klingensmith M, Dellin CM,
  Bagnell JA and Srinivasa SS (2013) {CHOMP}: Covariant {H}amiltonian
  optimization for motion planning.
\newblock \emph{The International Journal of Robotics Research} 32(9-10):
  1164--1193.

\end{thebibliography}

%%%%%%%%%%%%%%%%%%%%%%%%%%%%%%%%%%%%%%%%%%%%%%%%%%%%%%%%%%%%%%%%%%%%%%%%%%%%%%%%%%%%%%%%%%%%%%%%%%%%%%%%%%%%%%%%
\clearpage

\section*{Appendix A: The trajectory prior}

First, we review conditioning a distribution of state $\bm \theta$ on observations $\bm Y$ in general (for a full treatment see~\citep{rasmussen2006gaussian}). Let the observation be given by the following linear equation
\begin{equation}
	\bm Y = \mathbf{C}\bm{\theta} + \bm \epsilon, \quad \bm \epsilon \sim \mathcal{N}(\bm 0, \widetilde{\bm{\mathcal{K}}}_y).
\end{equation}
We can write their joint distribution as
\begin{equation}
	\mathcal{N} \Bigg( \left[ \begin{matrix} \widetilde{\bm \mu} \\ \mathbf{C}\widetilde{\bm \mu} \end{matrix} \right],
	\left[ \begin{matrix} \widetilde{\bm{\mathcal{K}}} & \widetilde{\bm{\mathcal{K}}}\mathbf{C}^\top \\ \mathbf{C}\widetilde{\bm{\mathcal{K}}} & \mathbf{C}\widetilde{\bm{\mathcal{K}}}\mathbf{C}^\top + \widetilde{\bm{\mathcal{K}}}_y \end{matrix}\right] \Bigg).
\end{equation}
The distribution of the state conditioned on the observations is then $\mathcal{N}(\bm \mu, \bm{\mathcal{K}})$ where
\begin{align}
	\bm \mu &= \widetilde{\bm \mu} + \widetilde{\bm{\mathcal{K}}}\mathbf{C}^\top (\mathbf{C}\widetilde{\bm{\mathcal{K}}}\mathbf{C}^\top + \widetilde{\bm{\mathcal{K}}}_y)^{-1}  (\bm Y - \mathbf{C}\widetilde{\bm \mu}) \\
	\bm{\mathcal{K}} &= \widetilde{\bm{\mathcal{K}}} - \widetilde{\bm{\mathcal{K}}}\mathbf{C}^\top (\mathbf{C}\widetilde{\bm{\mathcal{K}}}\mathbf{C}^\top + \widetilde{\bm{\mathcal{K}}}_y)^{-1}\mathbf{C}\widetilde{\bm{\mathcal{K}}} \label{eq:gen_prior_k}
\end{align}
Now, we are interested in conditioning just on the goal state $\bm\theta_N$ with mean $\bm{\mu}_N$ and covariance $\bm{\mathcal{K}}_N$. Therefore in the above equations we use $\mathbf{C} = [ \begin{matrix}\bm 0 & \hdots & \bm 0 & \mathbf{I} \end{matrix} ]$ and $\widetilde{\bm{\mathcal{K}}}_y = \bm{\mathcal{K}}_N$ to get
\begin{small}
\begin{align}
	{\bm{\mu}}&= \widetilde{\bm{\mu}} + \widetilde{\bm{\mathcal{K}}}(t_N,\bm t)^\top (\widetilde{\bm{\mathcal{K}}}(t_N,t_N) + \bm{\mathcal{K}}_N )^{-1} (\bm{\theta}_N - {\bm{\mu}}_N)\\
	{\bm{\mathcal{K}}}& = \widetilde{\bm{\mathcal{K}}} - \widetilde{\bm{\mathcal{K}}}(t_N,\bm t)^\top (\widetilde{\bm{\mathcal{K}}}(t_N,t_N) + \bm{\mathcal{K}}_N )^{-1} \widetilde{\bm{\mathcal{K}}}(t_N,\bm t)
\end{align}
\end{small}
where $\widetilde{\bm{\mathcal{K}}}(t_N,\bm t) = [ \begin{matrix} \widetilde{\bm{\mathcal{K}}}(t_N,t_0) & \dots & \widetilde{\bm{\mathcal{K}}}(t_N,t_N)] \end{matrix} ]$.

Using the Woodbury matrix identity we can write Eq.~\eqref{eq:gen_prior_k} as
\begin{equation}
	\bm{\mathcal{K}} = (\widetilde{\bm{\mathcal{K}}}^{-1} + \mathbf{C}^\top\widetilde{\bm{\mathcal{K}}}^{-1}_y\mathbf{C})^{-1}
\end{equation}
and substituting $\mathbf{C}$ and $\widetilde{\bm{\mathcal{K}}}_y$ as before for conditioning on the goal we get
\begin{small}
\begin{equation}
	\bm{\mathcal{K}} = \bigg(\widetilde{\bm{\mathcal{K}}}^{-1} + [ \begin{matrix}\bm 0 & \hdots & \bm 0 & \mathbf{I} \end{matrix} ]^\top \bm{\mathcal{K}}_N^{-1} [ \begin{matrix}\bm 0 & \hdots & \bm 0 & \mathbf{I} \end{matrix} ]\bigg)^{-1}.
\end{equation}
\end{small}%
From~\citep{barfoot2014batch} we know that the precision matrix of the distribution obtained from the LTV-SDE in Eq.~\eqref{eq:LVT-SDE} can be decomposed as $\widetilde{\bm{\mathcal{K}}}^{-1} = \widetilde{\bm{\mathbf{A}}}^{-\top}\widetilde{\bm{\mathbf{Q}}}^{-1}\widetilde{\bm{\mathbf{A}}}^{-1}$. Therefore,
\begin{align}
	\bm{\mathcal{K}}^{-1} &= \small{\left[ \begin{matrix} \widetilde{\bm{\mathbf{A}}}^{-1} \\ \begin{matrix} \bm 0 & \hdots & \bm 0 & \mathbf{I}	\end{matrix} \end{matrix} \right]^\top \left[ \begin{matrix} \widetilde{\bm{\mathbf{Q}}}^{-1} & \\ & \bm{\mathcal{K}}_N^{-1} \end{matrix} \right] \left[ \begin{matrix} \widetilde{\bm{\mathbf{A}}}^{-1} \\ \begin{matrix} \bm 0 & \hdots & \bm 0 & \mathbf{I}	\end{matrix} \end{matrix} \right]}\\
	&= \mathbf B^\top \mathbf Q^{-1} \mathbf B
\end{align}%
where
\begin{equation}
\footnotesize
	\mathbf B = \left[ \begin{matrix}
	\mathbf{I} & \bm 0 & \dots & \bm 0 & \bm 0\\
	-\mathbf\Phi(t_1,t_0) & \mathbf{I} & \dots & \bm 0 & \bm 0\\
	\bm 0 & -\mathbf\Phi(t_2,t_1) & \ddots & \vdots & \vdots \\
	\vdots & \vdots & \ddots & \mathbf{I} & \bm 0 \\
	\bm 0 & \bm 0 & \dots &-\mathbf\Phi(t_N,t_{N-1}) & \mathbf{I} \\
	\bm 0 & \bm 0 & \dots & \bm 0 & \mathbf{I} \end{matrix} \right],
\end{equation}%
and
\begin{align}
	\mathbf Q^{-1} &= \operatorname{diag}(\bm{\mathcal{K}}_0^{-1}, \mathbf Q_{0,1}^{-1}, \dots , \mathbf Q_{N-1,N}^{-1}, \bm{\mathcal{K}}_N^{-1}),\\
	\mathbf Q_{a,b} &= \int_{t_a}^{t_b} \mathbf \Phi(b,s) \mathbf F(s) \mathbf{Q}_c \mathbf F(s)^\top \mathbf \Phi(b,s)^\top \diff s
\end{align}

%%%%%%%%%%%%%%%%%%%%%%%%%%%%%%%%%%%%%%%%%%%%%%%%%%%%%%%%%%%%%%%%%%%%%%%%%%%%%%%%%%%%%%%%%%%%%%%%%%%%%%%%%%%%%%%%

\section*{Appendix B: Sparsity of the likelihood in GPMP2}

In Eq.~\eqref{eq:linear_system} we argue that matrix $\bm{\mathcal{K}}^{-1} + \mathbf{H}^\top \mathbf{\Sigma}_{obs}^{-1} \mathbf{H}$ is sparse. In Section~\ref{sec:ltvsde}, we proved the block-tridiagonal property of $\bm{\mathcal{K}}^{-1}$. In this section we prove that $\mathbf{H}^\top \mathbf{\Sigma}_{obs}^{-1} \mathbf{H}$ is also {block-tridiagonal}.

Given the isotropic definition of $\mathbf{\Sigma}_{obs}$ in Eq.~\eqref{eq:hyper_obs_matrix} and Eq.~\eqref{eq:hyper_obs_matrix_small} 
\begin{equation}\label{eq:obstacle_linear_system}
\mathbf{H}^\top \mathbf{\Sigma}_{obs}^{-1} \mathbf{H} = \sigma_{obs}^{-2} \mathbf{H}^\top  \mathbf{H}.
\end{equation}
Given the definition of $\bm{h}(\bm{\theta})$ in Eq.~\eqref{eq:vector_obs_cost}, the size of $\mathbf{H}$ is $M \times (N+1 + N\times n_{ip})$ by $(N+1) \times D$, therefore $\mathbf{H}^\top  \mathbf{H}$ has size $(N+1) \times D$ by $(N+1) \times D$.

For simplicity, we partition $\mathbf{H}$ and $\mathbf{H}^\top  \mathbf{H}$ by forming blocks corresponding to the system DOF $D$, and dimensionality $M$ of $\bm{h}$, and work with these block matrices in the remaining section. So $\mathbf{H}$ and $\mathbf{H}^\top \mathbf{H}$ have block-wise size $N+1 + N\times n_{ip}$ by $N+1$ and $N+1$ by $N+1$ respectively.
We define $\mathbf{A}(i,j)$ to be the block element at row $i$ and column $j$ of $\mathbf{A}$.

Given the definition of $\bm{h}(\bm{\theta})$ in Eq.~\eqref{eq:vector_obs_cost}, each element of $\mathbf{H}$ is defined by 
\begin{equation}
\mathbf{H}(i,j) = \frac{\partial \mathbf{h}(\bm{\theta}_{s_i})}{\partial \bm{\theta}_j} \Bigr|_{{\bm{\theta}}}
\end{equation}
for rows contain regular obstacle factors, where $s_i$ is the support state index connects the regular obstacle factor of row $i$, or
\begin{equation}
\mathbf{H}(i,j) = \frac{\partial \mathbf{h}^{intp}(\bm{\theta}_{s_i}, \bm{\theta}_{{s_i}+1})}{\partial \bm{\theta}_j} \Bigr|_{{\bm{\theta}}}
\end{equation}
for rows contain interpolated obstacle factors, where $s_i$ is the before support state index of interpolated obstacle factor of row $i$. Since $\mathbf{h}(\bm{\theta}_{s_i})$ is only a function of $\bm{\theta}_{s_i}$, and $\mathbf{h}^{intp}(\bm{\theta}_{s_i}, \bm{\theta}_{{s_i}+1})$ is only function of $\bm{\theta}_{s_i}$ and $\bm{\theta}_{{s_i}+1}$, they have zero partial derivatives with respect to any other states in $\bm{\theta}$, so for any block element in $\mathbf{H}$
\begin{equation}\label{eq:Hij_zero}
\mathbf{H}(i,j) = \mathbf{0}, \text{ if } j \neq s_i \text{ or } s_i+1.
\end{equation}
For each block element in $\mathbf{H}^\top \mathbf{H}$ 
\begin{align}
\mathbf{H}^\top \mathbf{H}(i,j) &= \sum_{k=1}^{N+1 + N\times n_{ip}} \mathbf{H}^\top(i,k) \mathbf{H}(k,j)  \\
&= \sum_{k=1}^{N+1 + N\times n_{ip}} \mathbf{H}(k,i)^\top \mathbf{H}(k,j).\label{eq:Hij}
\end{align}
For each $k$, non-zero $\mathbf{H}(k,i)^\top \mathbf{H}(k,j)$ is possible when the following condition is satisfied, 
\begin{equation}\label{eq:non_zero_Hkij}
\{ i = s_k \text{ or } s_k+1 \} \text{ and } \{ j = s_k \text{ or } s_k+1 \}.
\end{equation}
So for non-zero $\mathbf{H}^\top \mathbf{H}(i,j)$ 
\begin{equation}\label{eq:non_zero_Hij}
| i-j | \leq 1,
\end{equation}
since if $i$ and $j$ has difference larger than $1$, Eq.~\eqref{eq:non_zero_Hkij} is unsatisfied on every $k$, so $\mathbf{H}^\top \mathbf{H}(i,j)$ will be zero based on Eq.~\eqref{eq:Hij}. Given we know that $\mathbf{H}^\top \mathbf{H}$ is block tridiagonal, and Eq.~\eqref{eq:obstacle_linear_system}, we have proved that $\mathbf{H}^\top \mathbf{\Sigma}_{obs}^{-1} \mathbf{H}$ is also block tridiagonal.

\end{document}